\documentclass{article}

\usepackage[preprint]{neurips_data_2024}

\usepackage[utf8]{inputenc} 
\usepackage[T1]{fontenc}    
\usepackage{hyperref}       
\usepackage{url}            
\usepackage{booktabs}       
\usepackage{amsfonts}       
\usepackage{nicefrac}       
\usepackage{microtype}      
\usepackage{xcolor}         
\usepackage{microtype}
\usepackage{booktabs} 
\usepackage{multirow}
\usepackage{amsmath}
\usepackage{amsthm}
\usepackage{bbding}
\usepackage{graphicx}
\usepackage{makecell}
\usepackage{graphicx}
\usepackage{booktabs}
\usepackage{inconsolata}
\usepackage{subcaption}

\title{FinBen: An Holistic Financial Benchmark for Large Language Models}

\author{
\bf
Qianqian Xie$^{a}$,
Weiguang Han$^b$,  
Zhengyu Chen$^b$,   
Ruoyu Xiang$^a$, 
Xiao Zhang$^a$,
Yueru He$^a$,\\
\bf
Mengxi Xiao$^b$, Dong Li$^b$, Yongfu Dai$^g$, Duanyu Feng$^g$, Yijing Xu$^a$, Haoqiang Kang$^e$,\\
\bf
Ziyan Kuang$^l$, Chenhan Yuan$^c$, Kailai Yang$^c$, Zheheng Luo$^c$, Tianlin Zhang$^c$,\\
\bf
Zhiwei Liu$^c$, Guojun Xiong$^j$, Zhiyang Deng$^i$, Yuechen Jiang$^i$, Zhiyuan Yao$^i$,\\
\bf
Haohang Li$^i$, Yangyang Yu$^i$, Gang Hu$^h$, Jiajia Huang$^k$, Xiao-Yang Liu$^e$,\\
\bf
Alejandro Lopez-Lira$^d$, Benyou Wang$^f$, Yanzhao Lai$^m$, Hao Wang$^g$, Min Peng$^b$,\\
\bf
Sophia Ananiadou$^c$, Jimin Huang$^a$\\
$^a$The Fin AI, $^b$Wuhan University, $^c$The University of Manchester, $^d$University of Florida,\\
$^e$Columbia University, $^f$The Chinese University of Hong Kong, Shenzhen,\\
$^g$Sichuan University, $^h$Yunnan University, $^i$Stevens Institute of Technology\\
$^j$Stony Brook University, $^k$Nanjing Audit University,\\
$^l$Jiangxi Normal University, $^m$Southwest Jiaotong University\\
}

\begin{document}

\maketitle

\begin{abstract}
LLMs have transformed NLP and shown promise in various fields, yet their potential in finance is underexplored due to a lack of comprehensive evaluation benchmarks, the rapid development of LLMs, and the complexity of financial tasks. 
In this paper, we introduce FinBen, the first extensive open-source evaluation benchmark, including 36 datasets spanning 24 financial tasks, covering seven critical aspects: information extraction (IE), textual analysis, question answering (QA), text generation, risk management, forecasting, and decision-making. FinBen offers several key innovations: a broader range of tasks and datasets, the first evaluation of stock trading, novel agent and Retrieval-Augmented Generation (RAG) evaluation, and three novel open-source evaluation datasets for text summarization, question answering, and stock trading.
Our evaluation of 15 representative LLMs, including GPT-4, ChatGPT, and the latest Gemini, reveals several key findings: While LLMs excel in IE and textual analysis, they struggle with advanced reasoning and complex tasks like text generation and forecasting. GPT-4 excels in IE and stock trading, while Gemini is better at text generation and forecasting. Instruction-tuned LLMs improve textual analysis but offer limited benefits for complex tasks such as QA.
FinBen has been used to host the first financial LLMs shared task at the FinNLP-AgentScen workshop during IJCAI-2024, attracting 12 teams. Their novel solutions outperformed GPT-4, showcasing FinBen's potential to drive innovation in financial LLMs. All datasets, results, and codes are released for the research community\footnote{\url{https://github.com/The-FinAI/PIXIU}}.
\end{abstract}

\section{Introduction}
\label{section:introduction}
Recently, Large Language Models (LLMs)~\citep{brown2020language} such as ChatGPT\footnote{\url{https://openai.com/chatgpt}} and GPT-4~\citep{openai2023gpt4}, have reshaped the field of natural language processing (NLP) and exhibited remarkable capabilities in specialized domains across mathematics, coding, medicine, law, and finance~\citep{bubeck2023sparks}.
Within the financial domain, recent several studies~\citep{xie2023wall,lopez2023can,li2023chatgpt,xie2023pixiu} have shown the great potential of advanced LLMs such as GPT-4 on financial text analysis and prediction tasks. 
While their potential is evident, a comprehensive understanding of their capabilities and limitations for finance remains largely unexplored. 
This is due to a lack of extensive evaluation studies and benchmarks, and the inherent complexities associated with the professional nature of financial tasks.

\begin{figure*}[ht]
    \centering
    \begin{minipage}[b]{0.5\textwidth}
        \centering
        \scriptsize
        \captionof{table}{Comparison of different financial benchmarks based on the number of tasks and datasets used, and the task number distribution across various aspects including information extraction (IE), textual analysis (TA), question answering (QA), text generation (TG), risk management (RM), forecasting (FO), and decision-making (DM).}
        \setlength\tabcolsep{1pt}
        \renewcommand{\arraystretch}{1}
        \begin{tabular}{ccccccccccc}
        \toprule
        \multirow{1}{*}{\textbf{Benchmark}} & \textbf{Language} & \multirow{1}{*}{\textbf{Dataset}} & \textbf{Task} & \textbf{IE} & \textbf{TA} & \textbf{QA} & \textbf{TG} & \textbf{RM} & \textbf{FO} & \textbf{DM} \\
        \midrule
        CFBenchmark & Chinese & 8 & 7 & 1 & 3 & \XSolidBrush & 3 & \XSolidBrush & \XSolidBrush & \XSolidBrush \\
        Fin-Eva & Chinese & 1 & 1 & \XSolidBrush & \XSolidBrush & 1 & \XSolidBrush & \XSolidBrush & \XSolidBrush & \XSolidBrush \\
        PIXIU & English & 15 & 8 & 1 & 3 & 1 & 1 & 1 & 1 & \XSolidBrush \\
        FinanceBench & English & 1 & 1 & \XSolidBrush & \XSolidBrush & 1 & \XSolidBrush & \XSolidBrush & \XSolidBrush & \XSolidBrush \\
        BizBench & English & 8 & 5 & 2 & \XSolidBrush & 2 & 1 & \XSolidBrush & \XSolidBrush & \XSolidBrush \\
        FinBen & English & 36 & 24 & 6 & 8 & 3 & 1 & 4 & 1 & 1 \\
        \bottomrule
        \end{tabular}
        \label{tab:com}
    \end{minipage}%
    \hfill
    \begin{minipage}[b]{0.35\textwidth}
        \centering
        \includegraphics[width=\textwidth]{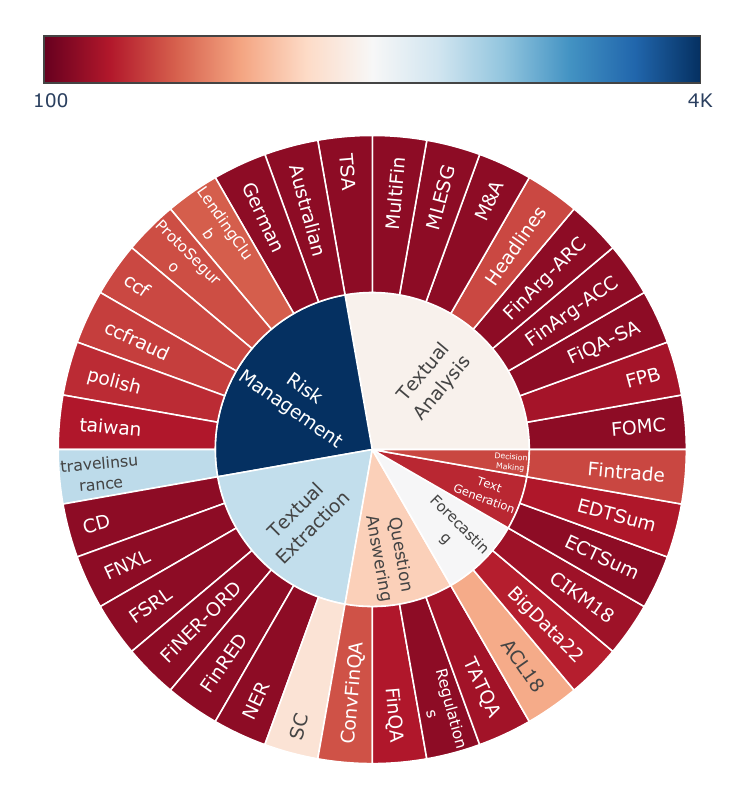}
        \caption{Evaluation datasets for FinBen, with colors indicating dataset sizes ranging from 100 to 4,000.}
        \label{fig:burst}
    \end{minipage}
\end{figure*}

Existing financial domain evaluation benchmarks including PIXIU~\citep{xie2023pixiu}, FinanceBench~\citep{islam2023financebench} and BizBench~\citep{koncel2023bizbench}, have \textbf{limited Evaluation Tasks} and primarily \textbf{focus on Financial NLP Tasks} (As shown in Table~\ref{tab:com}). Most existing benchmarks cover only a small number of evaluation tasks and are centered on evaluating NLP capabilities, such as information extraction and QA. While PIXIU stands out by covering the highest number of tasks, it includes only one evaluation task in most categories. This narrow focus limits their ability to comprehensively evaluate LLMs across the diverse and complex landscape of financial applications, such as forecasting, risk management, and decision-making. It is insufficient for a thorough evaluation of LLM capabilities, especially in the financial area.

To bridge this gap, we propose FinBen, a novel comprehensive open-source evaluation benchmark developed through the collaborative efforts of experts in both computer science and finance. FinBen comprises 36 datasets spanning 24 financial tasks, meticulously organized to assess LLMs across seven critical aspects: information extraction (IE), textual analysis (TA), question answering (QA), text generation (TG), risk management (RM), forecasting (FO), and decision-making (DM). Each category targets specific skills of financial data handling and analysis, ensuring a thorough evaluation of LLMs and showcasing their proficiency in managing complex financial scenarios.

FinBen introduces several innovations compared to existing benchmarks: 1) \textbf{New Tasks}: FinBen introduces a significantly larger number of tasks and datasets, making it the most holistic benchmark for financial LLMs with the highest number of tasks and datasets. This extensive range provides a more robust evaluation of LLM capabilities in diverse financial contexts. 2) \textbf{Broader Coverage}: Covering seven aspects of the financial sector, FinBen is the first benchmark to include the evaluation of stock trading, which is the fundamental task in the financial sector, involving complex decision-making processes that impact market dynamics and investment strategies. 3) \textbf{New Evaluation Strategy}: FinBen is the first benchmark to include agent-based evaluation and Retrieval-Augmented Generation (RAG) based evaluation. These innovative strategies provide a more dynamic and realistic assessment of LLMs, reflecting their ability to interact with and retrieve relevant information from vast datasets. 4) \textbf{Novel Datasets}: FinBen proposes three novel open-source datasets of text summarization, QA, and stock trading tasks for the research community, pushing the boundaries of what LLMs can achieve and setting a new standard for dataset comprehensiveness. 5) \textbf{Empowering Financial LLMs Research}: Based on FinBen, we hosted the first shared task (see Appendix \ref{append:finllm_challenge} for more details) focused on financial LLMs at the FinNLP-AgentScen workshop during IJCAI-2024 \footnote{\url{https://sites.google.com/nlg.csie.ntu.edu.tw/finnlp-agentscen}}. This event attracted 12 teams, leveraging our benchmark to develop novel LLMs-based solutions within the financial domain. Remarkably, the proposed methods achieved superior performance compared to GPT-4, demonstrating the benchmark's potential to foster innovation and advance the state-of-the-art (SOTA) in financial LLMs.

Based on FinBen, we access 15 representative general LLMs such as GPT-4, ChatGPT, and the latest Gemini, and financial LLMs, and have the following findings: 
1) \textbf{Superior Capabilities with Limitations}: While LLMs exhibit exceptional prowess in IE and textual analysis tasks, they underperform in areas necessitating advanced reasoning and complex IE, such as text generation and forecasting.
2) \textbf{Potential in Stock Trading}: SOTA LLMs have demonstrated considerable promise in stock trading applications. However, there remains significant room for improvement due to their limitations in reasoning and comprehensive forecasting abilities.
3) \textbf{Closed-Source Superiority}: Closed-source commercial LLMs continue to lead in performance within the financial domain. Specifically, GPT-4 excels in IE, text analysis, QA, and intricate stock trading tasks, while Gemini shows superior capabilities in text generation and forecasting. 
4) \textbf{Open-Source Improvements and Limitations}: While open-source, instruction-tuned financial LLMs have shown notable enhancements in textual analysis and IE tasks, the advantages of instruction-tuning are less pronounced when it comes to complex tasks such as QA, text generation, and forecasting.

In summary, the main contributions of this paper are: 1) we propose FinBen, the first comprehensive open-sourced evaluation benchmark for LLMs in the financial domain, 2) we propose a novel taxonomy covering seven aspects for organizing financial evaluation tasks, 3) we propose three novel evaluation datasets for the research community, and 4) we conduct systematic evaluation of 15 LLMs using FinBen, showcasing their limitations and advantages, and highlighting directions for future work.

\section{FinBen}
In this section, we delve into the specifics of FinBen, detailing the evaluation taxonomy, data sources, and evaluation tasks.

\subsection{The Taxonomy of Financial Evaluation Tasks}
In the dynamic landscape of financial technology, evaluating the capabilities of LLMs necessitates a comprehensive and structured approach. We propose a novel taxonomy for financial evaluation tasks, categorizing and assessing LLMs across seven financial domains inspired by established taxonomies in financial tasks~\cite{cao2022ai,li2023large,zhao2024revolutionizing}: \textbf{Information Extraction (IE)}, \textbf{Textual Analysis (TA)}, \textbf{Question Answering (QA)}, \textbf{Text Generation (TG)}, \textbf{Risk Management (RM)}, \textbf{Forecasting (FO)}, and \textbf{Decision-Making (DM)}. \textbf{Information Extraction} focuses on identifying key entities and relationships within financial documents, transforming unstructured data into structured insights~\cite{costantino2008information}. \textbf{Textual Analysis} delves into content and sentiment analysis of financial texts, aiding in market trend understanding~\cite{loughran2020textual}. \textbf{Question Answering} evaluates the model's ability to comprehend and respond to financial queries~\cite{maia2018www}. \textbf{Text Generation} assesses the production of coherent financial text~\cite{la2020end}. \textbf{Risk Management} involves evaluating creditworthiness, detecting fraud, and ensuring regulatory compliance~\cite{aziz2019machine}. \textbf{Forecasting} predicts future financial trends, enabling strategic responses to market dynamics~\cite{abu1996introduction}. Finally, \textbf{Decision-Making} assesses the model’s proficiency in making informed financial decisions, such as developing trading strategies and optimizing investment portfolios~\cite{paiva2019decision}.
\subsection{Data Sources}
FinBen's evaluation tasks are drawn from three primary data sources: 1) open-sourced datasets from existing studies comprising open-sourced datasets originally released for non-LLM evaluation settings. Domain experts have designed diverse prompts and reformulated these datasets into instruction-response pairs, making them suitable for evaluating the zero-shot performance of LLMs. 
2) datasets from existing evaluation benchmarks including datasets from existing evaluation benchmarks such as PIXIU. These datasets have already been transformed into the instruction tuning format, allowing for seamless integration and direct use in FinBen. 3) novel datasets introduced in this paper. These datasets are designed to address gaps in existing benchmarks and provide unique challenges for financial LLMs evaluation. Novel datasets include (As shown in Table \ref{tab:eval}): 
\textbf{EDTSum}: We develop the EDTSum dataset for the text summarization task, focusing on financial news. This dataset contains 2,000 news with their headlines as the ground-truth summarization manually selected and cleaned from~\cite{zhou2021trade}. This dataset represents a significant contribution to evaluating the ability of LLMs to generate concise and informative summaries.
\textbf{FinTrade}: We develop the FinTrade dataset specifically for the stock trading task, incorporating historical stock prices, news, and sentiment data. This dataset is the collection of 10 stocks over one year. Our contribution enables the assessments of LLMs in agent-based financial trading scenarios.
\textbf{Regulations}: We develop the regulations dataset for the long-form QA task related to Over-the-Counter (OTC) derivatives and general financial matters within the European Union. It precisely maps these inquiries to relevant articles from the European Market Infrastructure Regulation (EMIR) and other directives, with updates ranging from 2013 to 2021. It tests LLMs' ability to interpret and apply complex regulatory frameworks in real-world scenarios.

\subsection{Tasks}
Table \ref{tab:eval} and Figure \ref{fig:burst} shows all tasks, datasets, data statistics, and evaluation metrics covered by FinBen\footnote{For detailed instructions of each dataset, please see Appendix~\ref{app:instruction}}.

\paragraph{Information extraction:} It spans seven datasets across six information extraction tasks. \textbf{\textit{1) Named entity recognition}} extracts entities like LOCATION, ORGANIZATION, and PERSON from financial agreements and SEC filings, using the NER~\citep{alvarado2015domain} and FINER-ORD~\citep{shah2023finer} datasets. \textbf{\textit{2) Relation extraction}} identifies relationships such as "product/material produced" and "manufacturer" in financial news and earnings transcripts with the FINRED dataset~\citep{sharma2022finred}. \textbf{\textit{3) Causal classification}} discerns whether sentences from financial news and SEC filings convey causality using the SC dataset~\citep{mariko2020financial}. \textbf{\textit{4) Causal detection}} identifies cause and effect spans in financial texts with the CD dataset~\citep{mariko2020financial}. 
\textbf{\textit{5) Numeric labeling}} tags numeric spans in financial documents using the FNXL dataset~\cite{sharma2023financial}, focusing on automating the assignment of labels from a large taxonomy to numeral spans in sentences. \textbf{\textit{6) Textual analogy parsing}} involves identifying common attributes and comparative elements in textual analogies by extracting analogy frames, utilizing the FSRL dataset~\cite{lamm2018textual}, which maps analogous facts to semantic role representations and identifies the analogical relations between them. The evaluation of these tasks is focused on the F1 score~\citep{goutte2005probabilistic}, Entity F1 score~\citep{derczynski2016complementarity}, and the Exact Match Accuracy (EMAcc) metric~\citep{kim2023we}.

\begin{table}[h]
\centering
\scriptsize
\caption{The tasks, datasets, data statistics, and evaluation metrics included in FinBen. We use only test data for evaluation. Datasets marked with an asterisk (*) are newly constructed by us, comprising 10.32\% of the total data. EM Accuracy means the exact match accuracy.}
\scalebox{0.85}{
\begin{tabular}{lllll}
			\toprule
			\textbf{Data}
			&\textbf{Task}
			&\textbf{Test}
			&\textbf{Evaluation}
            &\textbf{License}\\
			\midrule
			\textcolor{black}{NER}~\citep{alvarado2015domain}
			&named entity recognition
			&980
			&Entity F1
            &CC BY-SA 3.0\\
			
			\textcolor{black}{FiNER-ORD}~\citep{shah2023finer}
			&\textcolor{black}{named entity recognition}
			&1080
			&\textcolor{black}{Entity F1}
            &CC BY-NC 4.0\\
			
			\textcolor{black}{FinRED~\citep{sharma2022finred}}
			&\textcolor{black}{relation extraction}
			&1,068
			&\textcolor{black}{F1, Entity F1}
            &Public\\
   
           \textcolor{black}{SC~\citep{mariko2020financial}}
			&\textcolor{black}{causal classification}
			&8,630
			&\textcolor{black}{F1,Entity F1}
            &CC BY 4.0\\

            \textcolor{black}{CD~\citep{mariko2020financial}}
			&\textcolor{black}{causal detection}
			&226
			&\textcolor{black}{F1,Entity F1}
            &CC BY 4.0\\
            
              \textcolor{black}{FNXL~\citep{sharma2023financial}}
			&\textcolor{black}{numeric labeling}
			&318
			&\textcolor{black}{F1,EM Accuracy}
            &Public\\

            \textcolor{black}{FSRL~\citep{lamm2018textual}}
			&\textcolor{black}{textual analogy parsing}
			&97
			&\textcolor{black}{F1, EM Accuracy}
            &MIT License\\
            \midrule

FPB~\citep{malo2014good}
			&sentiment analysis
			&970
			&F1, Accuracy
            &CC BY-SA 3.0\\
			
			FiQA-SA~\citep{maia2018www}
			&sentiment analysis
			&235
			&F1
            &Public\\

   \textcolor{black}{TSA~\citep{cortis2017semeval}}
			&\textcolor{black}{sentiment analysis}
			&561
			&\textcolor{black}{F1, Accuracy}
            &CC BY-NC-SA 4.0\\

\textcolor{black}{Headlines}~\citep{sinha2021impact}
			&news headline classification
			&2,283
			&Avg F1
            &CC BY-SA 3.0\\

            \textcolor{black}{FOMC~\citep{shah2023trillion}}
			&\textcolor{black}{hawkish-dovish classification}
			&496
			&\textcolor{black}{F1, Accuracy}
            &CC BY-NC 4.0\\
            
            \textcolor{black}{FinArg-ACC~\citep{sy2023fine}}
			&\textcolor{black}{argument unit classification}
			&969
			&\textcolor{black}{F1, Accuracy}
            &CC BY-NC-SA 4.0\\

            \textcolor{black}{FinArg-ARC~\citep{sy2023fine}}
			&\textcolor{black}{argument relation classification}
			&496
			&\textcolor{black}{F1, Accuracy}
            &CC BY-NC-SA 4.0\\

            \textcolor{black}{MultiFin~\citep{jorgensen2023multifin}}
			&\textcolor{black}{multi-class classification}
			&690
			&\textcolor{black}{F1, Accuracy}
            &Public\\

            \textcolor{black}{MA~\citep{yang2020generating}}
			&\textcolor{black}{deal completeness classification}
			&500
			&\textcolor{black}{F1, Accuracy}
            &Public\\

            \textcolor{black}{MLESG~\citep{chen2023multi}}
			&\textcolor{black}{ESG Issue Identification}
			&300
			&\textcolor{black}{F1, Accuracy}
            &CC BY-NC-ND\\
\midrule
            
			FinQA~\citep{chen2021finqa}
			&question answering
			&1,147
			&EM Accuracy
            &MIT License\\
            
\textcolor{black}{TATQA~\citep{zhu2021tat}}
			&\textcolor{black}{question answering}
			&1,668
			&\textcolor{black}{F1, EM Accuracy}
            &MIT License\\

            *Regulations
            & long-form question answering
            & 254
            & ROUGE, BERTScore
            & Public \\
            
ConvFinQA~\citep{chen2022convfinqa}
			&multi-turn question answering
			&1,490
			&EM Accuracy
            &MIT License\\
			\midrule

   \textcolor{black}{ECTSum~\citep{mukherjee2022ectsum}}
			&\textcolor{black}{text summarization}
			&495
			&\textcolor{black}{ROUGE, BERTScore, BARTScore}
            &Public\\
			
			\textcolor{black}{*EDTSum}
			&\textcolor{black}{text summarization}
			&2000
			&\textcolor{black}{ROUGE, BERTScore, BARTScore}
            &Public\\
   \midrule
   
			BigData22~\citep{soun2022accurate}
			&stock movement prediction
			&1,470
			&\textcolor{black}{Accuracy}, MCC
            &Public\\
			
			ACL18~\citep{xu2018stock}
			&stock movement prediction
			&3,720
			&\textcolor{black}{Accuracy}, MCC
            &MIT License\\
			
			CIKM18~\citep{wu2018hybrid}
			&stock movement prediction
			&1,140
			&\textcolor{black}{Accuracy}, MCC
            &Public\\
   \midrule			

			\textcolor{black}{German~\citep{misc_german_credit_data_144}}
			&\textcolor{black}{credit scoring}
			&1000
			&\textcolor{black}{F1, MCC}
            &CC BY 4.0\\
			
			\textcolor{black}{Australian~\citep{misc_australian_credit_approval_143}}
			&\textcolor{black}{credit scoring}
			&690
			&\textcolor{black}{F1, MCC}
            &CC BY 4.0\\

            \textcolor{black}{LendingClub~\citep{feng2023empowering}}
			&\textcolor{black}{credit scoring}
			&2,690
			&\textcolor{black}{F1, MCC}
            &CC0 1.0\\

            \textcolor{black}{ccf~\citep{feng2023empowering}}
			&\textcolor{black}{fraud detection}
			&2,278
			&\textcolor{black}{F1, MCC}
            &(DbCL) v1.0\\

            \textcolor{black}{ccfraud~\citep{feng2023empowering}}
			&\textcolor{black}{fraud detection}
			&2,097
			&\textcolor{black}{F1, MCC}
            &Public\\

            \textcolor{black}{polish~\citep{feng2023empowering}}
			&\textcolor{black}{financial distress identification}
			&1,736
			&\textcolor{black}{F1, MCC}
            &CC BY 4.0\\

            \textcolor{black}{taiwan~\citep{feng2023empowering}}
			&\textcolor{black}{financial distress identification}
			&1,364
			&\textcolor{black}{F1, MCC}
            &CC BY 4.0\\

           \textcolor{black}{ProtoSeguro~\citep{feng2023empowering}}
			&\textcolor{black}{	claim analysis}
			&2,381
			&\textcolor{black}{F1, MCC}
            &Public\\

            \textcolor{black}{travelinsurance~\citep{feng2023empowering}}
			&\textcolor{black}{	claim analysis}
			&3,800
			&\textcolor{black}{F1, MCC}
            &(ODbL) v1.0\\
   \midrule
            \textcolor{black}{*FinTrade}
			&\textcolor{black}{stock trading}
			&3,384
			&\makecell[l]{CR, SR, DV, AV, MD}
            &MIT License\\
			
			\bottomrule
		\end{tabular}}
\label{tab:eval}
\end{table}

\paragraph{Textual analysis:} This encompasses eight classification tasks for evaluating LLMs. \textbf{\textit{1) Sentiment analysis}} focuses on extracting sentiment information (positive, negative, or neutral) from financial texts, using three datasets: the Financial Phrase Bank (FPB)~\citep{malo2014good}, FiQA-SA~\citep{maia2018www}, and TSA~\citep{cortis2017semeval} dataset. \textbf{\textit{2) News headline classification}} analyzes additional information, like price movements in financial texts, using the Headlines dataset~\citep{sinha2021impact}.
\textbf{\textit{3) Hawkish-Dovish classification}} aims to classify sentences from monetary policy texts as 'hawkish' or 'dovish' focusing on the nuanced language and economic implications of financial texts, using the FOMC~\citep{shah2023trillion} dataset. \textbf{\textit{4) Argument unit classification}} categorizes sentences as claims or premises using the FinArg AUC dataset~\citep{sy2023fine}. \textbf{\textit{5) Argument relation detection}} identifies relationships (attack, support, or irrelevant) between social media posts using the FinArg ARC dataset~\citep{sy2023fine}. \textbf{\textit{6) Multi-class classification}} targets categorizing a variety of financial texts, including analyst reports, news articles, and investor comments, utilizing the MultiFin dataset~\citep{jorgensen2023multifin}. \textbf{\textit{7) Deal completeness classification }} predicts if mergers and acquisitions events are "completed" or remain "rumors" based on news and tweets, employing the MA dataset~\citep{yang2020generating}. \textbf{\textit{8) ESG issue identification}} focuses on detecting Environmental, Social, and Governance (ESG) concerns in financial documents using the MLESG dataset~\citep{chen2023multi}. For all datasets, evaluation utilizes the accuracy and F1 Score. 

\paragraph{Question answering.} It includes 3 datasets from 2 QA tasks, challenging LLMs to respond to financial queries. \textbf{\textit{1) QA}} focuses on solving questions through multi-step numerical reasoning with financial reports and tables, utilizing the FinQA~\citep{chen2021finqa} and TATQA~\citep{zhu2021tat} dataset. \textbf{\textit{2) Multi-turn QA}} is an extension of QA with multi-turn questions and answers based on financial earnings reports and tables, using the ConvFinQA dataset~\citep{chen2022convfinqa}. F1 score~\citep{derczynski2016complementarity} and the Exact Match Accuracy (EMAcc) metric~\citep{kim2023we} are used to evaluate these tasks. \textbf{\textit{2) Long-form QA}} involve presenting models with complex, detailed questions that require extensive and nuanced answers, often incorporating legal interpretations and practical applications. In our evaluation, we utilize our newly proposed Regulations dataset, which focuses on intricate questions and answers related to financial regulations like EMIR. We assess the model responses using ROUGE~\citep{lin2004rouge} and BERTScore~\citep{zhang2019bertscore}. 

\textbf{Text generation.} This task assesses the models' ability to produce coherent and informative text. Our focus is on \textbf{\textit{text summarization}}, utilizing the ECTSUM~\citep{mukherjee2022ectsum} dataset for summarizing earnings call transcripts. We also propose a novel dataset, EDTSUM, specifically designed for condensing financial news articles into concise summaries, constructed from original data in~\citep{zhou2021trade}. Evaluation employs ROUGE~\citep{lin2004rouge}, BERTScore~\citep{zhang2019bertscore}, and BART Score~\citep{yuan2021bartscore} to measure alignment, factual consistency, and information retention between machine-generated and expert summaries.

\textbf{Forecasting.} The forecasting task challenges models to adaptively predict future market and investor behaviors from emerging patterns. We focus on the\textbf{\textit{1) Stock movement prediction}} task, forecasting stock directions as either positive or negative, based on historical prices and tweets, utilizing three datasets: BigData22~\citep{soun2022accurate}, ACL18~\citep{xu2018stock} and CIKM18~\citep{wu2018hybrid}. 

\textbf{Risk management}. It challenges LLMs to accurately identify, extract, and analyze relevant risk-related information, interpret numerical data, and understand complex relationships. We include 4 tasks: \textbf{\textit{1) Credit scoring}} classifies individuals as "good" or "bad" credit risks using historical customer data, employing datasets including: German~\citep{misc_german_credit_data_144}, Australia~\citep{misc_australian_credit_approval_143} and LendingClub~\citep{feng2023empowering}. \textbf{\textit{2) Fraud detection}} involve categorizes transactions as "fraudulent" or "non-fraudulent", using two datasets: ccf~\citep{feng2023empowering} and ccFraud~\citep{feng2023empowering}.
\textbf{\textit{3) Financial distress identification}} aims to predict a company's bankruptcy risk, using the polish~\citep{feng2023empowering} and taiwan dataset~\citep{feng2023empowering}. Note that the dataset name describes only the region of the company, and the content within the datasets is in English.
\textbf{\textit{4) Claim analysis}} anonymizes client data for privacy, labeling a "target" to indicate claim status, using two datasets: PortoSeguro~\citep{feng2023empowering} and travelinsurance~\citep{feng2023empowering}.
It is noticed that the dataset name such as German and taiwan, only indicates customer sources and all content is in English.
F1 score and Matthews correlation coefficient (MCC)~\citep{chicco2020advantages} are used for evaluating these tasks. 

\textbf{Decision-making.} Strategic decision-making~\citep{punt2017strategic} evaluates the model's proficiency in synthesizing diverse information to formulate and implement trading strategies, a challenge even for experts. We innovatively introduce the SOTA financial LLM agent FinMem~\citep{yu2023finmem} to evaluate LLMs on the \textbf{\textit{stock trading}} task. We construct the novel FinTrade dataset, containing 10 stocks, simulating real-world trading through historical prices, news, and sentiment analysis. Performance is measured by Cumulative Return (CR)~\citep{ariel1987monthly}, Sharpe Ratio (SR)~\citep{sharpe1998sharpe}, Daily (DV) and Annualized volatility (AV)~\citep{zhou2023forecasting}, and Maximum Drawdown (MD)~\citep{magdon2004maximum}, offering a comprehensive assessment of profitability, risk management, and decision-making prowess.

\section{Evaluation}
We evaluate the zero-shot (from our evaluation) and few-shots (results from previous papers) performance of 15 representative general LLMs and financial LLMs on the FinBen benchmark, including:
1) ChatGPT: A LLM developed by OpenAI.
2) GPT-4~\citep{openai2023gpt}: The SOTA commercialized LLMs proposed by OpenAI.
3) Gemini Pro \cite{team2023gemini}: A multimodal AI LLM with 50T parameters, released by Google.
4) LLaMA2-7/70B-chat~\cite{touvron2023llama}: An open-sourced instruction-following LLM with 7B and 70B parameters developed by MetaAI.
5) LLaMA3-8B\footnote{\url{https://llama.meta.com/llama3/}}: An open-sourced LLMs developed by MetaAI, using more training data than LLaMA2.
6) ChatGLM3-6B \cite{du2022glm}: A conversational LLM with 6B parameters, jointly released by Zhipu AI and Tsinghua KEG.
7) Baichuan2-6B \cite{baichuan2023baichuan2}: An open-source LLM with 6B parameters, launched by Baichuan Intelligent Technology.
8) InternLM-7B \cite{team2023internlm}: An open-sourced 7B parameter base model tailored for practical scenarios, proposed by SenseTime.
9) Falcon-7B \cite{almazrouei2023falcon}: A 7B parameter causal decoder-only LLM model trained on 1500B tokens of RefinedWeb enhanced with curated corpora.
10) Mixtral 8$\times$7B \cite{jiang2024mixtral}: A LLM with the Sparse Mixture of Experts (SMoE) architecture.
11) Code LLaMA-7B \cite{roziere2023code}: An open-source LLM model for generating programming code, launched by Meta AI with 7B parameters.
12) FinGPT \cite{yang2023fingpt}: A 7B instruction finetuned financial LLM based on LLaMA 7B~\cite{touvron2023llama1} with sentiment analysis tasks.
13) FinMA-7B \cite{xie2023pixiu}: A 7B instruction finetuned financial LLM based on LLaMA 7B with multiple NLP and forecasting tasks.
14) DISC-FinLLM \cite{chen2023disc}: An open-sourced financial LLM, fine-tuned from Baichuan-13B-Chat \cite{baichuan2023baichuan2}.
15) CFGPT \cite{li2023cfgpt}: An open-source LLM, specifically designed for the financial sector and trained on Chinese financial datasets, which comprises 7B parameters.

\noindent
\textbf{Experimental Settings}
\label{sec:experimental_setting}
We set the maximum generation tokens for LLMs to 1024 and the batch size to 20,000 for all experiments. These experiments are exclusively conducted on 16 NVIDIA A100 80G GPUs, taking approximately 600 hours to complete. Including the GPT-4 API costs, the total expenditure amounts to approximately \$51,000.

\section{Results}
Table \ref{tab:overre} and Table \ref{table:performance_overview} shows the performance of 12 representative LLMs on all datasets in the FinBen. We also report results of non-LLM methods (traditional methods) in Appendix \ref{append:non_llm_result}.

\subsection{Information Extraction \& Textual Analysis Results}
As shown in Table \ref{tab:overre}, for IE tasks, GPT-4 demonstrates superior performance in named entity recognition tasks, including NER, FINER-ORD, and FinRED. InternLM 7B achieves the best results in causal classification (SC). However, for more complex information extraction tasks, such as causal detection (CD) and numerical understanding (FNXL and FSRL), even GPT-4's performance is limited, with Gemini showing only slightly better results, still falling short of expectations. Additionally, while financial domain-specific LLMs developed by instruction tuning such as FinMA 7B exhibit improvements over general domain LLMs such as LLaMA2 7B-chat, they continue to struggle with both named entity recognition and complex extraction tasks. These findings highlight significant opportunities for advancement in financial causal detection and numerical understanding for LLMs.

\begin{table}[ht]
\centering
\scriptsize
\renewcommand{\arraystretch}{1.2}
\caption{The zero-shot and few-shot performance of different LLMs on the FinBen. All results via our evaluations are the average of three runs. ``-'' represents the result that is currently unable to yield due to model size or availability, and ``*'' represents the result from the previous paper.}	
\scalebox{0.7}{
	\begin{tabular}{llccccccccccccc}
			\toprule
			\textbf{Dataset} 
			&\textbf{Metrics}
			&\makecell{\textbf{Chat}\\\textbf{GPT}}
			&\makecell{\textbf{GPT}\\\textbf{4}}
			&\makecell{\textbf{Gemini}}
   &\makecell{\textbf{LLaMA2}\\\textbf{7B-chat}}
			&\makecell{\textbf{LLaMA2}\\\textbf{70B}}
			&\makecell{\textbf{LLaMA3}\\\textbf{8B}}
			&\makecell{\textbf{FinMA}\\\textbf{7B}}
            &\makecell{\textbf{FinGPT}\\\textbf{7b-lora}}
            &\makecell{\textbf{InternLM}\\\textbf{7B}}
            &\makecell{\textbf{Falcon}\\\textbf{7B}}
            &\makecell{\textbf{Mixtral}\\\textbf{7B}}
            &\makecell{\textbf{CFGPT}\\\textbf{sft-7B-Full}}

			\\\midrule
			
\textcolor{black}{NER}
			&EntityF1&0.77\textcolor{black}{*}&\textbf{0.83}\textcolor{black}{*}&0.61&\textcolor{black}{0.18}&\textcolor{black}{0.04}&\textcolor{black}{0.08}&\textcolor{black}{0.69}&0.00&0.00
            &\textcolor{black}{0.00}&\textcolor{black}{0.24}&\textcolor{black}{0.00}\\

            \textcolor{black}{FINER-ORD}
			&EntityF1&\textcolor{black}{0.28}&\textcolor{black}{\textbf{0.77}}&\textcolor{black}{0.14}&\textcolor{black}{0.02}&\textcolor{black}{0.07}&\textcolor{black}{0.00}&\textcolor{black}{0.00}&\textcolor{black}{0.00}&0.00
            &\textcolor{black}{0.00}&\textcolor{black}{0.05}&\textcolor{black}{0.00}\\
    
 \textcolor{black}{FinRED}
			&F1&\textcolor{black}{0.00}&\textcolor{black}{\textbf{0.02}}&\textcolor{black}{0.00}&\textcolor{black}{0.00}&\textcolor{black}{0.00}&\textcolor{black}{0.00}&\textcolor{black}{0.00}&\textcolor{black}{0.00}&0.00 
            &0.00&\textcolor{black}{0.00}&\textcolor{black}{0.00}\\

            \textcolor{black}{SC}
			&F1&\textcolor{black}{0.80}&\textcolor{black}{{0.81}}&\textcolor{black}{0.74}&\textcolor{black}{0.85}&\textcolor{black}{0.61}&\textcolor{black}{0.69}&\textcolor{black}{0.19}&\textcolor{black}{0.00}&{\textbf{0.88}}
            &0.67&\textcolor{black}{0.83}&\textcolor{black}{{0.15}}\\

            \textcolor{black}{CD}
			&F1&\textcolor{black}{0.00}&\textcolor{black}{0.01}&\textcolor{black}{\textbf{0.03}}&\textcolor{black}{0.00}&\textcolor{black}{0.01}&\textcolor{black}{0.00}&\textcolor{black}{0.00}&\textcolor{black}{0.00}&0.00 
            &0.00&\textcolor{black}{0.00}&\textcolor{black}{0.00}\\

            \textcolor{black}{FNXL}
			&EntityF1&\textcolor{black}{0.00}&\textcolor{black}{0.00}&\textcolor{black}{0.00}&\textcolor{black}{0.00}&\textcolor{black}{0.00}&\textcolor{black}{0.00}&\textcolor{black}{0.00}&\textcolor{black}{0.00}&0.00 
            &0.00&\textcolor{black}{0.00}&\textcolor{black}{0.00}\\

            \textcolor{black}{FSRL}
			&EntityF1&\textcolor{black}{0.00}&\textcolor{black}{0.01}&\textcolor{black}{\textbf{0.03}}&\textcolor{black}{0.00}&\textcolor{black}{0.01}&\textcolor{black}{0.00}&\textcolor{black}{0.00}&\textcolor{black}{0.00}&0.00
            &0.00&\textcolor{black}{0.00}&\textcolor{black}{0.00}&\\  
\midrule  
			\multirow{2}{*}{FPB}
			&F1&0.78\textcolor{black}{*}&0.78\textcolor{black}{*}&0.77&\textcolor{black}{0.39}&{0.73}&\textcolor{black}{0.52}&\textbf{0.88}&0.00&0.69
            &\textcolor{black}{0.07}&\textcolor{black}{0.29}&\textcolor{black}{0.35}{*}\\
			&Acc&{0.78}\textcolor{black}{*}&0.76\textcolor{black}{*}&\textcolor{black}{0.77}&\textcolor{black}{0.41}&{0.72}&\textcolor{black}{0.52}&\textbf{0.88}&0.00&0.69
            &\textcolor{black}{0.05}&\textcolor{black}{0.37}&\textcolor{black}{0.26}{*}\\
   
			\multirow{1}{*}{FiQA-SA}
			&F1&0.60&0.80&0.81&\textcolor{black}{0.76}&\textbf{0.83}&\textcolor{black}{0.70}&\textcolor{black}{0.79}&{0.00}&0.81
            &\textcolor{black}{0.77}&\textcolor{black}{0.16}&\textcolor{black}{0.42}{*}\\
            
             \multirow{1}{*}{TSA}
			&RMSE$\downarrow$&0.53&0.50&0.37&\textcolor{black}{0.71}&0.57&\textcolor{black}{0.25}&0.80&0.00&\textbf{0.29}
            &0.50&\textcolor{black}{0.50}&\textcolor{black}{1.05}\\
            
            Headlines&AvgF1&0.77\textcolor{black}{*}&{0.86}\textcolor{black}{*}&0.78&\textcolor{black}{0.72}&{0.63}&\textcolor{black}{0.60}&\textbf{0.97}&{0.60}&0.60
            &\textcolor{black}{0.45}&\textcolor{black}{0.60}&\textcolor{black}{0.61}{*}\\

            \multirow{2}{*}{FOMC}
			&F1&\textcolor{black}{0.64}&\textcolor{black}{\textbf{0.71}}&\textcolor{black}{0.40}&\textcolor{black}{0.35}&\textcolor{black}{0.49}&\textcolor{black}{0.40}&\textcolor{black}{0.49}&\textcolor{black}{0.00}&0.36
            &\textcolor{black}{0.30}&\textcolor{black}{0.37}&\textcolor{black}{0.16}{*}\\
			&Acc&\textcolor{black}{0.6}&\textcolor{black}{\textbf{0.69}}&\textcolor{black}{0.60}&\textcolor{black}{0.49}&\textcolor{black}{0.47}&\textcolor{black}{0.41}&\textcolor{black}{0.46}&\textcolor{black}{0.00}&0.35
            &\textcolor{black}{0.30}&\textcolor{black}{0.35}&\textcolor{black}{0.21}{*}\\

            FinArg-ACC
			&MicroF1&0.50\textcolor{black}&\textbf{0.60}\textcolor{black}&0.31&\textcolor{black}{0.46}&{0.58}&\textcolor{black}{0.51}&{0.27}&0.00&0.39
            &0.23&\textcolor{black}{0.39}&\textcolor{black}{0.05}\\

            FinArg-ARC
			&MicroF1&0.39\textcolor{black}&0.40\textcolor{black}&\textbf{0.60}&\textcolor{black}{0.27}&{0.36}&\textcolor{black}{0.28}&{0.08}&0.00&0.33
            &0.32&\textcolor{black}{0.57}&\textcolor{black}{0.05}\\

            \textcolor{black}{MultiFin}&MicroF1&0.59\textcolor{black}&\textbf{0.65}\textcolor{black}&0.62&\textcolor{black}{0.20}&{0.63}&\textcolor{black}{0.39}&{0.14}&0.00&0.34
            &0.09&\textcolor{black}{0.37}&\textcolor{black}{0.05}\\

            \textcolor{black}{MA}
			&MicroF1&0.85\textcolor{black}&0.79\textcolor{black}&0.84&\textcolor{black}{0.70}&\textbf{0.86}&\textcolor{black}{0.34}&{0.45}&0.00&0.78
            &0.39&\textcolor{black}{0.34}&\textcolor{black}{0.25}\\

            \textcolor{black}{MLESG}
			&MicroF1&0.25\textcolor{black}&\textbf{0.35}\textcolor{black}&0.34&\textcolor{black}{0.03}&{0.31}&\textcolor{black}{0.12}&{0.00}&0.00&0.14
            &0.06&\textcolor{black}{0.17}&\textcolor{black}{0.01}&\\
\midrule

			FinQA
			&EmAcc&0.58\textcolor{black}{*}&\textbf{0.63}\textcolor{black}{*}&\textcolor{black}{0.00}&\textcolor{black}{0.00}&0.06&\textcolor{black}{0.00}&\textcolor{black}{0.04}&0.00&0.00
            &\textcolor{black}{0.00}&\textcolor{black}{0.00}&\textcolor{black}{0.00}\\
			 TATQA
			&EmAcc&0.00\textcolor{black}{*}&{0.13}\textcolor{black}{*}&\textcolor{black}{\textbf{0.18}}&\textcolor{black}{0.03}&\textcolor{black}{0.01}&\textcolor{black}{0.01}&0.00&0.00&0.00
            &0.00&\textcolor{black}{0.01}&\textcolor{black}{0.00}\\

            \multirow{2}{*}{Regulations}
			&Rouge-1&\textcolor{black}{0.12}&\textcolor{black}{0.11}&\textcolor{black}{\textbf{-}}&\textcolor{black}{0.24}&\textcolor{black}{-}&\textcolor{black}{0.10}&\textcolor{black}{0.12}&\textcolor{black}{0.01}&0.04
            &\textcolor{black}{0.03}&-&\textcolor{black}{0.14}\\
			&BertScore&\textcolor{black}{0.64}&\textcolor{black}{0.62}&\textcolor{black}{\textbf{-}}&\textcolor{black}{0.65}&\textcolor{black}{-}&\textcolor{black}{0.60}&\textcolor{black}{0.59}&\textcolor{black}{0.40}&0.57
            &\textcolor{black}{0.14}&-&\textcolor{black}{0.57}\\
            
			ConvFinQA
			&EmAcc&0.60\textcolor{black}{*}&\textbf{0.76}\textcolor{black}{*}&0.43&\textcolor{black}{0.00}&0.25&\textcolor{black}{0.00}&\textcolor{black}{0.20}&0.00&0.00
            &\textcolor{black}{0.00}&\textcolor{black}{0.31}&\textcolor{black}{0.01}\\

\midrule
			\multirow{4}{*}{EDTSUM}
			&Rouge-1&\textcolor{black}{0.17}&\textcolor{black}{0.20}&\textcolor{black}{\textbf{0.39}}&\textcolor{black}{0.17}&\textcolor{black}{0.25}&\textcolor{black}{0.14}&\textcolor{black}{0.13}&\textcolor{black}{0.00}&0.13
            &\textcolor{black}{0.15}&0.12&\textcolor{black}{0.01}\\
			&BertScore&\textcolor{black}{0.66}&\textcolor{black}{0.67}&\textcolor{black}{\textbf{0.72}}&\textcolor{black}{0.62}&\textcolor{black}{0.68}&\textcolor{black}{0.60}&\textcolor{black}{0.38}&\textcolor{black}{0.52}&0.48
            &\textcolor{black}{0.57}&0.61&\textcolor{black}{0.51}\\
			&BartScore&\textcolor{black}{-3.64}&\textcolor{black}{\textbf{-3.62}}&\textcolor{black}{-3.87}&\textcolor{black}{-3.99}&\textcolor{black}{-3.81}&\textcolor{black}{-4.94}&\textcolor{black}{-5.71}&\textcolor{black}{-7.23}&-4.60
            &\textcolor{black}{-6.1}&-4.47&\textcolor{black}{-7.08}\\

			\multirow{4}{*}{ECTSUM}
			&Rouge-1&\textcolor{black}{0.00}&\textcolor{black}{0.00}&\textcolor{black}{0.00}&\textcolor{black}{0.00}&\textcolor{black}{0.00}&\textcolor{black}{0.00}&\textcolor{black}{0.00}&\textcolor{black}{0.00}&0.00%
            &\textcolor{black}{0.00}&0.00&\textcolor{black}{0.00}\\
			&BertScore&\textcolor{black}{0.00}&\textcolor{black}{0.00}&\textcolor{black}{0.00}&\textcolor{black}{0.00}&\textcolor{black}{0.00}&\textcolor{black}{0.00}&\textcolor{black}{0.00}&\textcolor{black}{0.00}&0.00
            &\textcolor{black}{0.00}&0.00&\textcolor{black}{0.00}\\
			&BartScore&\textcolor{black}{-5.18}&\textcolor{black}{-5.18}&\textcolor{black}{-4.93}&\textcolor{black}{-5.18}&\textcolor{black}{\textbf{-4.86}}&\textcolor{black}{-5.18}&\textcolor{black}{-5.18}&\textcolor{black}{-5.18}&-5.18
            &-\textcolor{black}{-5.18}&-5.18&\textcolor{black}{-5.18}\\
            \midrule
            
			\multirow{2}{*}{BigData22}
			&Acc&0.53&\textcolor{black}{0.54}&\textbf{0.55}&\textcolor{black}{0.54}&\textcolor{black}{0.47}&\textcolor{black}{0.55}&\textcolor{black}{0.51}&0.45&0.56
            &\textcolor{black}{0.55}&\textcolor{black}{0.46}&\textcolor{black}{0.45}\\
			&MCC&-0.025&0.03&0.04&\textcolor{black}{0.05}&\textcolor{black}{0.00}&\textcolor{black}{0.02}&\textcolor{black}{0.02}&{0.00}&\textbf{0.08}
            &\textcolor{black}{0.00}&\textcolor{black}{0.02}&\textcolor{black}{0.03}\\
			
			\multirow{2}{*}{ACL18}
			&Acc&0.50&\textbf{0.52}&\textcolor{black}{0.52}&\textcolor{black}{0.51}&\textcolor{black}{0.51}&\textcolor{black}{0.52}&\textcolor{black}{0.51}&0.49&0.51
            &\textcolor{black}{0.51}&\textcolor{black}{0.49}&\textcolor{black}{0.48}\\
			&MCC&0.005&0.02&\textbf{0.04}&\textcolor{black}{0.01}&\textcolor{black}{0.01}&{\textcolor{black}{0.02}}&\textcolor{black}{0.03}&0.00&0.02
            &\textcolor{black}{0.00}&\textcolor{black}{0.00}&\textcolor{black}{-0.03}\\
			
			\multirow{2}{*}{CIKM18}
			&Acc&0.55&\textbf{0.57}&\textcolor{black}{0.54}&\textcolor{black}{0.55}&\textcolor{black}{0.49}&\textcolor{black}{0.57}&\textcolor{black}{0.50}&0.42&0.57
            &\textcolor{black}{0.47}&\textcolor{black}{0.42}&\textcolor{black}{0.41}\\
			&MCC&0.01&0.02&\textcolor{black}{0.02}&\textcolor{black}{-0.03}&\textcolor{black}{-0.07}&\textcolor{black}{0.03}&\textcolor{black}{\textbf{0.08}}&0.00&-0.03
            &\textcolor{black}{-0.06}&\textcolor{black}{-0.05}&\textcolor{black}{-0.07}\\\hline

			\multirow{2}{*}{German}
			&F1&\textcolor{black}{0.20}&\textcolor{black}{{0.55}}&\textcolor{black}{0.52}&\textbf{0.57}&\textcolor{black}{0.17}&\textcolor{black}{0.56}&\textcolor{black}{0.17}&\textcolor{black}{0.52}&0.41
            &\textcolor{black}{0.23}&\textcolor{black}{0.53}&\textcolor{black}{0.53}\\
			&MCC&\textcolor{black}{-0.10}&\textcolor{black}{-0.02}&\textcolor{black}{0.00}&\textcolor{black}{\textbf{0.03}}&\textcolor{black}{0.00}&\textcolor{black}{0.05}&\textcolor{black}{0.00}&\textcolor{black}{0.00}&-0.30
            &\textcolor{black}{-0.07}&\textcolor{black}{0.00}&\textcolor{black}{0.00}\\
			
			\multirow{2}{*}{Australian}
			&F1&\textcolor{black}{0.41}&\textcolor{black}{\textbf{0.74}}&\textcolor{black}{0.26}&\textcolor{black}{0.26}&\textcolor{black}{0.41}&\textcolor{black}{0.26}&\textcolor{black}{0.41}&\textcolor{black}{0.38}&0.34
            &\textcolor{black}{0.26}&\textcolor{black}{0.26}&\textcolor{black}{0.29}\\
			&MCC&\textcolor{black}{0.00}&\textcolor{black}{\textbf{0.47}}&\textcolor{black}{0.00}&\textcolor{black}{0.00}&\textcolor{black}{0.00}&\textcolor{black}{0.00}&\textcolor{black}{0.00}&\textcolor{black}{0.11}&0.13
            &\textcolor{black}{0.00}&\textcolor{black}{0.00}&\textcolor{black}{-0.10}\\

            \multirow{2}{*}{LendingClub}
			&F1&\textcolor{black}{0.20}&\textcolor{black}{0.55}&\textcolor{black}{0.65}&\textcolor{black}{\textbf{0.72}}&\textcolor{black}{0.17}&\textcolor{black}{0.10}&\textcolor{black}{0.61}&\textcolor{black}{0.00}&0.59 
            &0.02&\textcolor{black}{0.61}&\textcolor{black}{0.05}\\
			&MCC&\textcolor{black}{-0.10}&\textcolor{black}{-0.02}&\textcolor{black}{\textbf{0.19}}&\textcolor{black}{0.00}&\textcolor{black}{0.00}&\textcolor{black}{-0.15}&\textcolor{black}{0.00}&\textcolor{black}{0.00}&0.15 
            &-0.01&\textcolor{black}{0.08}&\textcolor{black}{0.01}&\\

            \multirow{2}{*}{ccf}
			&F1&\textcolor{black}{0.20}&\textcolor{black}{0.55}&\textcolor{black}{{0.96}}&\textcolor{black}{0.00}&\textcolor{black}{0.17}&\textcolor{black}{{0.01}}&\textcolor{black}{0.00}&\textcolor{black}{1.00}&\textbf{1.00} 
            &0.10&\textcolor{black}{0.00}&\textcolor{black}{0.00}\\
			&MCC&\textcolor{black}{-0.10}&\textcolor{black}{-0.02}&\textcolor{black}{-0.01}&\textcolor{black}{\textbf{0.00}}&\textcolor{black}{0.00}&\textcolor{black}{0.00}&\textcolor{black}{0.00}&\textcolor{black}{0.00}&0.00 
            &0.00&\textcolor{black}{0.00}&\textcolor{black}{0.00}\\

            \multirow{2}{*}{ccfraud}
			&F1&\textcolor{black}{0.20}&\textcolor{black}{0.55}&\textcolor{black}{\textbf{0.90}}&\textcolor{black}{0.25}&\textcolor{black}{0.17}&\textcolor{black}{0.36}&\textcolor{black}{0.01}&\textcolor{black}{0.00}&0.57 
            &0.62&\textcolor{black}{0.48}&\textcolor{black}{0.03}&\\
			&MCC&\textcolor{black}{-0.10}&\textcolor{black}{-0.02}&\textcolor{black}{0.00}&\textcolor{black}{-0.16}&\textcolor{black}{0.00}&\textcolor{black}{-0.03}&\textcolor{black}{-0.06}&\textcolor{black}{0.00}&-0.13 
            &-0.02&\textcolor{black}{\textbf{0.16}}&\textcolor{black}{0.01}\\

            \multirow{2}{*}{polish}
			&F1&\textcolor{black}{0.20}&\textcolor{black}{0.55}&\textcolor{black}{0.86}&\textcolor{black}{0.92}&\textcolor{black}{0.17}&\textcolor{black}{0.83}&\textcolor{black}{0.92}&\textcolor{black}{0.30}&0.92 
            &0.76&\textcolor{black}{\textbf{0.92}}&\textcolor{black}{0.40}\\
			&MCC&\textcolor{black}{-0.10}&\textcolor{black}{-0.02}&\textcolor{black}{\textbf{0.14}}&\textcolor{black}{0.00}&\textcolor{black}{0.00}&\textcolor{black}{-0.06}&\textcolor{black}{-0.01}&\textcolor{black}{0.00}&0.07 
            &0.05&\textcolor{black}{0.00}&\textcolor{black}{-0.02}\\

            \multirow{2}{*}{taiwan}
			&F1&\textcolor{black}{0.20}&\textcolor{black}{0.55}&\textcolor{black}{\textbf{0.95}}&\textcolor{black}{0.95}&\textcolor{black}{0.17}&\textcolor{black}{{0.26}}&\textcolor{black}{0.95}&\textcolor{black}{0.60}&0.95 
            &0.00&\textcolor{black}{0.95}&\textcolor{black}{0.70}\\
			&MCC&\textcolor{black}{-0.10}&\textcolor{black}{-0.02}&\textcolor{black}{0.00}&\textcolor{black}{\textbf{-0.01}}&\textcolor{black}{0.00}&\textcolor{black}{-0.07}&\textcolor{black}{0.00}&\textcolor{black}{{-0.02}}&{-0.01} 
            &0.00&\textcolor{black}{0.00}&\textcolor{black}{0.00}\\

            \multirow{2}{*}{portoseguro}
			&F1&\textcolor{black}{0.20}&\textcolor{black}{0.55}&\textcolor{black}{0.95}&\textcolor{black}{0.01}&\textcolor{black}{0.17}&\textcolor{black}{{0.94}}&\textcolor{black}{0.04}&\textbf{0.96}&{{0.96}} 
            &\textcolor{black}{0.95}&\textcolor{black}{0.72}&\textcolor{black}{0.00}\\
			&MCC&\textcolor{black}{-0.10}&\textcolor{black}{-0.02}&\textcolor{black}{0.00}&\textcolor{black}{-0.05}&\textcolor{black}{0.00}&\textcolor{black}{-0.01}&\textbf{0.01}&\textcolor{black}{{0.00}}&0.00 
            &0.00&\textcolor{black}{0.01}&\textcolor{black}{0.00}\\

            \multirow{2}{*}{travelinsurance}
			&F1&\textcolor{black}{0.20}&\textcolor{black}{0.55}&\textcolor{black}{0.00}&\textcolor{black}{0.00}&\textcolor{black}{0.17}&\textcolor{black}{0.00}&\textcolor{black}{0.00}&\textbf{0.98}&{0.89} 
            &\textcolor{black}{0.77}&\textcolor{black}{0.00}&\textcolor{black}{0.03}\\
			&MCC&\textcolor{black}{-0.10}&\textcolor{black}{-0.02}&\textcolor{black}{0.00}&\textcolor{black}{{0.00}}&\textcolor{black}{0.00}&\textcolor{black}{0.00}&\textcolor{black}{0.00}&\textcolor{black}{0.00}&{\textbf{0.12}} 
            &-0.03&\textcolor{black}{0.00}&\textcolor{black}{0.01}\\
			\bottomrule
        \end{tabular}}
\label{tab:overre}
\end{table}

Regarding TA tasks, instruction fine-tuned models like FinMA 7B exhibit the best performance in sentiment analysis tasks, including FPB, FiQA-SA, and Headlines. However, the generalization ability of FinMA 7B is limited due to the diversity of TA tasks in the financial domain. It performs even worse than general domain LLMs such as LLaMA2-7B-chat on other TA tasks, where GPT-4, Gemini, and LLaMA2 70B show superior results. This underscores the limitations of instruction fine-tuned models, which may also be constrained by the parameter size of their backbone models.

Models tailored for the Chinese language, such as CFGPT sft-7B-Full, which is fine-tuned on Chinese financial data, exhibit limited improvement on some datasets and even a decline in performance on others like MultiFin compared to its base model InternLM 7B. This trend suggests a language-based discrepancy, indicating that fine-tuning with Chinese data may adversely affect performance on English tasks. These findings underscore the complexities of cross-lingual adaptation in model training, highlighting the challenges in achieving consistent performance across different languages.

\subsection{Question Answering \& Text Generation Results}
In the QA tasks, closed-source commercial LLMs like GPT-4 and Gemini continue to lead across all datasets. While FinMA 7B shows improvement over its backbone models, it remains limited by model size and exhibits bottlenecks in numeric reasoning ability. For the regulations dataset, which is the first intersection dataset requiring both financial and legal knowledge, GPT-4 demonstrates its broad knowledge coverage effectively.
 
In the TG tasks, Gemini emerges as the frontrunner on the EDTSUM abstractive text summarization dataset, illustrating its prowess in generating coherent summaries. Nevertheless, all models face challenges with extractive summarization, which demands the generation of precise label sequences for sentences. Among open-source LLMs, LLaMA2 70B stands out in text summarization. Conversely, CFGPT sft-7B-Full consistently shows a decrease in performance compared to its foundational model, InternLM 7B.

\subsection{Forecasting \& Risk Management Results}
For forecasting, it is crucial to acknowledge that all LLMs fail to meet expected outcomes and lag behind traditional methodologies. This consistent observation with existing studies~\cite {xie2023pixiu} underlines a notable deficiency in LLMs' capacity to tackle forecasting as effectively as traditional methods. Even the best-performing models, such as GPT-4 and Gemini, only perform slightly better than random guessing. This reveals significant potential for enhancement in LLMs, including industry leaders like GPT-4 and Gemini, particularly in forecasting tasks that demand complex reasoning abilities.

In RM tasks, such as credit scoring, fraud detection, and identifying financial distress, data often exhibit significant imbalances. Instances representing individuals with low credit scores, those prone to fraud, and companies at risk of financial distress constitute only a small percentage of the overall dataset. In such scenarios, LLMs with low instruction-following abilities (such as LLaMA2-7B-chat and LLaMA2-70B) tend to classify all cases into a single class, resulting in an MCC score of 0. These tasks, with tabular inputs and highly imbalanced distribution, pose a significant challenge for LLMs in the financial domain.

\subsection{Decision Making Results}
The comparative analysis of various LLMs on the complex task of stock trading, is presented in Table \ref{table:performance_overview}\footnote{For detailed trading performance, please see Appendix~\ref{app:trading}}. This task requires models to understand, summarize, and reason with multimodal financial data (texts and time series), leading to sophisticated trading decisions that necessitate a range of skills, from fundamental comprehension and summarization to reasoning and decision-making. 

Among the evaluated LLMs, GPT-4 distinguishes itself by achieving the highest Sharpe Ratio (SR) over 1, indicating superior investment performance through optimal risk-return balance. It also records the minimal Maximum Drawdown (MDD), suggesting effective limitation of potential losses, thereby offering a more secure investment avenue compared to other models, including those using reinforcement learning methods like DQN, PPO, and A2C, which show significantly lower SR and higher MDD. 

Tables \ref{table:performance_overview} and \ref{tab:trading_all} reinforce these findings, highlighting GPT-4's exceptional performance in this challenging domain. Additional results and analyses from these models in Table \ref{tab:additional_trading} contrast their performances with the traditional \textit{Buy \& Hold} strategy, which considerably lags behind.

\begin{table}[h]
		\centering
		\scriptsize
\renewcommand{\arraystretch}{1.2}
 \caption{The average trading performance (95\% Confidence Interval) comparison for different LLMs across 10 stocks. The results include large LLMs only ($\geq 70B$), as models with smaller contexts have difficulty understanding the instructions and producing a static strategy of holding.}
\begin{tabular}{lllllll}
\toprule
 \textbf{Model} & \textbf{CR (\%)$\uparrow$} & \textbf{SR$\uparrow$} & \textbf{DV (\%)$\downarrow$} & \textbf{AV (\%)$\downarrow$} & \textbf{MD (\%)$\downarrow$} \\
\midrule
Buy \& Hold & -4.00 $\pm$ 22.39 & 0.02 $\pm$ 0.87 & 3.59 $\pm$ 1.34 & 56.43 $\pm$ 21.00 & 30.67 $\pm$ 17.48 \\
GPT-4 & \textbf{28.19 $\pm$ 25.27} & \textbf{1.51 $\pm$ 1.08} & 2.52 $\pm$ 1.30 & 39.88 $\pm$ 20.66 & \textbf{18.34 $\pm$ 9.77} \\
GPT-4o & -5.54 $\pm$ 19.12 & -0.19 $\pm$ 0.84 & 2.73 $\pm$ 1.30 & 43.62 $\pm$ 20.67 & 29.96 $\pm$ 18.89 \\
GPT3.5-Turbo & 4.48 $\pm$ 22.23 & 0.15 $\pm$ 0.82 & 2.84 $\pm$ 1.47 & 45.39 $\pm$ 23.35 & 28.83 $\pm$ 15.40 \\
llama2-70B & 4.02 $\pm$ 24.65 & 0.52 $\pm$ 1.48 & 2.18 $\pm$ 1.28 & 34.86 $\pm$ 20.38 & 25.55 $\pm$ 16.83 \\
llama3-70B & -2.57 $\pm$ 22.63 & -0.04 $\pm$ 1.19 & 2.71 $\pm$ 1.54 & 43.42 $\pm$ 24.65 & 29.31 $\pm$ 15.57 \\
gemini & 14.95 $\pm$ 28.04 & 1.03 $\pm$ 1.24 & \textbf{2.17 $\pm$ 1.39} & \textbf{34.67 $\pm$ 22.23} & 20.13 $\pm$ 11.36 \\
\bottomrule
\end{tabular}
\label{table:performance_overview}
\end{table}

\begin{table}[h]
\centering
\small
\caption{Traditional model performances on stock trading.}
\renewcommand{\arraystretch}{1.2}
\scalebox{0.8}{
\begin{tabular}{cccccc}
\toprule
Model 
& Cumulative Return  
& Sharpe Ratio 
& Standard Deviation
& Annualized Volatility 
& Max Drawdown 
\\ \midrule
A2C   
& -4.2232                
& -0.2586    
& 2.7522      
& 43.6898      
& 30.5819       
\\ 
PPO   
& -0.5586    
& 0.0085     
& 2.7531      
& 43.7048    
& 28.9496    
\\ 
DQN   
& -2.9924 
& -0.1656   
& 2.7486     
& 43.6319    
& 31.78        
\\ \bottomrule
\end{tabular}}
\label{tab:additional_trading}
\end{table}

In contrast, ChatGPT exhibits significantly lower performance metrics, indicating limitations in its financial decision-making capabilities. Gemini, on the other hand, secures the position of second-best performer, showcasing lower risk and volatility compared to GPT-4, yet maintaining commendable returns. When considering open-source models, LLaMA-70B, despite its lower volatility, yields the least profit among the LLMs, highlighting a trade-off between risk management and profitability.

For smaller models with parameters less than 70 billion, a marked inability to adhere to trading instructions consistently across transactions is noted. This is attributed to their limited comprehension, extraction capabilities, and constrained context windows. This limitation underscores the critical challenges smaller LLMs face in tasks requiring intricate financial reasoning and decision-making, thereby spotlighting the necessity for more advanced models to tackle decision making tasks effectively.

\section{Conclusion}
In this work, we present FinBen, a comprehensive benchmark specifically designed to evaluate LLMs in the financial domain. FinBen includes 36 diverse datasets spanning 24 tasks, meticulously organized to assess LLMs across seven critical aspects: information extraction, textual analysis, question answering, text generation, risk management, forecasting, and decision-making. This breadth of coverage sets FinBen apart from existing financial benchmarks, enabling a more robust and nuanced evaluation of LLM capabilities.
Our evaluation of 15 LLMs, including GPT-4, ChatGPT, and Gemini, reveals their key advantages and limitations, highlighting directions for future work. Looking ahead, FinBen aims to continuously evolve, incorporating additional languages and expanding the range of financial tasks to further enhance its applicability and impact. 

\textbf{Limitations:} We acknowledge several limitations that could impact FinBen's effectiveness and applicability. The restricted size of available datasets may affect the models' financial understanding and generalization across various contexts. Computational constraints limited our evaluation to the LLaMA 70B model, potentially overlooking the capabilities of larger models. Additionally, the tasks are based on American market data and English texts, which may limit the benchmark's applicability to global financial markets. Responsible usage and safeguards are essential to prevent potential misuse, such as financial misinformation or unethical market influence\footnote{For a detailed limitation concerning this work, please see Appendix.}.

\textbf{Ethical Statement:}
The authors take full responsibility for any potential legal issues arising from FinBen's development and dissemination. All data used are publicly available, non-personal, and shared under the MIT license, adhering to privacy and ethical guidelines. This manuscript and associated materials are for academic and educational use only and do not provide financial, legal, or investment advice. The authors disclaim any liability for losses or damages from using the material, and users agree to seek professional consultation and indemnify the authors against any claims arising from its use\footnote{For a detailed ethical and legal statement concerning this work, please see Appendix.}.
\bibliographystyle{ACM-Reference-Format}
\bibliography{neurips_2024}


\begin{thebibliography}{93}


\ifx \showCODEN    \undefined \def \showCODEN     #1{\unskip}     \fi
\ifx \showDOI      \undefined \def \showDOI       #1{#1}\fi
\ifx \showISBNx    \undefined \def \showISBNx     #1{\unskip}     \fi
\ifx \showISBNxiii \undefined \def \showISBNxiii  #1{\unskip}     \fi
\ifx \showISSN     \undefined \def \showISSN      #1{\unskip}     \fi
\ifx \showLCCN     \undefined \def \showLCCN      #1{\unskip}     \fi
\ifx \shownote     \undefined \def \shownote      #1{#1}          \fi
\ifx \showarticletitle \undefined \def \showarticletitle #1{#1}   \fi
\ifx \showURL      \undefined \def \showURL       {\relax}        \fi
\providecommand\bibfield[2]{#2}
\providecommand\bibinfo[2]{#2}
\providecommand\natexlab[1]{#1}
\providecommand\showeprint[2][]{arXiv:#2}

\bibitem[Abu-Mostafa and Atiya(1996)]%
        {abu1996introduction}
\bibfield{author}{\bibinfo{person}{Yaser~S Abu-Mostafa} {and} \bibinfo{person}{Amir~F Atiya}.} \bibinfo{year}{1996}\natexlab{}.
\newblock \showarticletitle{Introduction to financial forecasting}.
\newblock \bibinfo{journal}{\emph{Applied intelligence}}  \bibinfo{volume}{6} (\bibinfo{year}{1996}), \bibinfo{pages}{205--213}.
\newblock


\bibitem[Almazrouei et~al\mbox{.}(2023)]%
        {almazrouei2023falcon}
\bibfield{author}{\bibinfo{person}{Ebtesam Almazrouei}, \bibinfo{person}{Hamza Alobeidli}, \bibinfo{person}{Abdulaziz Alshamsi}, \bibinfo{person}{Alessandro Cappelli}, \bibinfo{person}{Ruxandra Cojocaru}, \bibinfo{person}{M{\'e}rouane Debbah}, \bibinfo{person}{{\'E}tienne Goffinet}, \bibinfo{person}{Daniel Hesslow}, \bibinfo{person}{Julien Launay}, \bibinfo{person}{Quentin Malartic}, {et~al\mbox{.}}} \bibinfo{year}{2023}\natexlab{}.
\newblock \showarticletitle{The falcon series of open language models}.
\newblock \bibinfo{journal}{\emph{arXiv preprint arXiv:2311.16867}} (\bibinfo{year}{2023}).
\newblock


\bibitem[Alvarado et~al\mbox{.}(2015)]%
        {alvarado2015domain}
\bibfield{author}{\bibinfo{person}{Julio Cesar~Salinas Alvarado}, \bibinfo{person}{Karin Verspoor}, {and} \bibinfo{person}{Timothy Baldwin}.} \bibinfo{year}{2015}\natexlab{}.
\newblock \showarticletitle{Domain adaption of named entity recognition to support credit risk assessment}. In \bibinfo{booktitle}{\emph{Proceedings of the Australasian Language Technology Association Workshop 2015}}. \bibinfo{pages}{84--90}.
\newblock


\bibitem[Araci(2019)]%
        {araci2019finbert}
\bibfield{author}{\bibinfo{person}{Dogu Araci}.} \bibinfo{year}{2019}\natexlab{}.
\newblock \bibinfo{title}{FinBERT: Financial Sentiment Analysis with Pre-trained Language Models}.
\newblock
\newblock
\showeprint[arxiv]{1908.10063}~[cs.CL]


\bibitem[Ariel(1987)]%
        {ariel1987monthly}
\bibfield{author}{\bibinfo{person}{Robert~A Ariel}.} \bibinfo{year}{1987}\natexlab{}.
\newblock \showarticletitle{A monthly effect in stock returns}.
\newblock \bibinfo{journal}{\emph{Journal of financial economics}} \bibinfo{volume}{18}, \bibinfo{number}{1} (\bibinfo{year}{1987}), \bibinfo{pages}{161--174}.
\newblock


\bibitem[Aziz and Dowling(2019)]%
        {aziz2019machine}
\bibfield{author}{\bibinfo{person}{Saqib Aziz} {and} \bibinfo{person}{Michael Dowling}.} \bibinfo{year}{2019}\natexlab{}.
\newblock \bibinfo{booktitle}{\emph{Machine learning and AI for risk management}}.
\newblock \bibinfo{publisher}{Springer International Publishing}.
\newblock


\bibitem[Baichuan(2023)]%
        {baichuan2023baichuan2}
\bibfield{author}{\bibinfo{person}{Baichuan}.} \bibinfo{year}{2023}\natexlab{}.
\newblock \showarticletitle{Baichuan 2: Open Large-scale Language Models}.
\newblock \bibinfo{journal}{\emph{arXiv preprint arXiv:2309.10305}} (\bibinfo{year}{2023}).
\newblock
\urldef\tempurl%
\url{https://arxiv.org/abs/2309.10305}
\showURL{%
\tempurl}


\bibitem[Brown et~al\mbox{.}(2020)]%
        {brown2020language}
\bibfield{author}{\bibinfo{person}{Tom Brown}, \bibinfo{person}{Benjamin Mann}, \bibinfo{person}{Nick Ryder}, \bibinfo{person}{Melanie Subbiah}, \bibinfo{person}{Jared~D Kaplan}, \bibinfo{person}{Prafulla Dhariwal}, \bibinfo{person}{Arvind Neelakantan}, \bibinfo{person}{Pranav Shyam}, \bibinfo{person}{Girish Sastry}, \bibinfo{person}{Amanda Askell}, {et~al\mbox{.}}} \bibinfo{year}{2020}\natexlab{}.
\newblock \showarticletitle{Language models are few-shot learners}.
\newblock \bibinfo{journal}{\emph{Advances in neural information processing systems}}  \bibinfo{volume}{33} (\bibinfo{year}{2020}), \bibinfo{pages}{1877--1901}.
\newblock


\bibitem[Bubeck et~al\mbox{.}(2023)]%
        {bubeck2023sparks}
\bibfield{author}{\bibinfo{person}{S{\'e}bastien Bubeck}, \bibinfo{person}{Varun Chandrasekaran}, \bibinfo{person}{Ronen Eldan}, \bibinfo{person}{Johannes Gehrke}, \bibinfo{person}{Eric Horvitz}, \bibinfo{person}{Ece Kamar}, \bibinfo{person}{Peter Lee}, \bibinfo{person}{Yin~Tat Lee}, \bibinfo{person}{Yuanzhi Li}, \bibinfo{person}{Scott Lundberg}, {et~al\mbox{.}}} \bibinfo{year}{2023}\natexlab{}.
\newblock \showarticletitle{Sparks of artificial general intelligence: Early experiments with gpt-4}.
\newblock \bibinfo{journal}{\emph{arXiv preprint arXiv:2303.12712}} (\bibinfo{year}{2023}).
\newblock


\bibitem[Cao(2022)]%
        {cao2022ai}
\bibfield{author}{\bibinfo{person}{Longbing Cao}.} \bibinfo{year}{2022}\natexlab{}.
\newblock \showarticletitle{Ai in finance: challenges, techniques, and opportunities}.
\newblock \bibinfo{journal}{\emph{ACM Computing Surveys (CSUR)}} \bibinfo{volume}{55}, \bibinfo{number}{3} (\bibinfo{year}{2022}), \bibinfo{pages}{1--38}.
\newblock


\bibitem[Chen et~al\mbox{.}(2023a)]%
        {chen2023multi}
\bibfield{author}{\bibinfo{person}{Chung-Chi Chen}, \bibinfo{person}{Yu-Min Tseng}, \bibinfo{person}{Juyeon Kang}, \bibinfo{person}{Ana{\"\i}s Lhuissier}, \bibinfo{person}{Min-Yuh Day}, \bibinfo{person}{Teng-Tsai Tu}, {and} \bibinfo{person}{Hsin-Hsi Chen}.} \bibinfo{year}{2023}\natexlab{a}.
\newblock \showarticletitle{Multi-Lingual ESG Issue Identification}. In \bibinfo{booktitle}{\emph{Proceedings of the Fifth Workshop on Financial Technology and Natural Language Processing and the Second Multimodal AI For Financial Forecasting}}. \bibinfo{pages}{111--115}.
\newblock


\bibitem[Chen et~al\mbox{.}(2023b)]%
        {chen2023disc}
\bibfield{author}{\bibinfo{person}{Wei Chen}, \bibinfo{person}{Qiushi Wang}, \bibinfo{person}{Zefei Long}, \bibinfo{person}{Xianyin Zhang}, \bibinfo{person}{Zhongtian Lu}, \bibinfo{person}{Bingxuan Li}, \bibinfo{person}{Siyuan Wang}, \bibinfo{person}{Jiarong Xu}, \bibinfo{person}{Xiang Bai}, \bibinfo{person}{Xuanjing Huang}, {et~al\mbox{.}}} \bibinfo{year}{2023}\natexlab{b}.
\newblock \showarticletitle{Disc-finllm: A chinese financial large language model based on multiple experts fine-tuning}.
\newblock \bibinfo{journal}{\emph{arXiv preprint arXiv:2310.15205}} (\bibinfo{year}{2023}).
\newblock


\bibitem[Chen et~al\mbox{.}(2022a)]%
        {chen2022finqa}
\bibfield{author}{\bibinfo{person}{Zhiyu Chen}, \bibinfo{person}{Wenhu Chen}, \bibinfo{person}{Charese Smiley}, {and} \bibinfo{person}{et~al. Sameena~Shah}.} \bibinfo{year}{2022}\natexlab{a}.
\newblock \bibinfo{title}{FinQA: A Dataset of Numerical Reasoning over Financial Data}.
\newblock
\newblock
\showeprint[arxiv]{2109.00122}~[cs.CL]


\bibitem[Chen et~al\mbox{.}(2021)]%
        {chen2021finqa}
\bibfield{author}{\bibinfo{person}{Zhiyu Chen}, \bibinfo{person}{Wenhu Chen}, \bibinfo{person}{Charese Smiley}, \bibinfo{person}{Sameena Shah}, \bibinfo{person}{Iana Borova}, \bibinfo{person}{Dylan Langdon}, \bibinfo{person}{Reema Moussa}, \bibinfo{person}{Matt Beane}, \bibinfo{person}{Ting-Hao Huang}, \bibinfo{person}{Bryan~R Routledge}, {et~al\mbox{.}}} \bibinfo{year}{2021}\natexlab{}.
\newblock \showarticletitle{FinQA: A Dataset of Numerical Reasoning over Financial Data}. In \bibinfo{booktitle}{\emph{Proceedings of the 2021 Conference on Empirical Methods in Natural Language Processing}}. \bibinfo{pages}{3697--3711}.
\newblock


\bibitem[Chen et~al\mbox{.}(2022b)]%
        {chen2022convfinqa}
\bibfield{author}{\bibinfo{person}{Zhiyu Chen}, \bibinfo{person}{Shiyang Li}, \bibinfo{person}{Charese Smiley}, \bibinfo{person}{Zhiqiang Ma}, \bibinfo{person}{Sameena Shah}, {and} \bibinfo{person}{William~Yang Wang}.} \bibinfo{year}{2022}\natexlab{b}.
\newblock \bibinfo{title}{ConvFinQA: Exploring the Chain of Numerical Reasoning in Conversational Finance Question Answering}.
\newblock
\newblock
\showeprint[arxiv]{2210.03849}~[cs.CL]


\bibitem[Chicco and Jurman(2020)]%
        {chicco2020advantages}
\bibfield{author}{\bibinfo{person}{Davide Chicco} {and} \bibinfo{person}{Giuseppe Jurman}.} \bibinfo{year}{2020}\natexlab{}.
\newblock \showarticletitle{The advantages of the Matthews correlation coefficient (MCC) over F1 score and accuracy in binary classification evaluation}.
\newblock \bibinfo{journal}{\emph{BMC genomics}} \bibinfo{volume}{21}, \bibinfo{number}{1} (\bibinfo{year}{2020}), \bibinfo{pages}{1--13}.
\newblock


\bibitem[Cortis et~al\mbox{.}(2017)]%
        {cortis2017semeval}
\bibfield{author}{\bibinfo{person}{Keith Cortis}, \bibinfo{person}{Andr{\'e} Freitas}, \bibinfo{person}{Tobias Daudert}, \bibinfo{person}{Manuela Huerlimann}, \bibinfo{person}{Manel Zarrouk}, \bibinfo{person}{Siegfried Handschuh}, {and} \bibinfo{person}{Brian Davis}.} \bibinfo{year}{2017}\natexlab{}.
\newblock \showarticletitle{Semeval-2017 task 5: Fine-grained sentiment analysis on financial microblogs and news}. In \bibinfo{booktitle}{\emph{Proceedings of the 11th international workshop on semantic evaluation (SemEval-2017)}}. \bibinfo{pages}{519--535}.
\newblock


\bibitem[Costantino and Coletti(2008)]%
        {costantino2008information}
\bibfield{author}{\bibinfo{person}{Marco Costantino} {and} \bibinfo{person}{Paolo Coletti}.} \bibinfo{year}{2008}\natexlab{}.
\newblock \bibinfo{booktitle}{\emph{Information extraction in finance}}. Vol.~\bibinfo{volume}{8}.
\newblock \bibinfo{publisher}{Wit Press}.
\newblock


\bibitem[Dai et~al\mbox{.}(2024)]%
        {dai2024laiw}
\bibfield{author}{\bibinfo{person}{Yongfu Dai}, \bibinfo{person}{Duanyu Feng}, \bibinfo{person}{Jimin Huang}, \bibinfo{person}{Haochen Jia}, \bibinfo{person}{Qianqian Xie}, \bibinfo{person}{Yifang Zhang}, \bibinfo{person}{Weiguang Han}, \bibinfo{person}{Wei Tian}, {and} \bibinfo{person}{Hao Wang}.} \bibinfo{year}{2024}\natexlab{}.
\newblock \bibinfo{title}{LAiW: A Chinese Legal Large Language Models Benchmark}.
\newblock
\newblock
\showeprint[arxiv]{2310.05620}~[cs.CL]


\bibitem[Derczynski(2016)]%
        {derczynski2016complementarity}
\bibfield{author}{\bibinfo{person}{Leon Derczynski}.} \bibinfo{year}{2016}\natexlab{}.
\newblock \showarticletitle{Complementarity, {F}-score, and {NLP} Evaluation}. In \bibinfo{booktitle}{\emph{Proceedings of the Tenth International Conference on Language Resources and Evaluation ({LREC}'16)}}, \bibfield{editor}{\bibinfo{person}{Nicoletta Calzolari}, \bibinfo{person}{Khalid Choukri}, \bibinfo{person}{Thierry Declerck}, \bibinfo{person}{Sara Goggi}, \bibinfo{person}{Marko Grobelnik}, \bibinfo{person}{Bente Maegaard}, \bibinfo{person}{Joseph Mariani}, \bibinfo{person}{Helene Mazo}, \bibinfo{person}{Asuncion Moreno}, \bibinfo{person}{Jan Odijk}, {and} \bibinfo{person}{Stelios Piperidis}} (Eds.). \bibinfo{publisher}{European Language Resources Association (ELRA)}, \bibinfo{address}{Portoro{\v{z}}, Slovenia}, \bibinfo{pages}{261--266}.
\newblock
\urldef\tempurl%
\url{https://aclanthology.org/L16-1040}
\showURL{%
\tempurl}


\bibitem[Du et~al\mbox{.}(2022)]%
        {du2022glm}
\bibfield{author}{\bibinfo{person}{Zhengxiao Du}, \bibinfo{person}{Yujie Qian}, \bibinfo{person}{Xiao Liu}, \bibinfo{person}{Ming Ding}, \bibinfo{person}{Jiezhong Qiu}, \bibinfo{person}{Zhilin Yang}, {and} \bibinfo{person}{Jie Tang}.} \bibinfo{year}{2022}\natexlab{}.
\newblock \showarticletitle{GLM: General Language Model Pretraining with Autoregressive Blank Infilling}. In \bibinfo{booktitle}{\emph{Proceedings of the 60th Annual Meeting of the Association for Computational Linguistics (Volume 1: Long Papers)}}. \bibinfo{pages}{320--335}.
\newblock


\bibitem[Feng et~al\mbox{.}(2023)]%
        {feng2023empowering}
\bibfield{author}{\bibinfo{person}{Duanyu Feng}, \bibinfo{person}{Yongfu Dai}, \bibinfo{person}{Jimin Huang}, \bibinfo{person}{Yifang Zhang}, \bibinfo{person}{Qianqian Xie}, \bibinfo{person}{Weiguang Han}, \bibinfo{person}{Alejandro Lopez-Lira}, {and} \bibinfo{person}{Hao Wang}.} \bibinfo{year}{2023}\natexlab{}.
\newblock \showarticletitle{Empowering many, biasing a few: Generalist credit scoring through large language models}.
\newblock \bibinfo{journal}{\emph{arXiv preprint arXiv:2310.00566}} (\bibinfo{year}{2023}).
\newblock


\bibitem[Goutte and Gaussier(2005)]%
        {goutte2005probabilistic}
\bibfield{author}{\bibinfo{person}{Cyril Goutte} {and} \bibinfo{person}{Eric Gaussier}.} \bibinfo{year}{2005}\natexlab{}.
\newblock \showarticletitle{A probabilistic interpretation of precision, recall and F-score, with implication for evaluation}. In \bibinfo{booktitle}{\emph{European conference on information retrieval}}. Springer, \bibinfo{pages}{345--359}.
\newblock


\bibitem[Han et~al\mbox{.}(2023a)]%
        {han2023mastering}
\bibfield{author}{\bibinfo{person}{Weiguang Han}, \bibinfo{person}{Jimin Huang}, \bibinfo{person}{Qianqian Xie}, \bibinfo{person}{Boyi Zhang}, \bibinfo{person}{Yanzhao Lai}, {and} \bibinfo{person}{Min Peng}.} \bibinfo{year}{2023}\natexlab{a}.
\newblock \bibinfo{title}{Mastering Pair Trading with Risk-Aware Recurrent Reinforcement Learning}.
\newblock
\newblock
\showeprint[arxiv]{2304.00364}~[q-fin.CP]


\bibitem[Han et~al\mbox{.}(2023b)]%
        {han2023select}
\bibfield{author}{\bibinfo{person}{Weiguang Han}, \bibinfo{person}{Boyi Zhang}, \bibinfo{person}{Qianqian Xie}, \bibinfo{person}{Min Peng}, \bibinfo{person}{Yanzhao Lai}, {and} \bibinfo{person}{Jimin Huang}.} \bibinfo{year}{2023}\natexlab{b}.
\newblock \showarticletitle{Select and Trade: Towards Unified Pair Trading with Hierarchical Reinforcement Learning}.
\newblock \bibinfo{journal}{\emph{arXiv preprint arXiv:2301.10724}} (\bibinfo{year}{2023}).
\newblock


\bibitem[Hofmann(1994)]%
        {misc_german_credit_data_144}
\bibfield{author}{\bibinfo{person}{Hans Hofmann}.} \bibinfo{year}{1994}\natexlab{}.
\newblock \bibinfo{title}{{Statlog (German Credit Data)}}.
\newblock \bibinfo{howpublished}{UCI Machine Learning Repository}.
\newblock
\newblock
\shownote{{DOI}: https://doi.org/10.24432/C5NC77}.


\bibitem[Hongyuan et~al\mbox{.}(2023)]%
        {hongyuan-etal-2023-finbart}
\bibfield{author}{\bibinfo{person}{Dong Hongyuan}, \bibinfo{person}{Che Wanxiang}, \bibinfo{person}{He Xiaoyu}, \bibinfo{person}{Zheng Guidong}, {and} \bibinfo{person}{Wen Junjie}.} \bibinfo{year}{2023}\natexlab{}.
\newblock \showarticletitle{{F}in{BART}: A Pre-trained Seq2seq Language Model for {C}hinese Financial Tasks}. In \bibinfo{booktitle}{\emph{Proceedings of the 22nd Chinese National Conference on Computational Linguistics}}, \bibfield{editor}{\bibinfo{person}{Maosong Sun}, \bibinfo{person}{Bing Qin}, \bibinfo{person}{Xipeng Qiu}, \bibinfo{person}{Jing Jiang}, {and} \bibinfo{person}{Xianpei Han}} (Eds.). \bibinfo{publisher}{Chinese Information Processing Society of China}, \bibinfo{address}{Harbin, China}, \bibinfo{pages}{906--917}.
\newblock
\urldef\tempurl%
\url{https://aclanthology.org/2023.ccl-1.77}
\showURL{%
\tempurl}


\bibitem[Islam et~al\mbox{.}(2023)]%
        {islam2023financebench}
\bibfield{author}{\bibinfo{person}{Pranab Islam}, \bibinfo{person}{Anand Kannappan}, \bibinfo{person}{Douwe Kiela}, \bibinfo{person}{Rebecca Qian}, \bibinfo{person}{Nino Scherrer}, {and} \bibinfo{person}{Bertie Vidgen}.} \bibinfo{year}{2023}\natexlab{}.
\newblock \showarticletitle{FinanceBench: A New Benchmark for Financial Question Answering}.
\newblock \bibinfo{journal}{\emph{arXiv preprint arXiv:2311.11944}} (\bibinfo{year}{2023}).
\newblock


\bibitem[Jiang et~al\mbox{.}(2024)]%
        {jiang2024mixtral}
\bibfield{author}{\bibinfo{person}{Albert~Q Jiang}, \bibinfo{person}{Alexandre Sablayrolles}, \bibinfo{person}{Antoine Roux}, \bibinfo{person}{Arthur Mensch}, \bibinfo{person}{Blanche Savary}, \bibinfo{person}{Chris Bamford}, \bibinfo{person}{Devendra~Singh Chaplot}, \bibinfo{person}{Diego de~las Casas}, \bibinfo{person}{Emma~Bou Hanna}, \bibinfo{person}{Florian Bressand}, {et~al\mbox{.}}} \bibinfo{year}{2024}\natexlab{}.
\newblock \showarticletitle{Mixtral of experts}.
\newblock \bibinfo{journal}{\emph{arXiv preprint arXiv:2401.04088}} (\bibinfo{year}{2024}).
\newblock


\bibitem[J{\o}rgensen et~al\mbox{.}(2023)]%
        {jorgensen2023multifin}
\bibfield{author}{\bibinfo{person}{Rasmus J{\o}rgensen}, \bibinfo{person}{Oliver Brandt}, \bibinfo{person}{Mareike Hartmann}, \bibinfo{person}{Xiang Dai}, \bibinfo{person}{Christian Igel}, {and} \bibinfo{person}{Desmond Elliott}.} \bibinfo{year}{2023}\natexlab{}.
\newblock \showarticletitle{MultiFin: A Dataset for Multilingual Financial NLP}. In \bibinfo{booktitle}{\emph{Findings of the Association for Computational Linguistics: EACL 2023}}. \bibinfo{pages}{864--879}.
\newblock


\bibitem[Kim et~al\mbox{.}(2023)]%
        {kim2023we}
\bibfield{author}{\bibinfo{person}{Kisub Kim}, \bibinfo{person}{Xin Zhou}, \bibinfo{person}{Dongsun Kim}, \bibinfo{person}{Julia Lawall}, \bibinfo{person}{Kui Liu}, \bibinfo{person}{Tegawend{\'e}~F Bissyand{\'e}}, \bibinfo{person}{Jacques Klein}, \bibinfo{person}{Jaekwon Lee}, {and} \bibinfo{person}{David Lo}.} \bibinfo{year}{2023}\natexlab{}.
\newblock \showarticletitle{How are We Detecting Inconsistent Method Names? An Empirical Study from Code Review Perspective}.
\newblock \bibinfo{journal}{\emph{arXiv preprint arXiv:2308.12701}} (\bibinfo{year}{2023}).
\newblock


\bibitem[Koncel-Kedziorski et~al\mbox{.}(2023)]%
        {koncel2023bizbench}
\bibfield{author}{\bibinfo{person}{Rik Koncel-Kedziorski}, \bibinfo{person}{Michael Krumdick}, \bibinfo{person}{Viet Lai}, \bibinfo{person}{Varshini Reddy}, \bibinfo{person}{Charles Lovering}, {and} \bibinfo{person}{Chris Tanner}.} \bibinfo{year}{2023}\natexlab{}.
\newblock \showarticletitle{Bizbench: A quantitative reasoning benchmark for business and finance}.
\newblock \bibinfo{journal}{\emph{arXiv preprint arXiv:2311.06602}} (\bibinfo{year}{2023}).
\newblock


\bibitem[La~Quatra and Cagliero(2020)]%
        {la2020end}
\bibfield{author}{\bibinfo{person}{Moreno La~Quatra} {and} \bibinfo{person}{Luca Cagliero}.} \bibinfo{year}{2020}\natexlab{}.
\newblock \showarticletitle{End-to-end training for financial report summarization}. In \bibinfo{booktitle}{\emph{Proceedings of the 1st Joint Workshop on Financial Narrative Processing and MultiLing Financial Summarisation}}. \bibinfo{pages}{118--123}.
\newblock


\bibitem[Lamm et~al\mbox{.}(2018)]%
        {lamm2018textual}
\bibfield{author}{\bibinfo{person}{Matthew Lamm}, \bibinfo{person}{Arun~Tejasvi Chaganty}, \bibinfo{person}{Christopher~D Manning}, \bibinfo{person}{Dan Jurafsky}, {and} \bibinfo{person}{Percy Liang}.} \bibinfo{year}{2018}\natexlab{}.
\newblock \showarticletitle{Textual analogy parsing: What's shared and what's compared among analogous facts}.
\newblock \bibinfo{journal}{\emph{arXiv preprint arXiv:1809.02700}} (\bibinfo{year}{2018}).
\newblock


\bibitem[Lee et~al\mbox{.}(2024)]%
        {lee2024survey}
\bibfield{author}{\bibinfo{person}{Jean Lee}, \bibinfo{person}{Nicholas Stevens}, \bibinfo{person}{Soyeon~Caren Han}, {and} \bibinfo{person}{Minseok Song}.} \bibinfo{year}{2024}\natexlab{}.
\newblock \bibinfo{title}{A Survey of Large Language Models in Finance (FinLLMs)}.
\newblock
\newblock
\showeprint[arxiv]{2402.02315}~[cs.CL]


\bibitem[Lei et~al\mbox{.}(2023)]%
        {lei2023cfbenchmark}
\bibfield{author}{\bibinfo{person}{Yang Lei}, \bibinfo{person}{Jiangtong Li}, \bibinfo{person}{Ming Jiang}, \bibinfo{person}{Junjie Hu}, \bibinfo{person}{Dawei Cheng}, \bibinfo{person}{Zhijun Ding}, {and} \bibinfo{person}{Changjun Jiang}.} \bibinfo{year}{2023}\natexlab{}.
\newblock \bibinfo{title}{CFBenchmark: Chinese Financial Assistant Benchmark for Large Language Model}.
\newblock
\newblock
\showeprint[arxiv]{2311.05812}~[cs.CL]


\bibitem[Li et~al\mbox{.}(2023a)]%
        {li2023cfgpt}
\bibfield{author}{\bibinfo{person}{Jiangtong Li}, \bibinfo{person}{Yuxuan Bian}, \bibinfo{person}{Guoxuan Wang}, \bibinfo{person}{Yang Lei}, \bibinfo{person}{Dawei Cheng}, \bibinfo{person}{Zhijun Ding}, {and} \bibinfo{person}{Changjun Jiang}.} \bibinfo{year}{2023}\natexlab{a}.
\newblock \bibinfo{title}{CFGPT: Chinese Financial Assistant with Large Language Model}.
\newblock
\newblock
\showeprint[arxiv]{2309.10654}~[cs.CL]


\bibitem[Li et~al\mbox{.}(2023c)]%
        {li2023chatgpt}
\bibfield{author}{\bibinfo{person}{Xianzhi Li}, \bibinfo{person}{Xiaodan Zhu}, \bibinfo{person}{Zhiqiang Ma}, \bibinfo{person}{Xiaomo Liu}, {and} \bibinfo{person}{Sameena Shah}.} \bibinfo{year}{2023}\natexlab{c}.
\newblock \showarticletitle{Are ChatGPT and GPT-4 General-Purpose Solvers for Financial Text Analytics? An Examination on Several Typical Tasks}.
\newblock \bibinfo{journal}{\emph{arXiv preprint arXiv:2305.05862}} (\bibinfo{year}{2023}).
\newblock


\bibitem[Li et~al\mbox{.}(2023b)]%
        {li2023large}
\bibfield{author}{\bibinfo{person}{Yinheng Li}, \bibinfo{person}{Shaofei Wang}, \bibinfo{person}{Han Ding}, {and} \bibinfo{person}{Hang Chen}.} \bibinfo{year}{2023}\natexlab{b}.
\newblock \bibinfo{title}{Large Language Models in Finance: A Survey}.
\newblock
\newblock
\showeprint[arxiv]{2311.10723}~[q-fin.GN]


\bibitem[Lin(2004)]%
        {lin2004rouge}
\bibfield{author}{\bibinfo{person}{Chin-Yew Lin}.} \bibinfo{year}{2004}\natexlab{}.
\newblock \showarticletitle{Rouge: A package for automatic evaluation of summaries}. In \bibinfo{booktitle}{\emph{Text summarization branches out}}. \bibinfo{pages}{74--81}.
\newblock


\bibitem[Liu et~al\mbox{.}(2023a)]%
        {liu2023fingpt}
\bibfield{author}{\bibinfo{person}{Xiao-Yang Liu}, \bibinfo{person}{Guoxuan Wang}, {and} \bibinfo{person}{Daochen Zha}.} \bibinfo{year}{2023}\natexlab{a}.
\newblock \showarticletitle{Fingpt: Democratizing internet-scale data for financial large language models}.
\newblock \bibinfo{journal}{\emph{arXiv preprint arXiv:2307.10485}} (\bibinfo{year}{2023}).
\newblock


\bibitem[Liu et~al\mbox{.}(2022)]%
        {liu2022finrlmeta}
\bibfield{author}{\bibinfo{person}{Xiao-Yang Liu}, \bibinfo{person}{Ziyi Xia}, \bibinfo{person}{Jingyang Rui}, \bibinfo{person}{Jiechao Gao}, \bibinfo{person}{Hongyang Yang}, \bibinfo{person}{Ming Zhu}, \bibinfo{person}{Christina~Dan Wang}, \bibinfo{person}{Zhaoran Wang}, {and} \bibinfo{person}{Jian Guo}.} \bibinfo{year}{2022}\natexlab{}.
\newblock \bibinfo{title}{FinRL-Meta: Market Environments and Benchmarks for Data-Driven Financial Reinforcement Learning}.
\newblock
\newblock
\showeprint[arxiv]{2211.03107}~[q-fin.TR]


\bibitem[Liu et~al\mbox{.}(2023b)]%
        {liu2023dynamic}
\bibfield{author}{\bibinfo{person}{Xiao-Yang Liu}, \bibinfo{person}{Ziyi Xia}, \bibinfo{person}{Hongyang Yang}, \bibinfo{person}{Jiechao Gao}, \bibinfo{person}{Daochen Zha}, \bibinfo{person}{Ming Zhu}, \bibinfo{person}{Christina~Dan Wang}, \bibinfo{person}{Zhaoran Wang}, {and} \bibinfo{person}{Jian Guo}.} \bibinfo{year}{2023}\natexlab{b}.
\newblock \bibinfo{title}{Dynamic Datasets and Market Environments for Financial Reinforcement Learning}.
\newblock
\newblock
\showeprint[arxiv]{2304.13174}~[cs.LG]


\bibitem[Liu et~al\mbox{.}(2020)]%
        {ijcai2020p622}
\bibfield{author}{\bibinfo{person}{Zhuang Liu}, \bibinfo{person}{Degen Huang}, \bibinfo{person}{Kaiyu Huang}, \bibinfo{person}{Zhuang Li}, {and} \bibinfo{person}{Jun Zhao}.} \bibinfo{year}{2020}\natexlab{}.
\newblock \showarticletitle{FinBERT: A Pre-trained Financial Language Representation Model for Financial Text Mining}. In \bibinfo{booktitle}{\emph{Proceedings of the Twenty-Ninth International Joint Conference on Artificial Intelligence, {IJCAI-20}}}, \bibfield{editor}{\bibinfo{person}{Christian Bessiere}} (Ed.). \bibinfo{publisher}{International Joint Conferences on Artificial Intelligence Organization}, \bibinfo{pages}{4513--4519}.
\newblock
\urldef\tempurl%
\url{https://doi.org/10.24963/ijcai.2020/622}
\showDOI{\tempurl}
\newblock
\shownote{Special Track on AI in FinTech}.


\bibitem[Lopez-Lira and Tang(2023)]%
        {lopez2023can}
\bibfield{author}{\bibinfo{person}{Alejandro Lopez-Lira} {and} \bibinfo{person}{Yuehua Tang}.} \bibinfo{year}{2023}\natexlab{}.
\newblock \showarticletitle{Can chatgpt forecast stock price movements? return predictability and large language models}.
\newblock \bibinfo{journal}{\emph{arXiv preprint arXiv:2304.07619}} (\bibinfo{year}{2023}).
\newblock


\bibitem[Loughran and McDonald(2020)]%
        {loughran2020textual}
\bibfield{author}{\bibinfo{person}{Tim Loughran} {and} \bibinfo{person}{Bill McDonald}.} \bibinfo{year}{2020}\natexlab{}.
\newblock \showarticletitle{Textual analysis in finance}.
\newblock \bibinfo{journal}{\emph{Annual Review of Financial Economics}}  \bibinfo{volume}{12} (\bibinfo{year}{2020}), \bibinfo{pages}{357--375}.
\newblock


\bibitem[Magdon-Ismail and Atiya(2004)]%
        {magdon2004maximum}
\bibfield{author}{\bibinfo{person}{Malik Magdon-Ismail} {and} \bibinfo{person}{Amir~F Atiya}.} \bibinfo{year}{2004}\natexlab{}.
\newblock \showarticletitle{Maximum drawdown}.
\newblock \bibinfo{journal}{\emph{Risk Magazine}} \bibinfo{volume}{17}, \bibinfo{number}{10} (\bibinfo{year}{2004}), \bibinfo{pages}{99--102}.
\newblock


\bibitem[Maia et~al\mbox{.}(2018)]%
        {maia2018www}
\bibfield{author}{\bibinfo{person}{Macedo Maia}, \bibinfo{person}{Siegfried Handschuh}, \bibinfo{person}{Andre Freitas}, \bibinfo{person}{Brian Davis}, \bibinfo{person}{Ross McDermott}, \bibinfo{person}{Manel Zarrouk}, {and} \bibinfo{person}{Alexandra Balahur}.} \bibinfo{year}{2018}\natexlab{}.
\newblock \showarticletitle{WWW'18 Open Challenge: Financial Opinion Mining and Question Answering}.
\newblock \bibinfo{journal}{\emph{WWW '18: Companion Proceedings of the The Web Conference 2018}}, \bibinfo{pages}{1941--1942}.
\newblock
\showISBNx{9781450356404}
\urldef\tempurl%
\url{https://doi.org/10.1145/3184558.3192301}
\showDOI{\tempurl}


\bibitem[Malo et~al\mbox{.}(2014)]%
        {malo2014good}
\bibfield{author}{\bibinfo{person}{Pekka Malo}, \bibinfo{person}{Ankur Sinha}, \bibinfo{person}{Pekka Korhonen}, \bibinfo{person}{Jyrki Wallenius}, {and} \bibinfo{person}{Pyry Takala}.} \bibinfo{year}{2014}\natexlab{}.
\newblock \showarticletitle{Good debt or bad debt: Detecting semantic orientations in economic texts}.
\newblock \bibinfo{journal}{\emph{Journal of the Association for Information Science and Technology}} \bibinfo{volume}{65}, \bibinfo{number}{4} (\bibinfo{year}{2014}), \bibinfo{pages}{782--796}.
\newblock


\bibitem[Mariko et~al\mbox{.}(2020)]%
        {mariko2020financial}
\bibfield{author}{\bibinfo{person}{Dominique Mariko}, \bibinfo{person}{Hanna~Abi Akl}, \bibinfo{person}{Estelle Labidurie}, \bibinfo{person}{Stephane Durfort}, \bibinfo{person}{Hugues De~Mazancourt}, {and} \bibinfo{person}{Mahmoud El-Haj}.} \bibinfo{year}{2020}\natexlab{}.
\newblock \showarticletitle{Financial document causality detection shared task (fincausal 2020)}.
\newblock \bibinfo{journal}{\emph{arXiv preprint arXiv:2012.02505}} (\bibinfo{year}{2020}).
\newblock


\bibitem[Mukherjee et~al\mbox{.}(2022)]%
        {mukherjee2022ectsum}
\bibfield{author}{\bibinfo{person}{Rajdeep Mukherjee}, \bibinfo{person}{Abhinav Bohra}, \bibinfo{person}{Akash Banerjee}, \bibinfo{person}{Soumya Sharma}, \bibinfo{person}{Manjunath Hegde}, \bibinfo{person}{Afreen Shaikh}, \bibinfo{person}{Shivani Shrivastava}, \bibinfo{person}{Koustuv Dasgupta}, \bibinfo{person}{Niloy Ganguly}, \bibinfo{person}{Saptarshi Ghosh}, {et~al\mbox{.}}} \bibinfo{year}{2022}\natexlab{}.
\newblock \showarticletitle{Ectsum: A new benchmark dataset for bullet point summarization of long earnings call transcripts}.
\newblock \bibinfo{journal}{\emph{arXiv preprint arXiv:2210.12467}} (\bibinfo{year}{2022}).
\newblock


\bibitem[OpenAI(2023a)]%
        {openai2023gpt4}
\bibfield{author}{\bibinfo{person}{OpenAI}.} \bibinfo{year}{2023}\natexlab{a}.
\newblock \bibinfo{title}{GPT-4 Technical Report}.
\newblock
\newblock
\showeprint[arxiv]{2303.08774}~[cs.CL]


\bibitem[OpenAI(2023b)]%
        {openai2023gpt}
\bibfield{author}{\bibinfo{person}{R OpenAI}.} \bibinfo{year}{2023}\natexlab{b}.
\newblock \showarticletitle{Gpt-4 technical report. arxiv 2303.08774}.
\newblock \bibinfo{journal}{\emph{View in Article}}  \bibinfo{volume}{2} (\bibinfo{year}{2023}), \bibinfo{pages}{13}.
\newblock


\bibitem[Paiva et~al\mbox{.}(2019)]%
        {paiva2019decision}
\bibfield{author}{\bibinfo{person}{Felipe~Dias Paiva}, \bibinfo{person}{Rodrigo Tom{\'a}s~Nogueira Cardoso}, \bibinfo{person}{Gustavo~Peixoto Hanaoka}, {and} \bibinfo{person}{Wendel~Moreira Duarte}.} \bibinfo{year}{2019}\natexlab{}.
\newblock \showarticletitle{Decision-making for financial trading: A fusion approach of machine learning and portfolio selection}.
\newblock \bibinfo{journal}{\emph{Expert Systems with Applications}}  \bibinfo{volume}{115} (\bibinfo{year}{2019}), \bibinfo{pages}{635--655}.
\newblock


\bibitem[Punt(2017)]%
        {punt2017strategic}
\bibfield{author}{\bibinfo{person}{Andr{\'e}~E Punt}.} \bibinfo{year}{2017}\natexlab{}.
\newblock \showarticletitle{Strategic management decision-making in a complex world: quantifying, understanding, and using trade-offs}.
\newblock \bibinfo{journal}{\emph{ICES Journal of Marine Science}} \bibinfo{volume}{74}, \bibinfo{number}{2} (\bibinfo{year}{2017}), \bibinfo{pages}{499--510}.
\newblock


\bibitem[Quinlan({[n.\,d.]})]%
        {misc_australian_credit_approval_143}
\bibfield{author}{\bibinfo{person}{Ross Quinlan}.} \bibinfo{year}{[n.\,d.]}\natexlab{}.
\newblock \bibinfo{title}{{Statlog (Australian Credit Approval)}}.
\newblock \bibinfo{howpublished}{UCI Machine Learning Repository}.
\newblock
\newblock
\shownote{{DOI}: https://doi.org/10.24432/C59012}.


\bibitem[Roziere et~al\mbox{.}(2023)]%
        {roziere2023code}
\bibfield{author}{\bibinfo{person}{Baptiste Roziere}, \bibinfo{person}{Jonas Gehring}, \bibinfo{person}{Fabian Gloeckle}, \bibinfo{person}{Sten Sootla}, \bibinfo{person}{Itai Gat}, \bibinfo{person}{Xiaoqing~Ellen Tan}, \bibinfo{person}{Yossi Adi}, \bibinfo{person}{Jingyu Liu}, \bibinfo{person}{Tal Remez}, \bibinfo{person}{J{\'e}r{\'e}my Rapin}, {et~al\mbox{.}}} \bibinfo{year}{2023}\natexlab{}.
\newblock \showarticletitle{Code llama: Open foundation models for code}.
\newblock \bibinfo{journal}{\emph{arXiv preprint arXiv:2308.12950}} (\bibinfo{year}{2023}).
\newblock


\bibitem[Salinas~Alvarado et~al\mbox{.}(2015)]%
        {salinas-alvarado-etal-2015-domain}
\bibfield{author}{\bibinfo{person}{Julio~Cesar Salinas~Alvarado}, \bibinfo{person}{Karin Verspoor}, {and} \bibinfo{person}{Timothy Baldwin}.} \bibinfo{year}{2015}\natexlab{}.
\newblock \showarticletitle{Domain Adaption of Named Entity Recognition to Support Credit Risk Assessment}. In \bibinfo{booktitle}{\emph{Proceedings of the Australasian Language Technology Association Workshop 2015}}, \bibfield{editor}{\bibinfo{person}{Ben Hachey} {and} \bibinfo{person}{Kellie Webster}} (Eds.). \bibinfo{address}{Parramatta, Australia}, \bibinfo{pages}{84--90}.
\newblock
\urldef\tempurl%
\url{https://aclanthology.org/U15-1010}
\showURL{%
\tempurl}


\bibitem[Shah et~al\mbox{.}(2023a)]%
        {shah2023trillion}
\bibfield{author}{\bibinfo{person}{Agam Shah}, \bibinfo{person}{Suvan Paturi}, {and} \bibinfo{person}{Sudheer Chava}.} \bibinfo{year}{2023}\natexlab{a}.
\newblock \showarticletitle{Trillion Dollar Words: A New Financial Dataset, Task {\&} Market Analysis}. In \bibinfo{booktitle}{\emph{Proceedings of the 61st Annual Meeting of the Association for Computational Linguistics (Volume 1: Long Papers)}}, \bibfield{editor}{\bibinfo{person}{Anna Rogers}, \bibinfo{person}{Jordan Boyd-Graber}, {and} \bibinfo{person}{Naoaki Okazaki}} (Eds.). \bibinfo{publisher}{Association for Computational Linguistics}, \bibinfo{address}{Toronto, Canada}, \bibinfo{pages}{6664--6679}.
\newblock
\urldef\tempurl%
\url{https://doi.org/10.18653/v1/2023.acl-long.368}
\showDOI{\tempurl}


\bibitem[Shah et~al\mbox{.}(2023b)]%
        {shah2023finer}
\bibfield{author}{\bibinfo{person}{Agam Shah}, \bibinfo{person}{Ruchit Vithani}, \bibinfo{person}{Abhinav Gullapalli}, {and} \bibinfo{person}{Sudheer Chava}.} \bibinfo{year}{2023}\natexlab{b}.
\newblock \showarticletitle{Finer: Financial named entity recognition dataset and weak-supervision model}.
\newblock \bibinfo{journal}{\emph{arXiv preprint arXiv:2302.11157}} (\bibinfo{year}{2023}).
\newblock


\bibitem[Shah et~al\mbox{.}(2022)]%
        {shah2022flue}
\bibfield{author}{\bibinfo{person}{Raj Shah}, \bibinfo{person}{Kunal Chawla}, \bibinfo{person}{Dheeraj Eidnani}, \bibinfo{person}{Agam Shah}, \bibinfo{person}{Wendi Du}, \bibinfo{person}{Sudheer Chava}, \bibinfo{person}{Natraj Raman}, \bibinfo{person}{Charese Smiley}, \bibinfo{person}{Jiaao Chen}, {and} \bibinfo{person}{Diyi Yang}.} \bibinfo{year}{2022}\natexlab{}.
\newblock \showarticletitle{When FLUE Meets FLANG: Benchmarks and Large Pretrained Language Model for Financial Domain}. In \bibinfo{booktitle}{\emph{Proceedings of the 2022 Conference on Empirical Methods in Natural Language Processing}}. \bibinfo{pages}{2322--2335}.
\newblock


\bibitem[Sharma et~al\mbox{.}(2023)]%
        {sharma2023financial}
\bibfield{author}{\bibinfo{person}{Soumya Sharma}, \bibinfo{person}{Subhendu Khatuya}, \bibinfo{person}{Manjunath Hegde}, \bibinfo{person}{Afreen Shaikh}, \bibinfo{person}{Koustuv Dasgupta}, \bibinfo{person}{Pawan Goyal}, {and} \bibinfo{person}{Niloy Ganguly}.} \bibinfo{year}{2023}\natexlab{}.
\newblock \showarticletitle{Financial Numeric Extreme Labelling: A dataset and benchmarking}. In \bibinfo{booktitle}{\emph{Findings of the Association for Computational Linguistics: ACL 2023}}. \bibinfo{pages}{3550--3561}.
\newblock


\bibitem[Sharma et~al\mbox{.}(2022)]%
        {sharma2022finred}
\bibfield{author}{\bibinfo{person}{Soumya Sharma}, \bibinfo{person}{Tapas Nayak}, \bibinfo{person}{Arusarka Bose}, \bibinfo{person}{Ajay~Kumar Meena}, \bibinfo{person}{Koustuv Dasgupta}, \bibinfo{person}{Niloy Ganguly}, {and} \bibinfo{person}{Pawan Goyal}.} \bibinfo{year}{2022}\natexlab{}.
\newblock \showarticletitle{FinRED: A dataset for relation extraction in financial domain}. In \bibinfo{booktitle}{\emph{Companion Proceedings of the Web Conference 2022}}. \bibinfo{pages}{595--597}.
\newblock


\bibitem[Sharpe(1998)]%
        {sharpe1998sharpe}
\bibfield{author}{\bibinfo{person}{William~F Sharpe}.} \bibinfo{year}{1998}\natexlab{}.
\newblock \showarticletitle{The sharpe ratio}.
\newblock \bibinfo{journal}{\emph{Streetwise--the Best of the Journal of Portfolio Management}}  \bibinfo{volume}{3} (\bibinfo{year}{1998}), \bibinfo{pages}{169--85}.
\newblock


\bibitem[Sinha and Khandait(2020)]%
        {sinha2020impact}
\bibfield{author}{\bibinfo{person}{Ankur Sinha} {and} \bibinfo{person}{Tanmay Khandait}.} \bibinfo{year}{2020}\natexlab{}.
\newblock \bibinfo{title}{Impact of News on the Commodity Market: Dataset and Results}.
\newblock
\newblock
\showeprint[arxiv]{2009.04202}~[cs.CL]


\bibitem[Sinha and Khandait(2021)]%
        {sinha2021impact}
\bibfield{author}{\bibinfo{person}{Ankur Sinha} {and} \bibinfo{person}{Tanmay Khandait}.} \bibinfo{year}{2021}\natexlab{}.
\newblock \showarticletitle{Impact of news on the commodity market: Dataset and results}. In \bibinfo{booktitle}{\emph{Advances in Information and Communication: Proceedings of the 2021 Future of Information and Communication Conference (FICC), Volume 2}}. Springer, \bibinfo{pages}{589--601}.
\newblock


\bibitem[Soun et~al\mbox{.}(2022)]%
        {soun2022accurate}
\bibfield{author}{\bibinfo{person}{Yejun Soun}, \bibinfo{person}{Jaemin Yoo}, \bibinfo{person}{Minyong Cho}, \bibinfo{person}{Jihyeong Jeon}, {and} \bibinfo{person}{U Kang}.} \bibinfo{year}{2022}\natexlab{}.
\newblock \showarticletitle{Accurate Stock Movement Prediction with Self-supervised Learning from Sparse Noisy Tweets}. In \bibinfo{booktitle}{\emph{2022 IEEE International Conference on Big Data (Big Data)}}. IEEE, \bibinfo{pages}{1691--1700}.
\newblock


\bibitem[Sy et~al\mbox{.}(2023)]%
        {sy2023fine}
\bibfield{author}{\bibinfo{person}{Eugene Sy}, \bibinfo{person}{Tzu-Cheng Peng}, \bibinfo{person}{Shih-Hsuan Huang}, \bibinfo{person}{Heng-Yu Lin}, {and} \bibinfo{person}{Yung-Chun Chang}.} \bibinfo{year}{2023}\natexlab{}.
\newblock \showarticletitle{Fine-Grained Argument Understanding with BERT Ensemble Techniques: A Deep Dive into Financial Sentiment Analysis}. In \bibinfo{booktitle}{\emph{Proceedings of the 35th Conference on Computational Linguistics and Speech Processing (ROCLING 2023)}}. \bibinfo{pages}{242--249}.
\newblock


\bibitem[Team et~al\mbox{.}(2023)]%
        {team2023gemini}
\bibfield{author}{\bibinfo{person}{Gemini Team}, \bibinfo{person}{Rohan Anil}, \bibinfo{person}{Sebastian Borgeaud}, \bibinfo{person}{Yonghui Wu}, \bibinfo{person}{Jean-Baptiste Alayrac}, \bibinfo{person}{Jiahui Yu}, \bibinfo{person}{Radu Soricut}, \bibinfo{person}{Johan Schalkwyk}, \bibinfo{person}{Andrew~M Dai}, \bibinfo{person}{Anja Hauth}, {et~al\mbox{.}}} \bibinfo{year}{2023}\natexlab{}.
\newblock \showarticletitle{Gemini: a family of highly capable multimodal models}.
\newblock \bibinfo{journal}{\emph{arXiv preprint arXiv:2312.11805}} (\bibinfo{year}{2023}).
\newblock


\bibitem[Team(2023)]%
        {team2023internlm}
\bibfield{author}{\bibinfo{person}{InternLM Team}.} \bibinfo{year}{2023}\natexlab{}.
\newblock \bibinfo{title}{Internlm: A multilingual language model with progressively enhanced capabilities}.
\newblock
\newblock


\bibitem[Touvron et~al\mbox{.}(2023a)]%
        {touvron2023llama1}
\bibfield{author}{\bibinfo{person}{Hugo Touvron}, \bibinfo{person}{Thibaut Lavril}, \bibinfo{person}{Gautier Izacard}, \bibinfo{person}{Xavier Martinet}, \bibinfo{person}{Marie-Anne Lachaux}, \bibinfo{person}{Timoth{\'e}e Lacroix}, \bibinfo{person}{Baptiste Rozi{\`e}re}, \bibinfo{person}{Naman Goyal}, \bibinfo{person}{Eric Hambro}, \bibinfo{person}{Faisal Azhar}, {et~al\mbox{.}}} \bibinfo{year}{2023}\natexlab{a}.
\newblock \showarticletitle{Llama: Open and efficient foundation language models}.
\newblock \bibinfo{journal}{\emph{arXiv preprint arXiv:2302.13971}} (\bibinfo{year}{2023}).
\newblock


\bibitem[Touvron et~al\mbox{.}(2023b)]%
        {touvron2023llama}
\bibfield{author}{\bibinfo{person}{Hugo Touvron}, \bibinfo{person}{Louis Martin}, \bibinfo{person}{Kevin Stone}, \bibinfo{person}{Peter Albert}, \bibinfo{person}{Amjad Almahairi}, \bibinfo{person}{Yasmine Babaei}, \bibinfo{person}{Nikolay Bashlykov}, \bibinfo{person}{Soumya Batra}, \bibinfo{person}{Prajjwal Bhargava}, \bibinfo{person}{Shruti Bhosale}, {et~al\mbox{.}}} \bibinfo{year}{2023}\natexlab{b}.
\newblock \showarticletitle{Llama 2: Open foundation and fine-tuned chat models}.
\newblock \bibinfo{journal}{\emph{arXiv preprint arXiv:2307.09288}} (\bibinfo{year}{2023}).
\newblock


\bibitem[Wang et~al\mbox{.}(2023)]%
        {wang2023fingpt}
\bibfield{author}{\bibinfo{person}{Neng Wang}, \bibinfo{person}{Hongyang Yang}, {and} \bibinfo{person}{Christina~Dan Wang}.} \bibinfo{year}{2023}\natexlab{}.
\newblock \bibinfo{title}{FinGPT: Instruction Tuning Benchmark for Open-Source Large Language Models in Financial Datasets}.
\newblock
\newblock
\showeprint[arxiv]{2310.04793}~[cs.CL]


\bibitem[Wu et~al\mbox{.}(2018)]%
        {wu2018hybrid}
\bibfield{author}{\bibinfo{person}{Huizhe Wu}, \bibinfo{person}{Wei Zhang}, \bibinfo{person}{Weiwei Shen}, {and} \bibinfo{person}{Jun Wang}.} \bibinfo{year}{2018}\natexlab{}.
\newblock \showarticletitle{Hybrid deep sequential modeling for social text-driven stock prediction}. In \bibinfo{booktitle}{\emph{Proceedings of the 27th ACM international conference on information and knowledge management}}. \bibinfo{pages}{1627--1630}.
\newblock


\bibitem[Wu et~al\mbox{.}(2023)]%
        {wu2023bloomberggpt}
\bibfield{author}{\bibinfo{person}{Shijie Wu}, \bibinfo{person}{Ozan Irsoy}, \bibinfo{person}{Steven Lu}, \bibinfo{person}{Vadim Dabravolski}, \bibinfo{person}{Mark Dredze}, \bibinfo{person}{Sebastian Gehrmann}, \bibinfo{person}{Prabhanjan Kambadur}, \bibinfo{person}{David Rosenberg}, {and} \bibinfo{person}{Gideon Mann}.} \bibinfo{year}{2023}\natexlab{}.
\newblock \bibinfo{title}{BloombergGPT: A Large Language Model for Finance}.
\newblock
\newblock
\showeprint[arxiv]{2303.17564}~[cs.LG]


\bibitem[Xie et~al\mbox{.}(2023a)]%
        {xie2023wall}
\bibfield{author}{\bibinfo{person}{Qianqian Xie}, \bibinfo{person}{Weiguang Han}, \bibinfo{person}{Yanzhao Lai}, \bibinfo{person}{Min Peng}, {and} \bibinfo{person}{Jimin Huang}.} \bibinfo{year}{2023}\natexlab{a}.
\newblock \showarticletitle{The Wall Street Neophyte: A Zero-Shot Analysis of ChatGPT Over MultiModal Stock Movement Prediction Challenges}.
\newblock \bibinfo{journal}{\emph{arXiv preprint arXiv:2304.05351}} (\bibinfo{year}{2023}).
\newblock


\bibitem[Xie et~al\mbox{.}(2023b)]%
        {xie2023pixiu}
\bibfield{author}{\bibinfo{person}{Qianqian Xie}, \bibinfo{person}{Weiguang Han}, \bibinfo{person}{Xiao Zhang}, \bibinfo{person}{Yanzhao Lai}, \bibinfo{person}{Min Peng}, \bibinfo{person}{Alejandro Lopez-Lira}, {and} \bibinfo{person}{Jimin Huang}.} \bibinfo{year}{2023}\natexlab{b}.
\newblock \showarticletitle{PIXIU: A Large Language Model, Instruction Data and Evaluation Benchmark for Finance}.
\newblock \bibinfo{journal}{\emph{arXiv preprint arXiv:2306.05443}} (\bibinfo{year}{2023}).
\newblock


\bibitem[Xu and Cohen(2018)]%
        {xu2018stock}
\bibfield{author}{\bibinfo{person}{Yumo Xu} {and} \bibinfo{person}{Shay~B Cohen}.} \bibinfo{year}{2018}\natexlab{}.
\newblock \showarticletitle{Stock movement prediction from tweets and historical prices}. In \bibinfo{booktitle}{\emph{Proceedings of the 56th Annual Meeting of the Association for Computational Linguistics (Volume 1: Long Papers)}}. \bibinfo{pages}{1970--1979}.
\newblock


\bibitem[Yang et~al\mbox{.}(2023a)]%
        {yang2023fingpt}
\bibfield{author}{\bibinfo{person}{Hongyang Yang}, \bibinfo{person}{Xiao-Yang Liu}, {and} \bibinfo{person}{Christina~Dan Wang}.} \bibinfo{year}{2023}\natexlab{a}.
\newblock \bibinfo{title}{FinGPT: Open-Source Financial Large Language Models}.
\newblock
\newblock
\showeprint[arxiv]{2306.06031}~[q-fin.ST]


\bibitem[Yang et~al\mbox{.}(2020a)]%
        {yang2020generating}
\bibfield{author}{\bibinfo{person}{Linyi Yang}, \bibinfo{person}{Eoin~M Kenny}, \bibinfo{person}{Tin Lok~James Ng}, \bibinfo{person}{Yi Yang}, \bibinfo{person}{Barry Smyth}, {and} \bibinfo{person}{Ruihai Dong}.} \bibinfo{year}{2020}\natexlab{a}.
\newblock \showarticletitle{Generating plausible counterfactual explanations for deep transformers in financial text classification}.
\newblock \bibinfo{journal}{\emph{arXiv preprint arXiv:2010.12512}} (\bibinfo{year}{2020}).
\newblock


\bibitem[Yang et~al\mbox{.}(2023b)]%
        {yang2023investlm}
\bibfield{author}{\bibinfo{person}{Yi Yang}, \bibinfo{person}{Yixuan Tang}, {and} \bibinfo{person}{Kar~Yan Tam}.} \bibinfo{year}{2023}\natexlab{b}.
\newblock \bibinfo{title}{InvestLM: A Large Language Model for Investment using Financial Domain Instruction Tuning}.
\newblock
\newblock
\showeprint[arxiv]{2309.13064}~[q-fin.GN]


\bibitem[Yang et~al\mbox{.}(2020b)]%
        {yang2020finbert}
\bibfield{author}{\bibinfo{person}{Yi Yang}, \bibinfo{person}{Mark Christopher~Siy UY}, {and} \bibinfo{person}{Allen Huang}.} \bibinfo{year}{2020}\natexlab{b}.
\newblock \bibinfo{title}{FinBERT: A Pretrained Language Model for Financial Communications}.
\newblock
\newblock
\showeprint[arxiv]{2006.08097}~[cs.CL]


\bibitem[Yu et~al\mbox{.}(2023)]%
        {yu2023finmem}
\bibfield{author}{\bibinfo{person}{Yangyang Yu}, \bibinfo{person}{Haohang Li}, \bibinfo{person}{Zhi Chen}, \bibinfo{person}{Yuechen Jiang}, \bibinfo{person}{Yang Li}, \bibinfo{person}{Denghui Zhang}, \bibinfo{person}{Rong Liu}, \bibinfo{person}{Jordan~W. Suchow}, {and} \bibinfo{person}{Khaldoun Khashanah}.} \bibinfo{year}{2023}\natexlab{}.
\newblock \bibinfo{title}{FinMem: A Performance-Enhanced LLM Trading Agent with Layered Memory and Character Design}.
\newblock
\newblock
\showeprint[arxiv]{2311.13743}~[q-fin.CP]


\bibitem[Yuan et~al\mbox{.}(2021)]%
        {yuan2021bartscore}
\bibfield{author}{\bibinfo{person}{Weizhe Yuan}, \bibinfo{person}{Graham Neubig}, {and} \bibinfo{person}{Pengfei Liu}.} \bibinfo{year}{2021}\natexlab{}.
\newblock \showarticletitle{Bartscore: Evaluating generated text as text generation}.
\newblock \bibinfo{journal}{\emph{Advances in Neural Information Processing Systems}}  \bibinfo{volume}{34} (\bibinfo{year}{2021}), \bibinfo{pages}{27263--27277}.
\newblock


\bibitem[Zhang et~al\mbox{.}(2023a)]%
        {zhang2023instruction}
\bibfield{author}{\bibinfo{person}{Shengyu Zhang}, \bibinfo{person}{Linfeng Dong}, \bibinfo{person}{Xiaoya Li}, \bibinfo{person}{Sen Zhang}, \bibinfo{person}{Xiaofei Sun}, \bibinfo{person}{Shuhe Wang}, \bibinfo{person}{Jiwei Li}, \bibinfo{person}{Runyi Hu}, \bibinfo{person}{Tianwei Zhang}, \bibinfo{person}{Fei Wu}, {and} \bibinfo{person}{Guoyin Wang}.} \bibinfo{year}{2023}\natexlab{a}.
\newblock \bibinfo{title}{Instruction Tuning for Large Language Models: A Survey}.
\newblock
\newblock
\showeprint[arxiv]{2308.10792}~[cs.CL]


\bibitem[Zhang et~al\mbox{.}(2019)]%
        {zhang2019bertscore}
\bibfield{author}{\bibinfo{person}{Tianyi Zhang}, \bibinfo{person}{Varsha Kishore}, \bibinfo{person}{Felix Wu}, \bibinfo{person}{Kilian~Q Weinberger}, {and} \bibinfo{person}{Yoav Artzi}.} \bibinfo{year}{2019}\natexlab{}.
\newblock \showarticletitle{Bertscore: Evaluating text generation with bert}.
\newblock \bibinfo{journal}{\emph{arXiv preprint arXiv:1904.09675}} (\bibinfo{year}{2019}).
\newblock


\bibitem[Zhang et~al\mbox{.}(2023b)]%
        {zhang2023cgce}
\bibfield{author}{\bibinfo{person}{Xuanyu Zhang}, \bibinfo{person}{Bingbing Li}, {and} \bibinfo{person}{Qing Yang}.} \bibinfo{year}{2023}\natexlab{b}.
\newblock \bibinfo{title}{CGCE: A Chinese Generative Chat Evaluation Benchmark for General and Financial Domains}.
\newblock
\newblock
\showeprint[arxiv]{2305.14471}~[cs.CL]


\bibitem[Zhang et~al\mbox{.}(2024)]%
        {zhang2024dolares}
\bibfield{author}{\bibinfo{person}{Xiao Zhang}, \bibinfo{person}{Ruoyu Xiang}, \bibinfo{person}{Chenhan Yuan}, \bibinfo{person}{Duanyu Feng}, \bibinfo{person}{Weiguang Han}, \bibinfo{person}{Alejandro Lopez-Lira}, \bibinfo{person}{Xiao-Yang Liu}, \bibinfo{person}{Sophia Ananiadou}, \bibinfo{person}{Min Peng}, \bibinfo{person}{Jimin Huang}, {and} \bibinfo{person}{Qianqian Xie}.} \bibinfo{year}{2024}\natexlab{}.
\newblock \bibinfo{title}{D\'olares or Dollars? Unraveling the Bilingual Prowess of Financial LLMs Between Spanish and English}.
\newblock
\newblock
\showeprint[arxiv]{2402.07405}~[cs.CL]


\bibitem[Zhang et~al\mbox{.}(2023c)]%
        {zhang2023xuanyuan}
\bibfield{author}{\bibinfo{person}{Xuanyu Zhang}, \bibinfo{person}{Qing Yang}, {and} \bibinfo{person}{Dongliang Xu}.} \bibinfo{year}{2023}\natexlab{c}.
\newblock \bibinfo{title}{XuanYuan 2.0: A Large Chinese Financial Chat Model with Hundreds of Billions Parameters}.
\newblock
\newblock
\showeprint[arxiv]{2305.12002}~[cs.CL]


\bibitem[Zhao et~al\mbox{.}(2024)]%
        {zhao2024revolutionizing}
\bibfield{author}{\bibinfo{person}{Huaqin Zhao}, \bibinfo{person}{Zhengliang Liu}, \bibinfo{person}{Zihao Wu}, \bibinfo{person}{Yiwei Li}, \bibinfo{person}{Tianze Yang}, \bibinfo{person}{Peng Shu}, \bibinfo{person}{Shaochen Xu}, \bibinfo{person}{Haixing Dai}, \bibinfo{person}{Lin Zhao}, \bibinfo{person}{Gengchen Mai}, {et~al\mbox{.}}} \bibinfo{year}{2024}\natexlab{}.
\newblock \showarticletitle{Revolutionizing finance with llms: An overview of applications and insights}.
\newblock \bibinfo{journal}{\emph{arXiv preprint arXiv:2401.11641}} (\bibinfo{year}{2024}).
\newblock


\bibitem[Zhou et~al\mbox{.}(2023)]%
        {zhou2023forecasting}
\bibfield{author}{\bibinfo{person}{Xianzheng Zhou}, \bibinfo{person}{Hui Zhou}, {and} \bibinfo{person}{Huaigang Long}.} \bibinfo{year}{2023}\natexlab{}.
\newblock \showarticletitle{Forecasting the equity premium: Do deep neural network models work?}
\newblock \bibinfo{journal}{\emph{Modern Finance}} \bibinfo{volume}{1}, \bibinfo{number}{1} (\bibinfo{year}{2023}), \bibinfo{pages}{1--11}.
\newblock


\bibitem[Zhou et~al\mbox{.}(2021)]%
        {zhou2021trade}
\bibfield{author}{\bibinfo{person}{Zhihan Zhou}, \bibinfo{person}{Liqian Ma}, {and} \bibinfo{person}{Han Liu}.} \bibinfo{year}{2021}\natexlab{}.
\newblock \bibinfo{title}{Trade the Event: Corporate Events Detection for News-Based Event-Driven Trading}.
\newblock
\newblock
\showeprint[arxiv]{2105.12825}~[cs.CL]


\bibitem[Zhu et~al\mbox{.}(2021)]%
        {zhu2021tat}
\bibfield{author}{\bibinfo{person}{Fengbin Zhu}, \bibinfo{person}{Wenqiang Lei}, \bibinfo{person}{Youcheng Huang}, \bibinfo{person}{Chao Wang}, \bibinfo{person}{Shuo Zhang}, \bibinfo{person}{Jiancheng Lv}, \bibinfo{person}{Fuli Feng}, {and} \bibinfo{person}{Tat-Seng Chua}.} \bibinfo{year}{2021}\natexlab{}.
\newblock \showarticletitle{TAT-QA: A question answering benchmark on a hybrid of tabular and textual content in finance}.
\newblock \bibinfo{journal}{\emph{arXiv preprint arXiv:2105.07624}} (\bibinfo{year}{2021}).
\newblock


\end{thebibliography}

\section*{Checklist}

\begin{enumerate}

\item For all authors...
\begin{enumerate}
  \item Do the main claims made in the abstract and introduction accurately reflect the paper's contributions and scope?
    \answerYes
  \item Did you describe the limitations of your work?
    \answerYes
    See Limitation (Section \ref{sec:limitation}).
  \item Did you discuss any potential negative societal impacts of your work?
    \answerYes
    See Ethical Statement (Section \ref{sec:ethical}).
  \item Have you read the ethics review guidelines and ensured that your paper conforms to them?
    \answerYes
    See Ethical Statement (Section \ref{sec:ethical}).
\end{enumerate}

\item If you are including theoretical results...
\begin{enumerate}
  \item Did you state the full set of assumptions of all theoretical results?
    \answerNA
	\item Did you include complete proofs of all theoretical results?
    \answerNA
\end{enumerate}

\item If you ran experiments (e.g. for benchmarks)...
\begin{enumerate}
  \item Did you include the code, data, and instructions needed to reproduce the main experimental results (either in the supplemental material or as a URL)?
    \answerYes
    See Introduction (Section \ref{section:introduction}).
  \item Did you specify all the training details (e.g., data splits, hyperparameters, how they were chosen)?
    \answerNA
    Our benchmark only includes the evaluation process.
	\item Did you report error bars (e.g., with respect to the random seed after running experiments multiple times)?
    \answerYes
    See Table \ref{table:performance_overview}.
	\item Did you include the total amount of compute and the type of resources used (e.g., type of GPUs, internal cluster, or cloud provider)?
    \answerYes
    See Experimental Settings (Section \ref{sec:experimental_setting}).
\end{enumerate}

\item If you are using existing assets (e.g., code, data, models) or curating/releasing new assets...
\begin{enumerate}
  \item If your work uses existing assets, did you cite the creators?
    \answerYes
  \item Did you mention the license of the assets?
    \answerYes
  \item Did you include any new assets either in the supplemental material or as a URL?
    \answerYes
    Our Introduction (Section \ref{section:introduction}) contains a link for all data used in FinBen.
  \item Did you discuss whether and how consent was obtained from people whose data you're using/curating?
    \answerYes
    Our dataset statistics (Table \ref{tab:eval}) contains licenses for all used datasets.
  \item Did you discuss whether the data you are using/curating contains personally identifiable information or offensive content?
    \answerYes
    See Ethical Statement (Section \ref{sec:ethical}).
\end{enumerate}

\item If you used crowdsourcing or conducted research with human subjects...
\begin{enumerate}
  \item Did you include the full text of instructions given to participants and screenshots, if applicable?
    \answerNA
  \item Did you describe any potential participant risks, with links to Institutional Review Board (IRB) approvals, if applicable?
    \answerNA
  \item Did you include the estimated hourly wage paid to participants and the total amount spent on participant compensation?
    \answerNA
\end{enumerate}

\end{enumerate}


\appendix

\section{Contributions}
\textbf{Science Leadership}: Qianqian Xie, Min Peng, Sophia Ananiadou, Alejandro Lopez-Lira, Hao Wang, Yanzhao Lai, Benyou Wang, Xiao-yang Liu, Gang Hu, Jiajia Huang, Jimin Huang.

\textbf{Contributors}: Mengxi Xiao, Dong Li, Weiguang Han, Zhengyu Chen, Ruoyu Xiang, Xiao Zhang, Yueru He,  Yongfu Dai, Duanyu Feng, Yijing Xu, Haoqiang Kang, Ziyan Kuang, Chenhan Yuan, Kailai Yang, Zheheng Luo, Tianlin Zhang, Zhiwei Liu, Guojun Xiong, Zhiyang Deng, Yuechen Jiang, Zhiyuan Yao, Haohang Li, Yangyang Yu

\section{Other LLMs Performance}
Table \ref{tab:additional-per} presents other LLMs' performance in the FinBen.

\begin{table}[htb!]
		\centering
		\scriptsize
  \caption{The zero-shot and few-shots performance of other LLMs on the FinBen.}
		\begin{tabular}{llcccccccccc}
			\toprule
			\textbf{Dataset} 
			&\textbf{Metrics}
			&\makecell{\textbf{Baichuan}\\\textbf{7B}}
			&\makecell{\textbf{CodeLLaMA}\\\textbf{7B}}
			&\makecell{\textbf{DISC-}\\\textbf{FinLLM}}
                &\makecell{\textbf{ChatGLM3}\\\textbf{6B}}

			\\\midrule
			\textcolor{black}{NER}
			&EntityF1&0.00&{0.07}&{0.12}&\textbf{0.25}\\

            \textcolor{black}{FINER-ORD}
			&EntityF1&{0.00}&\textcolor{black}{{0.00}}&\textcolor{black}{0.00}&\textbf{0.02}\\

            \textcolor{black}{FinRED}
			&F1&\textbf{0.00}&\textcolor{black}{{0.00}}&\textcolor{black}{0.00}&0.00\\

            \textcolor{black}{SC}
			&F1&\textcolor{black}{0.74}&\textbf{{0.85}}&\textcolor{black}{0.00}&0.81\\

            \textcolor{black}{CD}
			&F1&\textbf{0.00}&\textcolor{black}{{0.00}}&\textcolor{black}{0.00}&0.00\\

\textcolor{black}{FNXL}
			&EntityF1&\textbf{0.00}&\textcolor{black}{{0.00}}&\textcolor{black}{0.00}&0.00\\

            \textcolor{black}{FSRL}
			&EntityF1&\textbf{0.00}&\textcolor{black}{{0.00}}&\textcolor{black}{0.00}&0.00\\

   \midrule
			\multirow{2}{*}{FPB}
	&F1&{0.17}&{0.34}\textcolor{black}&\textcolor{black}{0.29}&\textbf{0.74}\\
			&Acc&0.23\textcolor{black}&{0.39}\textcolor{black}&\textcolor{black}{0.26}&\textbf{0.74}\\
   
		\multirow{1}{*}{FiQA-SA}
			&F1&0.32&\textbf{0.66}&\textcolor{black}{0.32}&0.56\\

            \multirow{1}{*}{TSA}
            &RMSE$\downarrow$&0.44&0.43&\textbf{0.32}&0.35\\

            \textcolor{black}{Headlines}
			&AvgF1&{0.60}\textcolor{black}&0.60\textcolor{black}&\textcolor{black}{0.60}&\textbf{0.66}\\

            \multirow{2}{*}{FOMC}
			&F1&\textcolor{black}{0.17}&\textcolor{black}{{0.14}}&{0.19}&\textbf{0.47}\\
			&Acc&\textcolor{black}{0.25}&\textcolor{black}{{0.27}}&{0.28}&\textbf{0.46}\\

            \textcolor{black}{FinArg-ACC}
			&MicroF1&\textbf{0.36}&0.28\textcolor{black}{}&\textcolor{black}{0.29}&0.25\\

            \textcolor{black}{FinArg-ARC}
			&MicroF1&0.27&0.25\textcolor{black}{}&{0.29}&\textbf{0.50}\\

            \textcolor{black}{MultiFin}
			&MicroF1&0.12&0.21\textcolor{black}{}&{0.29}&\textbf{0.47}\\

            \textcolor{black}{M\&A}
			&MicroF1&0.33&{0.54}\textcolor{black}{}&\textcolor{black}{0.29}&\textbf{0.79}\\

            \textcolor{black}{MLESG}
			&MicroF1&0.04&0.10\textcolor{black}{}&\textbf{0.29}&0.16\\\midrule

			FinQA
			&EmAcc&\textbf{0.00}&{0.00}&\textcolor{black}{0.00}&0.00\\

                TATQA
			&EmAcc&{0.00}&{0.00}&\textcolor{black}{0.00}&\textbf{0.07}\\

            \multirow{2}{*}{Regulations}
			&Rouge-1&\textcolor{black}{0.13}&\textcolor{black}{-}&\textcolor{black}{-}&\textbf{0.26}\\
			&BertScore&\textcolor{black}{0.60}&\textcolor{black}{-}&\textcolor{black}{-}&\textbf{0.65}\\

			ConvFinQA
			&EmAcc&\textbf{0.00}&{0.00}&\textcolor{black}{0.00}&0.00\\

            \midrule

			\multirow{4}{*}{EDTSUM}
			&Rouge-1&{0.02}&\textcolor{black}{0.10}&\textbf{0.22}&0.13\\
			&BertScore&\textcolor{black}{0.47}&\textcolor{black}{\textbf{0.67}}&\textcolor{black}{0.61}&0.47\\
			&BartScore&\textcolor{black}{-6.18}&\textcolor{black}{\textbf{-3.62}}&\textcolor{black}{-4.13}&-4.78\\
			
			\multirow{4}{*}{ECTSUM}
			&Rouge-1&\textbf{0.00}&\textcolor{black}{0.00}&\textcolor{black}{0.00}&0.00\\
			&BertScore&\textbf{0.00}&\textcolor{black}{0.00}&\textcolor{black}{0.00}&0.00\\
			&BartScore&\textbf{-5.18}&\textcolor{black}{-5.18}&\textcolor{black}{-5.18}&-5.18\\\midrule

            \multirow{2}{*}{BigData22}
			&Acc&\textbf{0.53}&{0.52}&\textcolor{black}{0.44}&0.47\\
		&MCC&{-0.01}&-0.01&\textcolor{black}{-0.05}&\textbf{0.00}\\
			
			\multirow{2}{*}{ACL18}
			&Acc&0.50&\textbf{0.51}&\textcolor{black}{0.50}&0.50\\
			&MCC&-0.01&0.00&\textbf{0.02}&0.02\\
			
			\multirow{2}{*}{CIKM18}
			&Acc&0.48&\textbf{0.51}&\textcolor{black}{0.44}&0.42\\
			&MCC&\textbf{0.02}&0.02&\textcolor{black}{-0.03}&0.02\\

   \midrule
			\multirow{2}{*}{German}
			&F1&\textcolor{black}{0.52}&\textcolor{black}{\textbf{0.66}}&\textcolor{black}{0.52}&0.41\\
			&MCC&\textbf{0.00}&\textcolor{black}{0.00}&\textcolor{black}{{0.00}}&-0.30\\
			
			\multirow{2}{*}{Australian}
			&F1&\textcolor{black}{0.26}&\textcolor{black}{\textbf{0.43}}&\textcolor{black}{0.26}&0.27\\
			&MCC&\textbf{0.00}&\textcolor{black}{{0.00}}&\textcolor{black}{0.00}&-0.02\\

            \multirow{2}{*}{LendingClub}
			&F1&\textcolor{black}{0.59}&\textcolor{black}{\textbf{0.81}}&\textcolor{black}{0.72}&0.72\\
			&MCC&\textbf{0.02}&{0.00}&\textcolor{black}{{0.00}}&-0.03\\

            \multirow{2}{*}{ccf}
			&F1&{0.01}&\textcolor{black}{{0.00}}&{0.66}&\textbf{1.00}\\
			&MCC&\textbf{0.00}&\textcolor{black}{0.00}&\textcolor{black}{{-0.04}}&0.00\\

            \multirow{2}{*}{ccfraud}
			&F1&\textcolor{black}{0.05}&\textcolor{black}{{0.06}}&{0.46}&\textbf{0.67}\\
			&MCC&\textcolor{black}{-0.04}&\textcolor{black}{0.00}&\textcolor{black}{\textbf{0.02}}&-0.15\\

            \multirow{2}{*}{polish}
			&F1&\textbf{0.92}&\textcolor{black}{{0.47}}&{0.92}&0.05\\
			&MCC&\textcolor{black}{-0.01}&\textbf{0.04}&\textcolor{black}{{0.00}}&0.00\\

            \multirow{2}{*}{taiwan}
			&F1&\textbf{0.95}&\textcolor{black}{{0.36}}&{0.95}&0.05\\
			&MCC&{0.00}&\textcolor{black}{-0.03}&\textcolor{black}{0.00}&\textbf{0.01}\\

            \multirow{2}{*}{portoseguro}
			&F1&\textcolor{black}{0.00}&{{0.88}}&\textcolor{black}{0.63}&\textbf{0.95}\\
			&MCC&{0.00}&\textcolor{black}{-0.01}&\textcolor{black}{{-0.02}}&\textbf{0.06}\\

            \multirow{2}{*}{travelinsurance}
			&F1&{0.00}&{0.02}&\textcolor{black}{0.00}&\textbf{0.97}\\
			&MCC&{0.00}&{0.00}&\textcolor{black}{{0.00}}&\textbf{0.03}\\

			\bottomrule
		\end{tabular}
		\label{tab:additional-per}
	\end{table}    
\section{Instructions}
\label{app:instruction}
For detail instruction of each dataset, please see Table~\ref{tab:prompt} and Table~\ref{tab:times}.
\begin{table*}[htb!]
    \centering
    \scriptsize
  \caption{Quantification task datasets prompt overview.}
    \scalebox{0.85}{
    \begin{tabular}{ll}
        \toprule
        \textbf{Data} & \textbf{Prompt} \\
        \midrule
        FPB & \makecell[l]{``Analyze the sentiment of this statement extracted from a financial news article.\\Provide your answer as either negative, positive or neutral.\\For instance, 'The company's stocks plummeted following the scandal.' would be classified as negative."} \\
        \midrule
        FiQA-SA & \makecell[l]{``What is the sentiment of the following financial \textcolor{blue}{\{category\}}:\\Positive, Negative, or Neutral?"} \\
        \midrule
        \textcolor{black}{Headlines} & \makecell[l]{``Consider whether the headline mentions the price of gold.\\Is there a Price or Not in the gold commodity market indicated in the news headline?\\Please answer Yes or No."} \\
        \midrule

        \textcolor{black}{NER} & \makecell[l]{``In the sentences extracted from financial agreements in U.S. SEC filings,\\identify the named entities that represent a person ('PER'), an organization ('ORG'),\\or a location ('LOC'). The required answer format is: 'entity name, entity type'.\\For instance, in 'Elon Musk, CEO of SpaceX, announced the launch from Cape Canaveral.',\\the entities would be: 'Elon Musk, PER; SpaceX, ORG; Cape Canaveral, LOC'"} \\
        \midrule

        \textcolor{black}{FiNER-ORD} & \makecell[l]{``In the list of tokens, identify \textcolor{blue}{\{tid\}}each accordingly. \\ If the entity spans multiple tokens, use the prefix B-PER, B-LOC, or B-ORG for the first token, and I-PER,\\
        I-LOC, or I-ORG for the subsequent tokens of that entity. \\ The beginning of each separate entity should always be labeled with a B-PER, B-LOC, or B-ORG prefix.\\ If the token does not fit into any of the three named categories, or is not a named entity, label it as 'O'."} \\
        \midrule

        \textcolor{black}{FinQA} & \makecell[l]{``Given the financial data and expert analysis, please answer this question:"} \\
        \midrule

        Regulations & \makecell[l]{``Please answer following questions."} \\
        \midrule

        \textcolor{black}{ConvFinQA} & \makecell[l]{``In the context of this series of interconnected finance-related queries and the additional information \\provided by the pretext, table data, and post text from a company's financial filings,\\please provide a response to the final question. This may require extracting information\\ from the context and performing mathematical calculations. Please take into account the information provided in\\ the preceding questions and their answers when formulating your response:"} \\
        \midrule
        
        \textcolor{black}{BigData22} & \makecell[l]{\textcolor{black}{``
                Contemplate the data and tweets to guess whether the closing price of \textcolor{blue}{\{tid\}} will surge or decline at \textcolor{blue}{\{point\}}.}\\ \textcolor{black}{Please declare with either Rise or Fall."}} \\
        \midrule

        \textcolor{black}{ACL18} & \makecell[l]{``Scrutinize the data and tweets to envisage if the closing price of \textcolor{blue}{\{tid\}}will swell or contract at \textcolor{blue}{\{point\}}.\\Respond with either Rise or Fall."} \\
        \midrule

        \textcolor{black}{CIKM18} & \makecell[l]{``Reflect on the provided data and tweets to anticipate if the closing price of \textcolor{blue}{\{tid\}}is going to increase or decrease at \textcolor{blue}{\{point\}}.\\Respond with either Rise or Fall."} \\
        \midrule
        
        \textcolor{black}{ECTSum } & \makecell[l]{\textcolor{black}{``Given the following article, please produce a list of 0 and 1, each separated by ' ' to indicate which sentences}\\ 
            \textcolor{black}{should be included in the final summary. The article's sentences have been split by ' '. Please mark each sentence}\\ 
            \textcolor{black}{with 1 if it should be included in the summary and 0 if it should not."}}\\
        \midrule

        \textcolor{black}{EDTSum } & \makecell[l]{\textcolor{black}{``You are given a text that consists of multiple sentences. Your task is to perform abstractive summarization on this text. Use} \\
            \textcolor{black}{your understanding of the content to express the main ideas and crucial details in a shorter, coherent, and natural sounding text."}}\\
        \midrule
        
        \textcolor{black}{German} & \makecell[l]{\textcolor{black}{``Assess the creditworthiness of a customer using the following table attributes for financial status. Respond with either} \\
            \textcolor{black}{'good' or 'bad'. And the table attributes including 13 categorical attributes and 7 numerical attributes are as follows:"}}\\
        \midrule

        \textcolor{black}{Australian} & \makecell[l]{``Assess the creditworthiness of a customer using the following table attributes for financial status. Respond with either \\'good' or 'bad'. And the table attributes including 13 categorical attributes \\
        and 7 numerical attributes and values have been changed to meaningless symbols to protect confidentiality of the data. :"}\\
        \midrule

        \textcolor{black}{FOMC} & \makecell[l]{\textcolor{black}{``Examine the excerpt from a central bank's release below. Classify it as HAWKISH if it advocates for a tightening} \\of monetary policy, DOVISH if it suggests an easing of monetary policy, or NEUTRAL if the stance is unbiased.\\Your response should return only HAWKISH, DOVISH, or NEUTRAL."}\\
        \midrule

        \textcolor{black}{TSA} & \makecell[l]{``Given the following financial text, return a sentiment score for Ashtead as a floating-point number \\ranging from -1 (indicating a very negative or bearish sentiment) to 1 (indicating a very positive or bullish sentiment), \\with 0 designating neutral sentiment. Return only the numerical score first, \\follow it with a brief reasoning behind your score."} \\
        \midrule

        \textcolor{black}{FinArg - ACC} & \makecell[l]{``Analyze sentences from earnings conference calls and identify \\their argumentative function. \\Each sentence is either a premise, offering evidence or reasoning, or a claim, \\asserting a conclusion or viewpoint. Return only premise or claim."} \\
        \midrule

        \textcolor{black}{FinArg - ARC} & \makecell[l]{``In this task, you are given a pair of sentences. \\Your objective is to ascertain the type of argumentative relation between these two sentences. \\The relation could either be 'NoRelation', indicating no discernible relation between the sentences, \\'Support', indicating that the first sentence supports the second, or 'Attack', indicating that the first sentence disputes \\or contradicts the second. Return only one of the three classifications: 'norelation', 'support', or 'attack'."} \\
        \midrule

        \textcolor{black}{MultiFin} & \makecell[l]{``In this task, you're working with English headlines from the MULTIFIN dataset. \\This dataset is made up of real-world article headlines from a large accounting firm's websites. \\Your objective is to categorize each headline according to its primary topic. \\The potential categories are \textcolor{blue}{\{category\}}. \\Your response should only include the category that best fits the headline."} \\
        \midrule

        \textcolor{black}{MA} & \makecell[l]{``In this task, you will be given Mergers and Acquisitions news articles or tweets. \\Your task is to classify each article or tweet based on whether the mentioned deal was completed or remained a rumour. \\Your response should be a single word - either 'complete' or 'rumour' - \\representing the outcome of the deal mentioned in the provided text."} \\
        \midrule

        \textcolor{black}{MLESG} & \makecell[l]{``You're given English news articles related to Environmental, Social, and Corporate Governance (ESG) issues. \\Your task is to classify each article based on the ESG issue it pertains to, according to the MSCI ESG rating guidelines. \\The ESG issues include \textcolor{blue}{\{category\}}. \\Your output should be the most relevant ESG issue label, followed by a brief rationale based on the article content."} \\

                \bottomrule
		\end{tabular}}
		\label{tab:prompt}
	   \end{table*}

\begin{table*}[htb!]
    \centering
    \scriptsize
    \caption{The example prompts of remaining tasks. FiQA-SA has two types of text, including news headlines and tweets. We will fill the detailed text type into \textcolor{blue}{\{category\}} for each data sample. For stock movement prediction data such as BigData22, we will fill \textcolor{blue}{\{tid\}} and \textcolor{blue}{\{point\}} with the detailed stock name and time from each data sample.}
    \scalebox{0.8}{
    \begin{tabular}{ll}
        \toprule
        \textbf{Data} & \textbf{Prompt} \\
        \midrule
        \textcolor{black}{FinRED} & \makecell[l]{``Given the following sentence, identify the head, tail, and relation of each triplet present in the sentence. \\The relations you should be looking for are \textcolor{blue}{\{category\}}. \\If a relation exists between two entities, provide your answer in the format \textcolor{blue}{\{category\}}. \\If there are multiple triplets in a sentence, provide each one on a new line."} \\ \midrule
        \textcolor{black}{SC} & \makecell[l]{``In this task, you are provided with sentences extracted from financial news and SEC data. \\Your goal is to classify each sentence into either 'causal' or 'noise' based on whether or not it indicates a causal relationship between financial events. \\Please return only the category 'causal' or 'noise'."} \\
        \midrule

        \textcolor{black}{CD} & \makecell[l]{``Your job in this task is to perform sequence labeling on a provided text section, marking the chunks that represent the cause of an event and the effects\\ that result from it. For each token in the text, assign a label to indicate its role in representing cause or effect. \\The labels you should use are 'B-CAUSE', 'I-CAUSE', 'B-EFFECT', 'I-EFFECT', and 'O'.\\ A 'B-' prefix is used to denote the beginning of a cause or effect sequence, \\while an 'I-' prefix is used for continuation of a cause or effect sequence. \\If a token is not part of either a cause or effect sequence, label it as 'O'. \\Provide your answer as a sequence of 'token:label' pairs, with each pair on a new line."}\\
        \midrule

        \textcolor{black}{TATQA} & \makecell[l]{``Please answer the given financial question based on the context. Context: \textcolor{blue}{\{context\}}Question: What is the amount of total sales in 2019?"} \\
        \midrule

        \textcolor{black}{FNXL} & \makecell[l]{``In the task of Financial Numeric Extreme Labelling (FNXL), \\your job is to identify and label the semantic role of each token in a sentence. \\The labels can include \textcolor{blue}{\{category\}}"} \\
        \midrule

        \textcolor{black}{FSRL} & \makecell[l]{``In the task of Textual Analogy Parsing (TAP), your job is to identify and label the semantic role of each token in a sentence.\\ The labels can include \textcolor{blue}{\{category\}}."} \\
        \midrule

        \textcolor{black}{LendingClub} & \makecell[l]{``Assess the client's loan status based on the following loan records from Lending Club. \\Respond with only 'good' or 'bad', and do not provide any additional information. \\For instance, 'The client has a stable income, no previous debts, and owns a property.' should be classified as 'good'."} \\
        \midrule

        \textcolor{black}{ccf} & \makecell[l]{``Detect the credit card fraud using the following financial table attributes. \\Respond with only 'yes' or 'no', and do not provide any additional information. \\Therein, the data contains 28 numerical input variables V1, V2, ..., \\and V28 which are the result of a PCA transformation and 1 input variable Amount which has not been transformed with PCA. \\The feature 'Amount' is the transaction Amount, this feature can be used for example-dependant cost-sensitive learning. \\For instance, 'The client has attributes:\textcolor{blue}{\{category\}}"} \\
        \midrule

        \textcolor{black}{ccfraud} & \makecell[l]{``Detect the credit card fraud with the following financial profile. \\Respond with only 'good' or 'bad', and do not provide any additional information. For instance, \\'The client is a female, the state number is 25, the number of cards is 1, the credit balance is 7000, \\the number of transactions is 16, the number of international transactions is 0, \\the credit limit is 6.' should be classified as 'good'."} \\
        \midrule

        \textcolor{black}{polish} & \makecell[l]{``Predict whether the company will face bankruptcy based on the financial profile attributes provided in the following text. \\Respond with only 'no' or 'yes', and do not provide any additional information."} \\
        \midrule

        \textcolor{black}{taiwan} & \makecell[l]{``Predict whether the company will face bankruptcy based on the financial profile attributes provided in the following text. \\Respond with only 'no' or 'yes', and do not provide any additional information."} \\
        \midrule

        \textcolor{black}{Porto-Seguro} & \makecell[l]{``Identify whether or not to files a claim for the auto insurance policy holder using the following table attributes about individual financial profile. \\Respond with only 'yes' or 'no', and do not provide any additional information.\\ And the table attributes that belong to similar groupings are tagged as such in the feature names (e.g., ind, reg, car, calc). \\In addition, feature names include the postfix bin to indicate binary features and cat to indicate categorical features. \\Features without these designations are either continuous or ordinal. \\Values of -1 indicate that the feature was missing from the observation."} \\
        \midrule

        \textcolor{black}{travelinsurace} & \makecell[l]{``Identify the claim status of insurance companies using the following table attributes for travel insurance status. \\Respond with only 'yes' or 'no', and do not provide any additional information.\\ And the table attributes including 5 categorical attributes and 4 numerical attributes are as follows:\textcolor{blue}{\{category\}}"} \\
        \midrule

    \textcolor{black}{FinTrade} & \makecell[l]{``Given the information, can you make an investment decision? Just summarize the reason of the decision.\\
    please consider only the available short-term information, the mid-term information, the long-term information, the \\reflection-term information.\\
    please consider the momentum of the historical stock price.\\
    When cumulative return is positive or zero, you are a risk-seeking investor.\\
    But when cumulative return is negative, you are a risk-averse investor.\\ 
    please consider how much share of the stock the investor holds now.  \\ 
    You should provide exactly one of the following investment decisions: buy or sell.\\
    When it is really hard to make a 'buy'-or-'sell' decision, you could go with 'hold' option.\\
    You also need to provide the id of the information to support your decision.\\
    \textcolor{blue}{\{investment\_info\}}\\
    \textcolor{blue}{\{gr\.complete\_json\_suffix\_v2\}}\\
    Your output should strictly conforms the following json format without any additional contents:\\ \{"investment\_decision" : string, "summary\_reason": string, "short\_memory\_index": number,\\ "middle\_memory\_index": number, "long\_memory\_index": number, "reflection\_memory\_index": number\}"} \\
        
                \bottomrule
		\end{tabular}}
\label{tab:times}
        
	\end{table*}

\section{Related Work}
\subsection{Financial Large Language Models}
Recent years have seen a significant surge in research on finance-specific LLMs, expanding on the groundwork laid by general-purpose language models~\citep{lee2024survey,liu2023dynamic,xie2023wall,zhang2024dolares,dai2024laiw}.
 Financial pre-trained language models (FinPLMs) like FinBERT~\citep{araci2019finbert,yang2020finbert,ijcai2020p622}, derived from BERT, and FLANG~\citep{shah2022flue}, based on ELECTRA, have been developed using domain-specific data for enhanced performance in tasks like sentiment analysis and stock prediction. 
 The open-source release of Meta AI's LLaMA~\citep{touvron2023llama1,touvron2023llama} has fueled further innovation in Financial LLMs (FinLLMs), with models like FinMA~\citep{xie2023pixiu}, InvestLM~\citep{yang2023investlm}, and FinGPT~\citep{wang2023fingpt,liu2023fingpt} leveraging advanced tuning strategies~\citep{zhang2023instruction} for financial applications. BloombergGPT~\citep{wu2023bloomberggpt} stands out as a BLOOM-based, closed-source model tailored for the financial industry. Additionally, the Chinese financial sector has seen the emergence of models like XuanYuan 2.0~\citep{zhang2023xuanyuan}, integrating broad and specialized knowledge, FinBART~\citep{hongyuan-etal-2023-finbart} for financial communication, and CFGPT~\citep{li2023cfgpt}, which includes a comprehensive dataset for targeted pre-training and fine-tuning.

\subsection{Financial Evaluation Benchmarks}
Financial evaluation benchmarks, such as the pioneering FLUE~\citep{shah2022flue}, have been introduced to measure model performance in the financial sector, covering five key NLP tasks: financial sentiment analysis~\citep{shah2022flue}, news headline classification~\citep{sinha2020impact}, named entity recognition (NER)~\citep{salinas-alvarado-etal-2015-domain}, structure boundary detection and question answering (QA)~\citep{chen2022finqa}.
Building upon FLUE, FLARE~\citep{xie2023pixiu} added the evaluation of time-series processing capabilities, i.e., forecasting stock price movements. 
In addition, in Chinese financial benchmarks, there are more recently released Chinese datasets like CFBenchmark~\citep{lei2023cfbenchmark}, DISC-FINSFT~\citep{chen2023disc}, and CGCE~\citep{zhang2023cgce}.
However, these benchmarks have a limited scope and have not yet addressed more complex financial NLP tasks such as event detection~\citep{zhou2021trade}, and realistic financial tasks, despite the fact that there were previous efforts on stock trading~\citep{liu2022finrlmeta,han2023mastering,han2023select}.

\section{Trading Accumulative Returns}
\label{app:trading}
Table~\ref{tab:trading_all} and the below Figures show detailed trading performance, 
\begin{table*}[htb]
		\centering
		\scriptsize
\caption{The overall trading performance comparison for different LLMs across various stocks. The results include large LLMs only ($\geq 70B$), as models with smaller contexts have difficulty understanding the instructions and producing a static strategy of holding.}
\scalebox{0.8}{
\begin{tabular}{lllllll}
\toprule
\textbf{Ticker} & \textbf{Model} & \textbf{CR (\%)} & \textbf{SR} & \textbf{DV (\%)} & \textbf{AV (\%)} & \textbf{MD (\%)} \\
\midrule
\multirow{7}{*}{TSLA} 
    & Buy \& Hold & -25.2137 & -0.7203 & 4.4099 & 70.0043 & 57.6765 \\
    & GPT-4 & \textbf{68.3089} & \textbf{2.8899} & \textbf{2.9780} & \textbf{47.2739} & \textbf{10.7996} \\
    & GPT-4o & -0.8789 & -0.0321 & 3.4531 & 54.8156 & 44.6842 \\
    & GPT3.5-Turbo & 25.2137 & 0.7203 & 4.4099 & 70.0043 & 51.3186 \\
    & llama2-70B & -31.4144 & -1.0412 & 3.8014 & 60.3450 & 48.6173 \\
    & llama3-70B & -16.4424 & -0.4847 & 4.2743 & 67.8519 & 55.5486 \\
    & gemini & -0.3790 & -0.0148 & 3.2271 & 51.2280 & 35.6707 \\
\midrule
\multirow{7}{*}{NFLX} 
    & Buy \& Hold & 34.6251 & 1.3696 & 3.1852 & 50.5634 & 20.9263 \\
    & GPT-4 & \textbf{36.4485} & \textbf{2.0088} & 2.2860 & 36.2894 & \textbf{15.8495} \\
    & GPT-4o & 5.5829 & 0.2592 & 2.7132 & 43.0702 & 17.4715 \\
    & GPT3.5-Turbo & 7.9337 & 0.4610 & 2.1680 & 34.4160 & 17.9578 \\
    & llama2-70B & 33.8460 & 1.4741 & 2.8928 & 45.9216 & 20.3910 \\
    & llama3-70B & 21.7374 & 0.9513 & 2.8788 & 45.6989 & 21.3478 \\
    & gemini & 11.6298 & 1.0073 & \textbf{1.4546} & \textbf{23.0906} & 16.5106 \\
\midrule
\multirow{7}{*}{AMZN} 
    & Buy \& Hold & -16.4428 & -0.7448 & 2.7812 & 44.1508 & 33.8847 \\
    & GPT-4 & 10.5539 & 0.4923 & 2.7012 & 42.8802 & 22.9294 \\
    & GPT-4o & 11.3626 & 0.7334 & 1.9520 & 30.9864 & 19.5964 \\
    & GPT3.5-Turbo & \textbf{19.9636} & 0.9611 & 2.6171 & 41.5454 & 19.2191 \\
    & llama2-70B & 8.3595 & \textbf{1.9715} & \textbf{0.5342} & \textbf{8.4804} & \textbf{0.0000} \\
    & llama3-70B & 11.1479 & 0.5405 & 2.5986 & 41.2509 & 28.2174 \\
    & gemini & -2.3838 & -0.5321 & 0.5645 & 8.9605 & 6.4291 \\
\midrule
\multirow{7}{*}{MSFT} 
    & Buy \& Hold & 17.2161 & 0.9710 & 2.2339 & 35.4623 & 15.0097 \\
    & GPT-4 & 25.7826 & \textbf{1.5818} & 2.0535 & 32.5989 & \textbf{14.9889} \\
    & GPT-4o & -5.3731 & -0.5209 & \textbf{1.2997} & \textbf{20.6317} & 18.8223 \\
    & GPT3.5-Turbo & 20.4179 & 1.3600 & 1.8915 & 30.0259 & 20.3212 \\
    & llama2-70B & \textbf{27.7664} & 1.5708 & 2.2270 & 35.3524 & 15.0097 \\
    & llama3-70B & 21.1983 & 1.2628 & 2.1149 & 33.5724 & 15.0097 \\
    & gemini & 21.5081 & 1.3701 & 1.9777 & 31.3957 & 17.5051 \\
\midrule
\multirow{7}{*}{AAPL} 
    & Buy \& Hold & 12.7371 & 0.7759 & 2.0682 & 32.8323 & 20.6590 \\
    & GPT-4 & \textbf{21.2335} & \textbf{1.9274} & 1.3879 & 22.0328 & 6.4237 \\
    & GPT-4o & -6.7540 & -0.5693 & 1.4948 & 23.7285 & 20.7600 \\
    & GPT3.5-Turbo & 0.7110 & 0.0758 & \textbf{1.1817} & \textbf{18.7581} & \textbf{6.0818} \\
    & llama2-70B & 11.4856 & 1.1550 & 1.2529 & 19.8885 & 9.2776 \\
    & llama3-70B & -16.0835 & -1.1985 & 1.6907 & 26.8394 & 25.9520 \\
    & gemini & 18.1718 & 1.7214 & 1.3300 & 21.1134 & 9.6467 \\
\midrule
\multirow{7}{*}{GOOG} 
    & Buy \& Hold & 6.3107 & 0.3081 & 2.5806 & 40.9660 & 21.1907 \\
    & GPT-4 & 13.2811 & 0.9667 & 1.7308 & 27.4762 & 12.2209 \\
    & GPT-4o & 16.5072 & 1.0654 & 1.9520 & 30.9872 & \textbf{11.8863} \\
    & GPT3.5-Turbo & 0.9990 & 0.0614 & 2.0490 & 32.5265 & 20.9316 \\
    & llama2-70B & 17.0030 & \textbf{1.1057} & 1.9374 & 30.7546 & 13.2088 \\
    & llama3-70B & 17.5630 & 0.8872 & 2.4942 & 39.5934 & 19.2783 \\
    & gemini & \textbf{38.7956} & 3.0341 & \textbf{1.6110} & \textbf{25.5732} & 13.7311 \\
\midrule
\multirow{7}{*}{DIS} 
    & Buy \& Hold & -0.0700 & -0.0037 & 2.3667 & 37.5695 & 22.7722 \\
    & GPT-4 & \textbf{31.3383} & \textbf{2.3931} & 1.6498 & 26.1904 & 12.3417 \\
    & GPT-4o & -20.2500 & -1.3737 & 1.8573 & 29.4830 & 27.0246 \\
    & GPT3.5-Turbo & -7.1533 & -0.5109 & 1.7641 & 28.0048 & 20.4278 \\
    & llama2-70B & -3.8257 & -1.4323 & \textbf{0.3365} & \textbf{5.3420} & \textbf{4.1451} \\
    & llama3-70B & -25.5829 & -1.5579 & 2.0690 & 32.8437 & 31.3391 \\
    & gemini & 8.6692 & 0.8015 & 1.3627 & 21.6321 & 18.4815 \\
\midrule
\multirow{7}{*}{GM} 
    & Buy \& Hold & 0.3393 & 0.0179 & 2.3823 & 37.8181 & 23.0317 \\
    & GPT-4 & 10.5648 & 0.7671 & 1.7351 & 27.5443 & 11.1285 \\
    & GPT-4o & -7.0147 & -0.5263 & 1.6792 & 26.6569 & 21.5978 \\
    & GPT3.5-Turbo & -17.6385 & -0.9692 & 2.2928 & 36.3976 & 23.0317 \\
    & llama2-70B & 8.4911 & \textbf{2.6369} & \textbf{0.4057} & \textbf{6.4402} & \textbf{2.1318} \\
    & llama3-70B & 25.9335 & 1.9823 & 1.6483 & 26.1657 & 13.2485 \\
    & gemini & 18.6257 & 2.4672 & 0.9511 & 15.0989 & 3.0369 \\
\midrule
\multirow{7}{*}{NIO} 
    & Buy \& Hold & -49.4263 & -1.1895 & 5.2351 & 83.1048 & 52.2083 \\
    & GPT-4 & \textbf{24.7684} & \textbf{0.9438} & 3.3063 & 52.4861 & 29.3384 \\
    & GPT-4o & -48.3748 & -1.5026 & 4.0562 & 64.3897 & 59.4037 \\
    & GPT3.5-Turbo & -28.9321 & -1.0096 & 3.6105 & 57.3149 & 39.5907 \\
    & llama2-70B & -49.6947 & -2.7868 & \textbf{2.2466} & \textbf{35.6639} & 42.6221 \\
    & llama3-70B & -28.6668 & -0.7094 & 5.0912 & 80.8202 & 37.1544 \\
    & gemini & 14.5673 & 0.6212 & 2.9543 & 46.8977 & \textbf{23.0110} \\
\midrule
\multirow{7}{*}{COIN} 
    & Buy \& Hold & -18.4787 & -0.3369 & 6.9098 & 109.6904 & 60.5084 \\
    & GPT-4 & 25.7631 & 0.5619 & 5.7761 & 91.6934 & 35.7526 \\
    & GPT-4o & -14.2451 & -0.2892 & 6.2049 & 98.4997 & 65.3090 \\
    & GPT3.5-Turbo & 25.1141 & 0.4772 & 6.6312 & 105.2669 & 53.9628 \\
    & llama2-70B & 15.1836 & 0.4395 & \textbf{4.3528} & \textbf{69.0979} & \textbf{35.3249} \\
    & llama3-70B & 19.8876 & 0.3749 & 6.6842 & 106.1076 & 55.7225 \\
    & gemini & \textbf{89.4782} & \textbf{1.7648} & 6.3879 & 101.4048 & 40.3246 \\
\bottomrule
\end{tabular}}
\label{tab:trading_all}
\end{table*}
\graphicspath{ {./} }
\begin{figure}[h!]
\centering
  \includegraphics[width=0.5\textwidth]{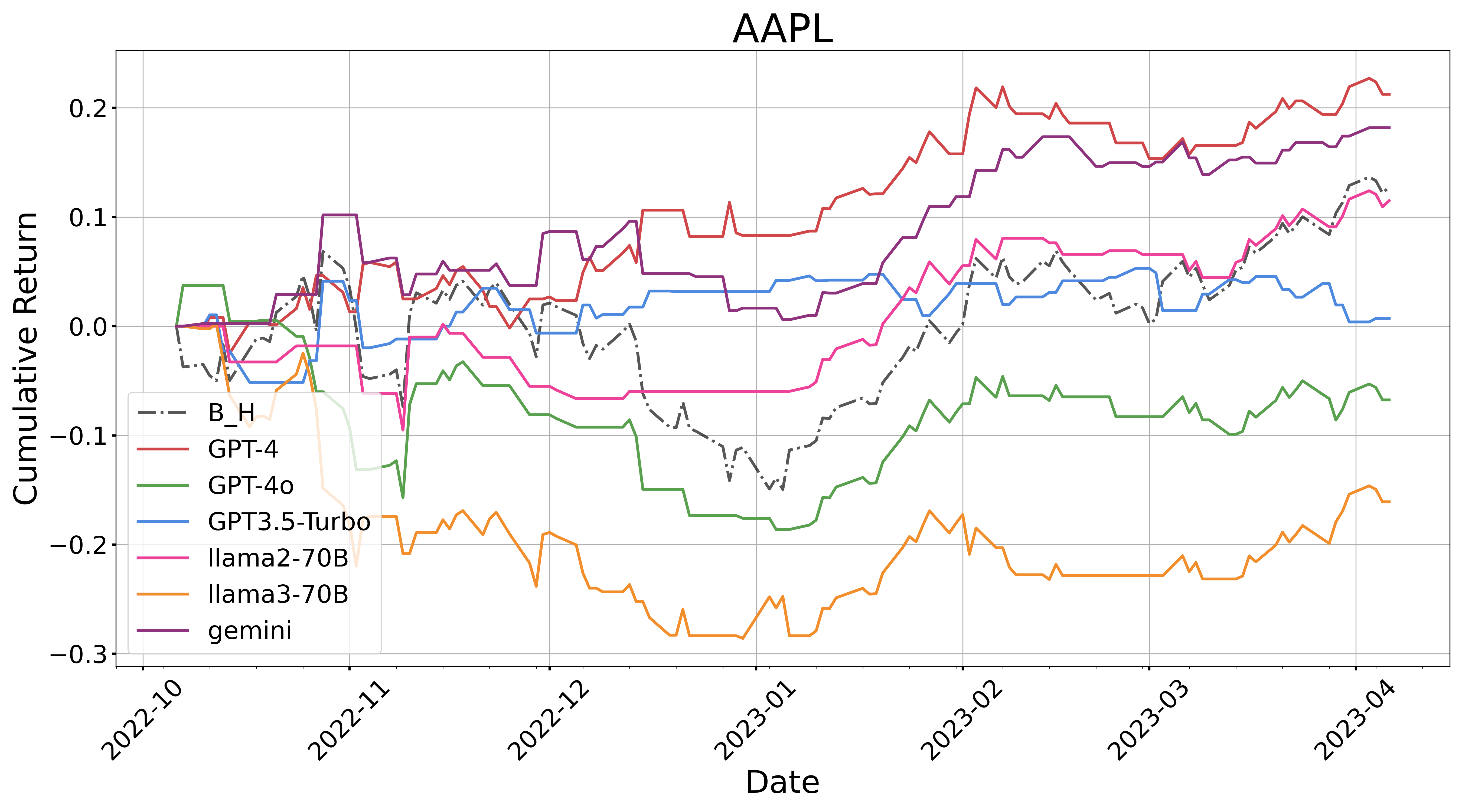}
  \caption{Accumulative Returns of LLM Trading Strategies on AAPL}
  \label{fig:AAPL}
\end{figure}
\begin{figure}[h!]
\centering
  \includegraphics[width=0.5\textwidth]{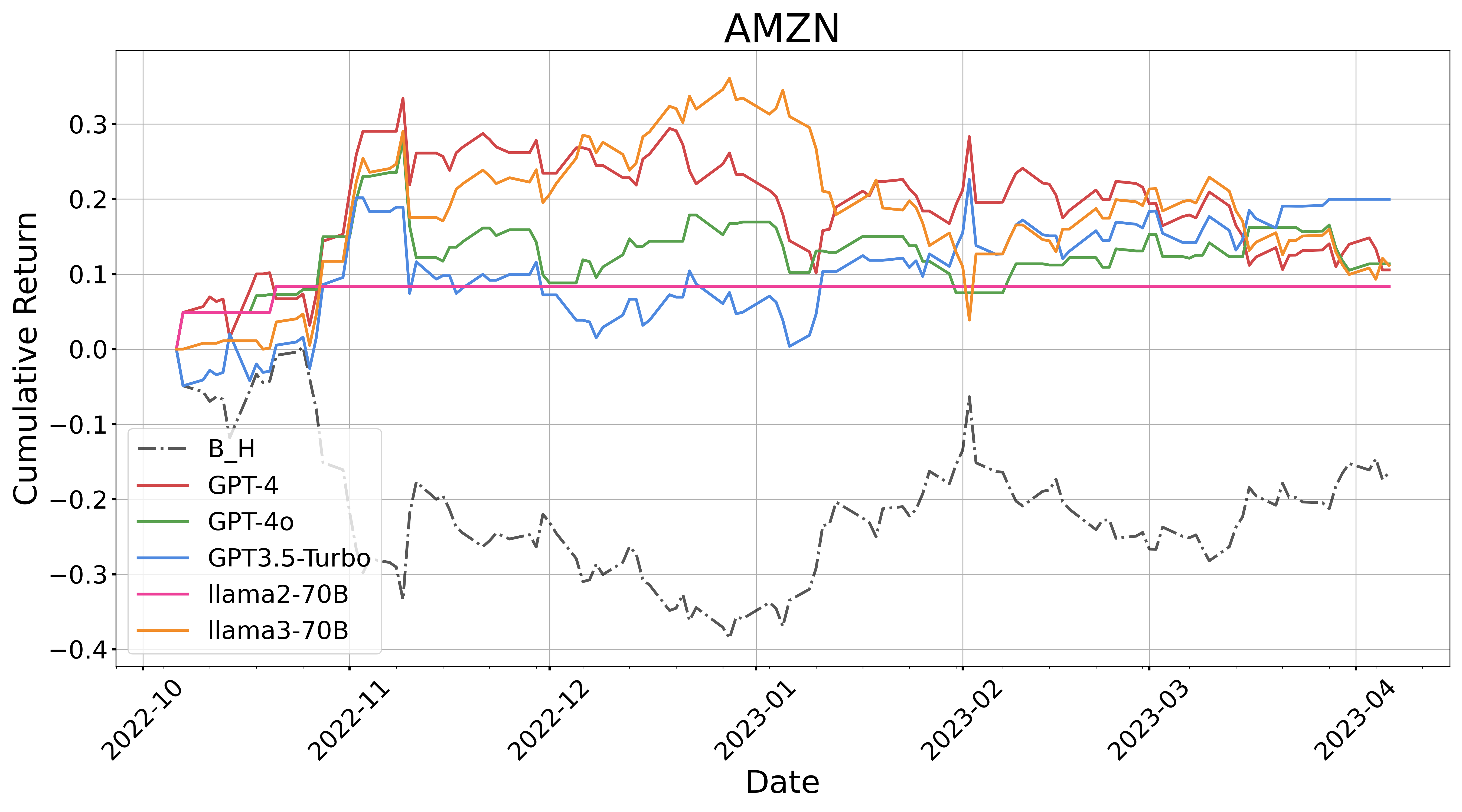}
  \caption{Accumulative Returns of LLM Trading Strategies on AMZN}
  \label{fig:AMZN}
\end{figure}
\begin{figure}[h!]
  \centering
  \includegraphics[width=0.5\textwidth]{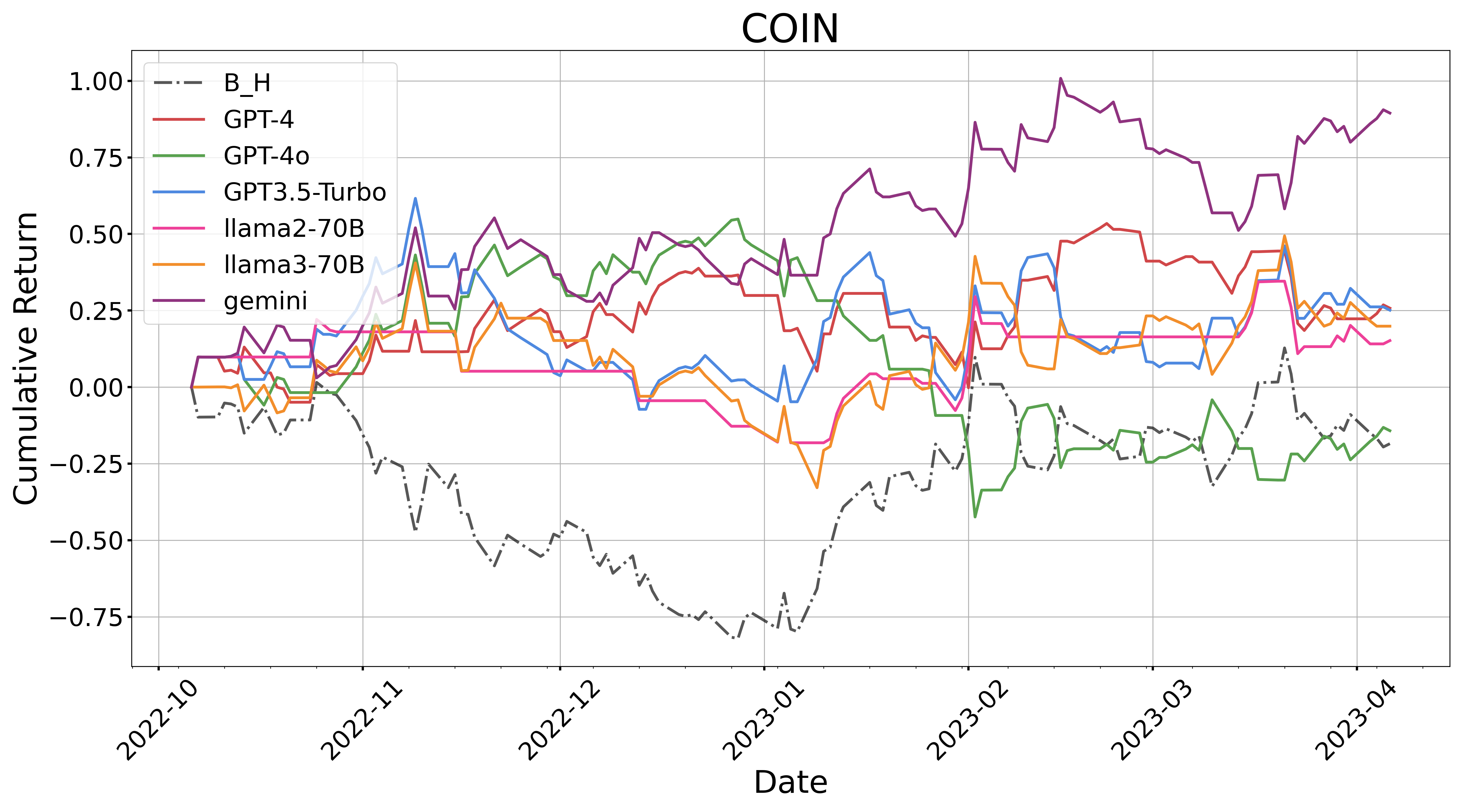}
  \caption{Accumulative Returns of LLM Trading Strategies on COIN}
  \label{fig:COIN}
\end{figure}
\begin{figure}[h!]
  \centering
  \includegraphics[width=0.5\textwidth]{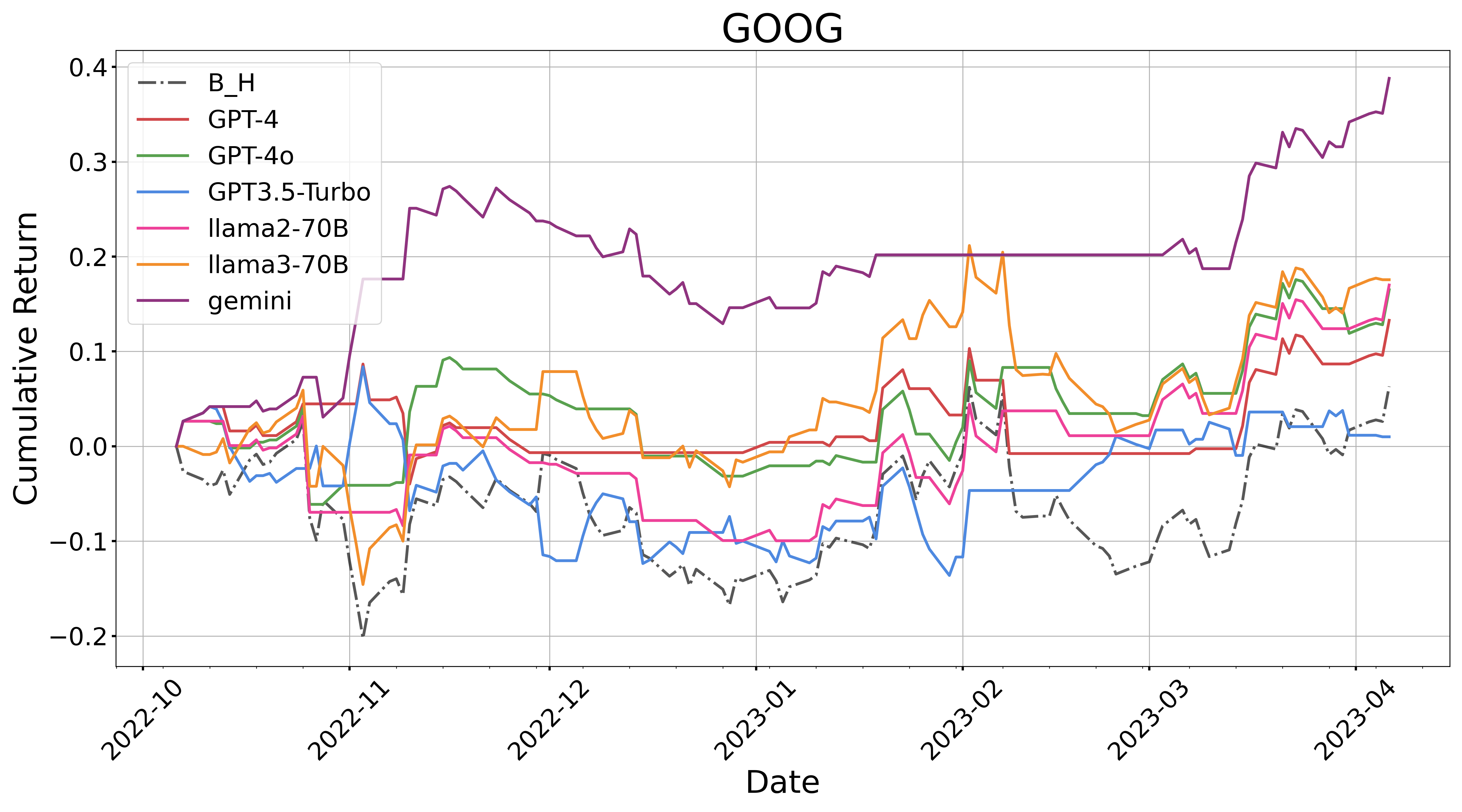}
  \caption{Accumulative Returns of LLM Trading Strategies on GOOG}
  \label{fig:GOOG}
\end{figure}
\begin{figure}[h!]
  \centering
  \includegraphics[width=0.5\textwidth]{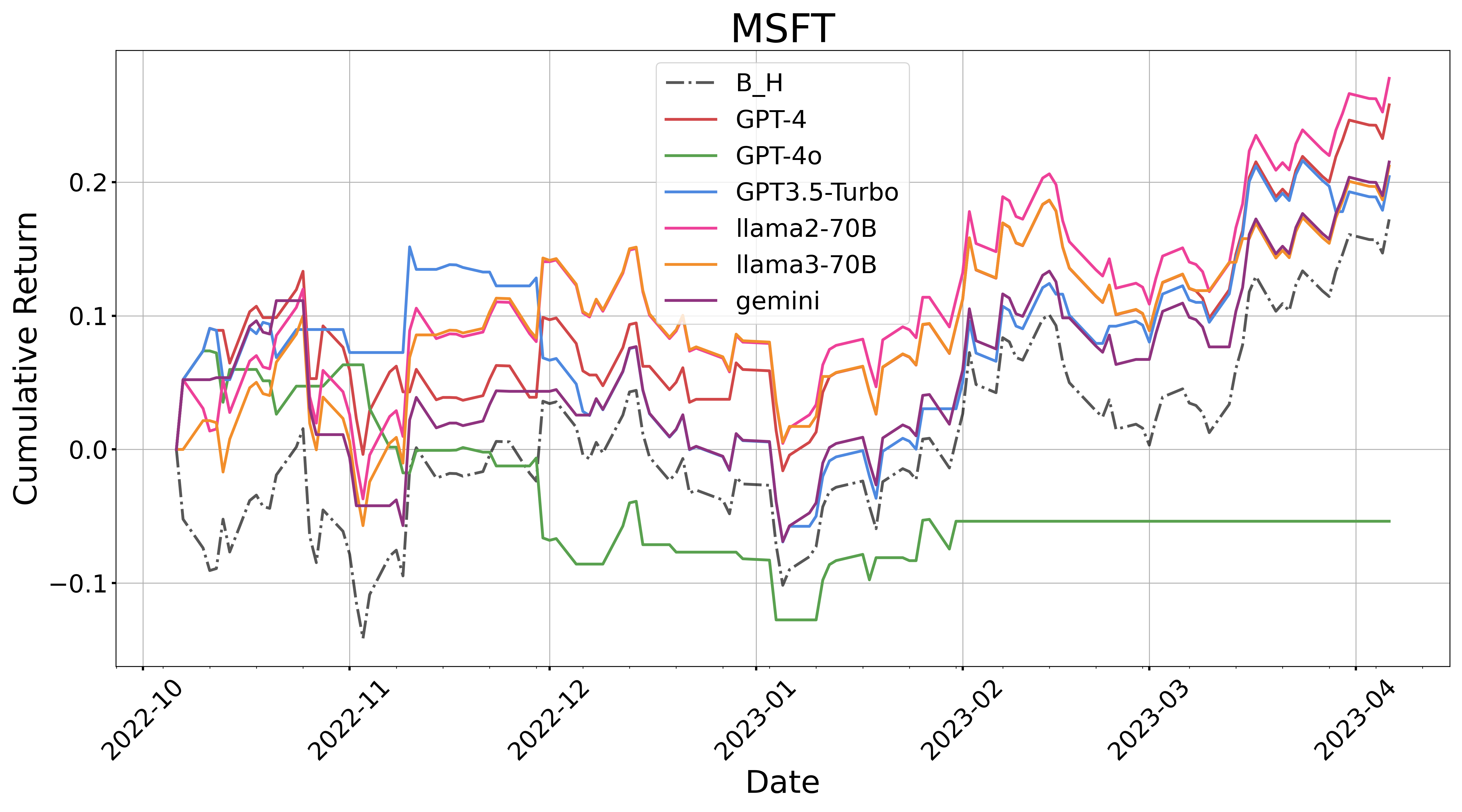}
  \caption{Accumulative Returns of LLM Trading Strategies on MSFT}
  \label{fig:MSFT}
\end{figure}
\begin{figure}[h!]
 \centering
  \includegraphics[width=0.5\textwidth]{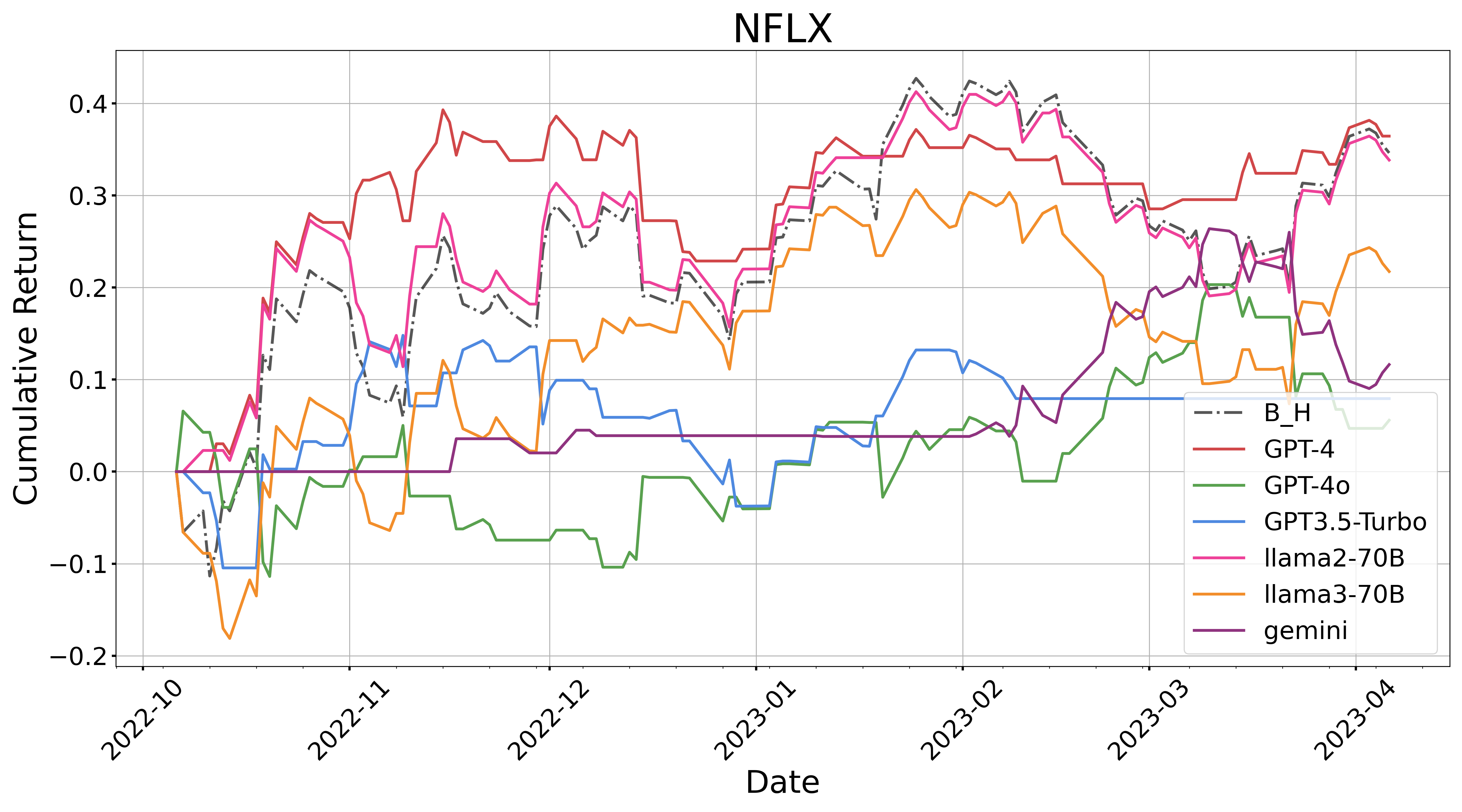}
   \caption{Accumulative Returns of LLM Trading Strategies on NFLX}
  \label{fig:NFLX}
\end{figure}
\begin{figure}[h!]
  \centering
  \includegraphics[width=0.5\textwidth]{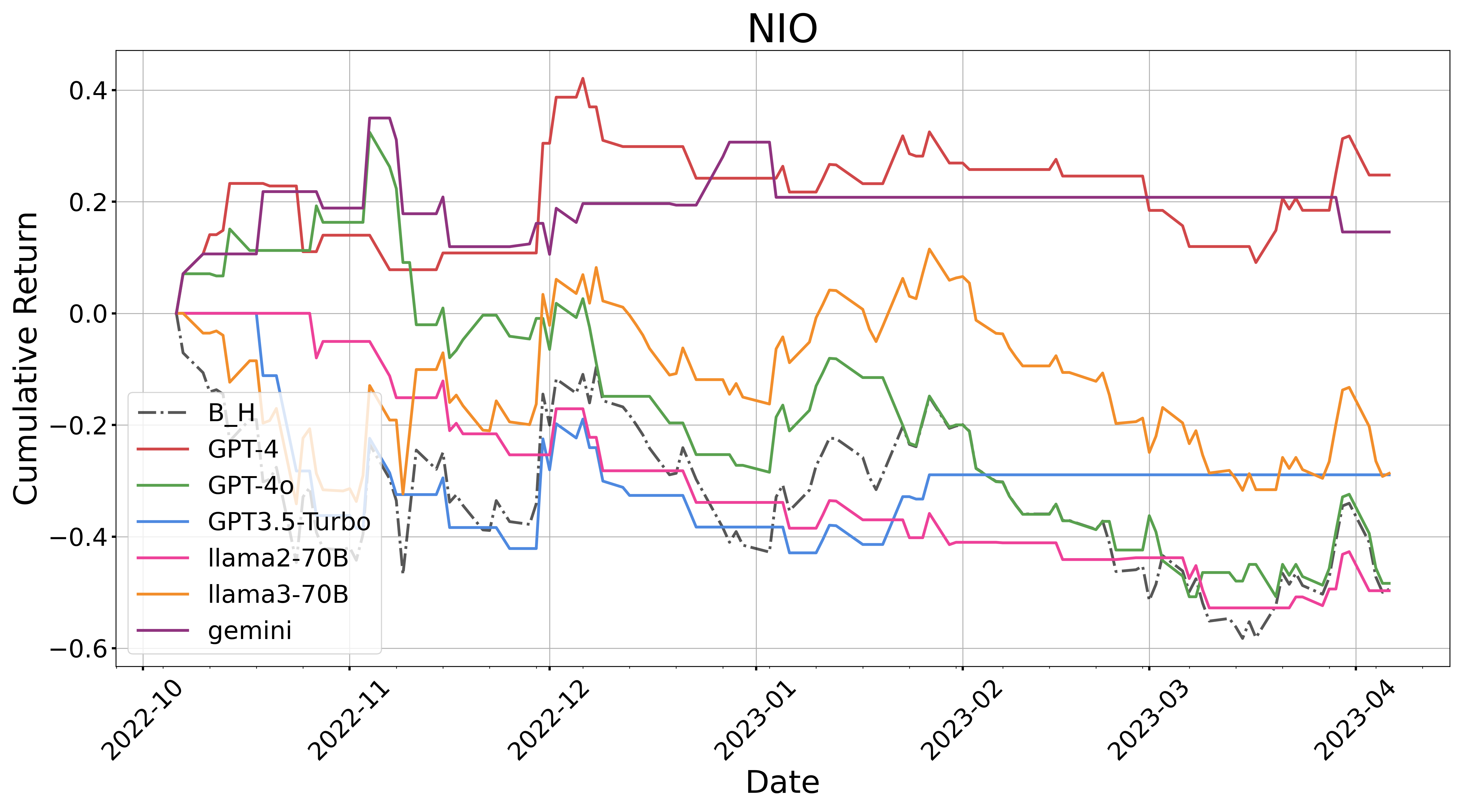}
  \caption{Accumulative Returns of LLM Trading Strategies on NIO}
  \label{fig:NIO}
\end{figure}
\begin{figure}[h!]
  \centering
  \includegraphics[width=0.5\textwidth]{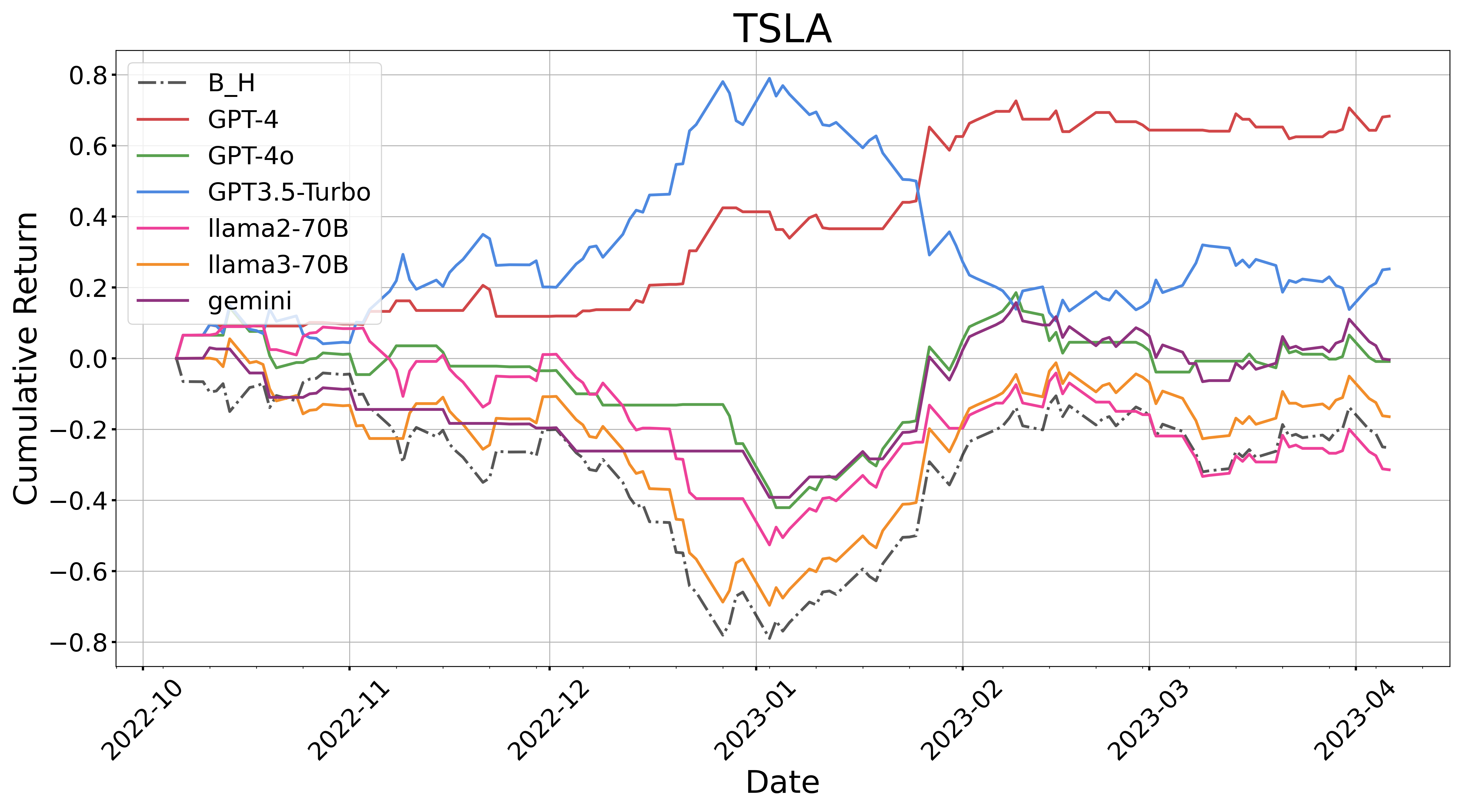}
  \caption{Accumulative Returns of LLM Trading Strategies on TSLA}
  \label{fig:TSLA}
\end{figure}
\begin{figure}[h!]
  \centering
  \includegraphics[width=0.5\textwidth]{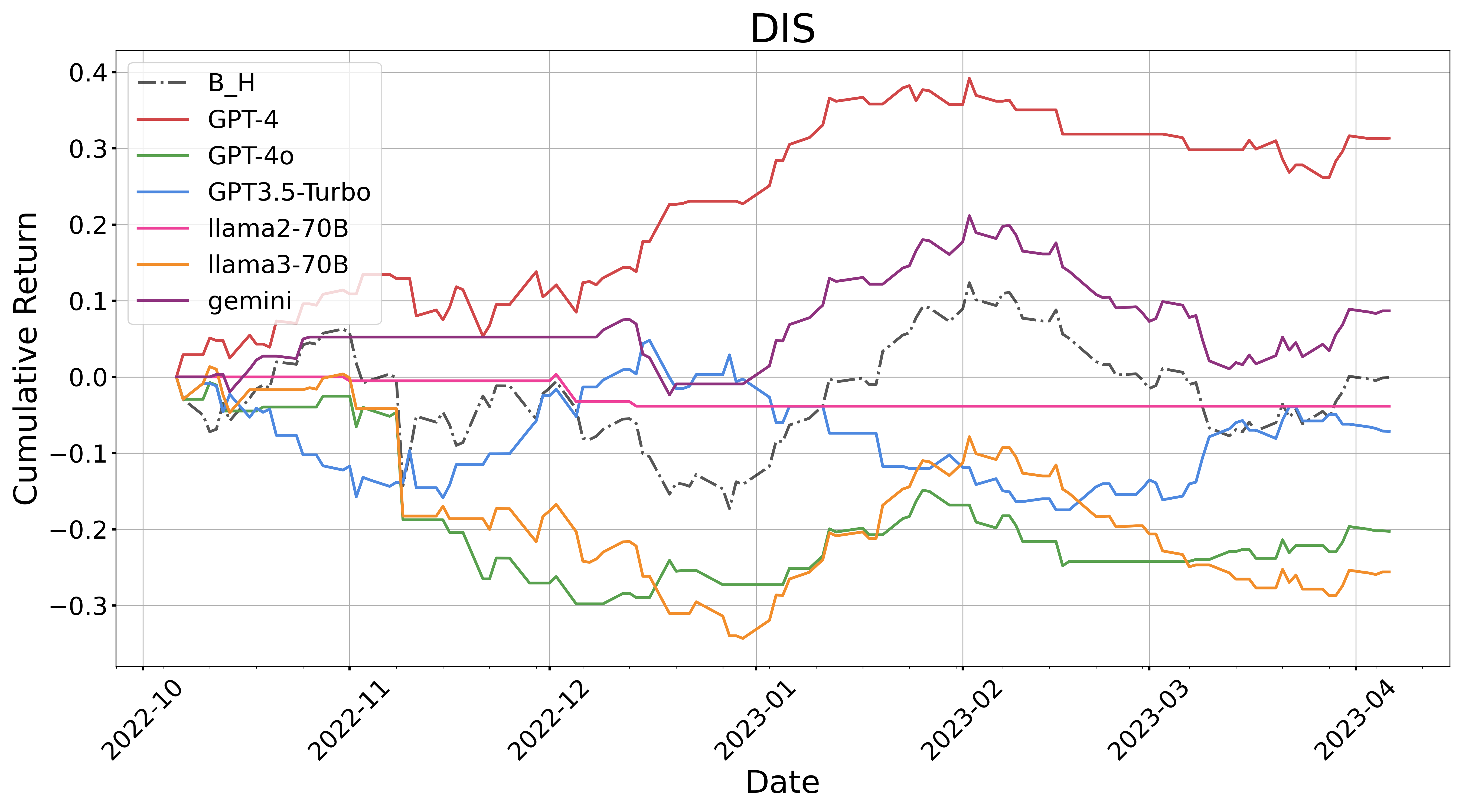}
  \caption{Accumulative Returns of LLM Trading Strategies on DIS}
  \label{fig:DIS}
\end{figure}
\begin{figure}[h!]
  \centering
  \includegraphics[width=0.5\textwidth]{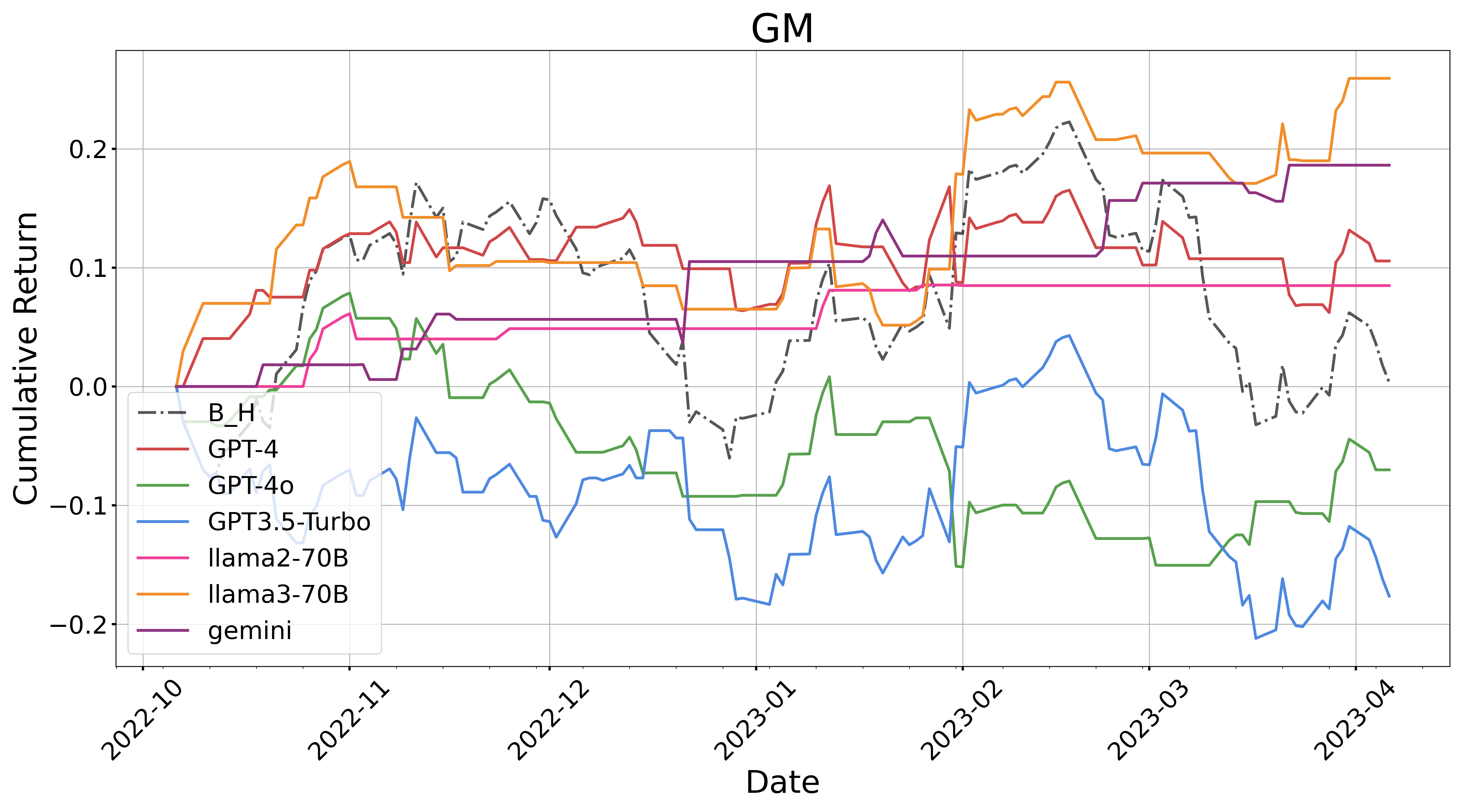}
  \caption{Accumulative Returns of LLM Trading Strategies on GM}
  \label{fig:GM}
\end{figure}

\section{FinLLM challenge}
\label{append:finllm_challenge}


Based on our proposed FinBen, we organized the FinLLM Share Task during the FinNLP-AgentScen Workshop at IJCAI 2024\footnote{\url{https://sites.google.com/nlg.csie.ntu.edu.tw/finnlp-agentscen/shared-task-finllm?authuser=0}}, known as the FinLLM Challenge. This challenge not only tests the abilities of LLMs but also promotes ongoing research into their application within the financial sector, highlighting FinBen's critical contribution to the advancement of financial analytics.

The FinLLM Challenge is a specialized shared task tailored for LLMs, targeting a comprehensive range of financial problems through three subtasks: financial classification, financial text summarization, and single stock trading. To rigorously evaluate the capabilities of financial LLMs, we have curated three distinct datasets corresponding to each of these subtasks, as detailed in Table \ref{tab:tasks and datasets}. This structured approach ensures a holistic and effective assessment of LLM performance across diverse financial scenarios.

\begin{table}[h]
   \scriptsize
\centering
\renewcommand{\arraystretch}{1.2}
     \caption{Tasks and Datasets of FinLLM Challenge.}
   \begin{tabular}{cccccc}
    \hline
        Category & Tasks  
        & \multicolumn{2}{c}{Datasets}  & Evaluation Metrics \\
        & &  Training set & Test set & \\
        \hline
        Task 1 &  Financial Classification  & 7.75k  & 969 & F1 Score, Acc \\
        Task 2 &  Financial Text Summarization  & 8k & 2k &  ROUGE-1, ROUGE-2, ROUGE-L, BERTScore \\ 
        \multirow{2}{*}{Task 3} 
        &\multirow{2}{*}{ Single Stock Trading}
        & \multirow{2}{*}{ 291 }
        & \multirow{2}{*}{ 225 }
       & Sharpe Ratio, Cumulative Return,  \\ 
        & & & & Maximum Drawdown, Daily and Annualized Volatility,
        \\ \hline
    \end{tabular}    
    \label{tab:tasks and datasets}   
\end{table}

\subsection{Tasks and Datasets}
\noindent
\textbf{Task 1: Financial Classification.} This task, inherited from FinBen's financial classification task, focuses on argument unit classification to test the capabilities of LLMs to identify and categorize texts as premises or claims. It consists of 7.75k training data and 969 test data to categorize sentences as claims or premises.  We use two metrics to evaluate classification capability, like F1 and Accuracy. F1 score is used as the final ranking metric.

\noindent
\textbf{Task 2: Financial Text Summarization.} This task, inherited from FinBen's generation task, is designed to test the capabilities of LLMs to generate coherent summaries. It provides 8k training data and 2k test data for abstracting financial news articles into concise summaries. We utilize three metrics, such as ROUGE (1, 2, and L) and BERTScore, to evaluate generated summaries in terms of Relevance.  ROUGE -1 score is used as the final ranking metric.  

\noindent
\textbf{Task 3: Single Stock Trading.} This task, inherited from FinBen's Trading task, aims to evaluate LLMs’ ability to make sophisticated decisions in trading activities, which is currently restricted by human’s limited ability to process large volumes of data rapidly. It specifically provides 291 data different from FinBen datasets, to evaluate LLMs on sophisticated stock Decisions. We offer a comprehensive assessment of profitability, risk management, and decision-making prowess by a series of metrics, such as Sharpe Ratio (SR), Cumulative Return (CR), Daily (DV) and Annualized volatility (AV), and Maximum Drawdown (MD).  Sharpe Ratio (SR) score is used as the final ranking metric.

\subsection{Model Cheating Detection}
To measure the risk of data leakage from the test set used in training, we introduce the Data Leakage Test (DLT). The DLT calculates the difference in perplexity between the training set and the test set. A larger difference indicates a lower likelihood of model cheating, while a smaller difference suggests a higher likelihood. For our FinLLM Challenge, we invite Top-3 participant teams per task for cheating detection.

\subsection{Participants and Automatic Evaluation}

There are 35 teams registered for FinLLM Challenge, with 11 teams submitting their system description papers. Participants can opt to join one or more task(s). 

As shown in Table \ref{tab:task1}, the top 3 teams achieved outstanding performance in Task 1. Their models' F1 scores were comparable to LlaMA3-8B, although slightly inferior to GPT-4 and LLaMA2-70B, yet significantly outperformed FinMA and other models. The results in Table \ref{tab:task1} further demonstrate that our FinLLM share task provides an excellent framework for participating teams to achieve superior experimental outcomes.

\begin{table}[h]
    \scriptsize
    \centering
    \renewcommand{\arraystretch}{1.2}
     \caption{The Result of Taks 1: Financial Classification}
     \scalebox{0.8}{
   \begin{tabular}{cccc}
    \hline
        Teams & ACC & F1 & MCC \\
        \hline
        Team Barclays & 0.7626 & 0.5237 & 0.7427 \\
        Albatross  & 0.7574 & 0.5174 & 0.7555 \\
        L3iTC & 0.7544 & 0.5149 & 0.7581 \\
        Wealth Guide & 0.7513 & 0.5018 & 0.7406 \\
        Finance Wizard & 0.7286 & 0.4554 & 0.7008 \\
        CatMemo & 0.711 & 0.4199 & 0.6818 \\ 
         Upaya& 0.709 & 0.4166 & 0.6941 \\
        Vidra & 0.7079 & 0.4141 & 0.69 \\ 
        jt & 0.4933 & 0.0141 & 0.5905 \\  
        \hline
    \end{tabular}}    
    \label{tab:task1}   
\end{table}

As illustrated in Table \ref{tab:task2}, in terms of the Rouge-1 metric, the models of these three teams surpassed all other models, demonstrating superior performance. The results in Table 2 indicate that, for financial generation tasks, our provided dataset and model framework help participating teams leverage their strengths and achieve better outcomes.

\begin{table}[h]
  \scriptsize
  \centering
  \renewcommand{\arraystretch}{1.2}
     \caption{The Result of Taks 2: Financial Text Summarization}
   \scalebox{0.8}{
   \begin{tabular}{cccccc}
    \hline
        Teams & Rouge-1 & Rouge-2 & Rouge-L & BertScore & BartScore \\
        \hline
         Wealth Guide & 0.308893532 & 0.179468097 & 0.281924302 & 0.85959909 & -4.961457408\\
         Albatross  & 0.369077581 & 0.201058395 & 0.322684316 & 0.872049115 & -3.933526929 \\
         LBZ & 0.534616211 &0.358105428 & 0.492179554 &0.911732047 &-3.407560172 \\
         L3iTC & 0.366093426 & 0.187210467 & 0.304610677 &0.875037043 & -4.257126737 \\
         Finance Wizard & 0.521037018 & 0.34060938 & 0.473530112 & 0.90836845 & -3.497988865\\
         Vidra & 0.284955468 & 0.134760859 & 0.228638961 & 0.858682767 & -4.169740305 \\ 
         Revelata & 0.500411369 & 0.333023818 & 0.464356474 & 0.907018743 & -3.805486962 \\
         Upaya & 0.529459817 & 0.358203218 & 0.486046685 & 0.910644962 & -3.45155009 \\
        \hline
    \end{tabular}}    
    \label{tab:task2}   
\end{table}

As shown in Table\ref{tab:task3}, the Top-1 Wealth Guide team excelled in the Sharpe Ratio metric, surpassing other teams and demonstrating outstanding performance. While it may not match the performance of GPT-4, it still outperforms other large models. These results from Table 3 once again underscore the significance of organizing the FinLLM share task. The FinLLM Challenge not only assesses the performance of large language models (LLMs) but also fosters further research into applying LLMs in the financial domain.

\begin{table}[h]
   \scriptsize
\centering
\renewcommand{\arraystretch}{1.2}
     \caption{The Result of Taks 3: Single Stock Trading}
     \scalebox{0.8}{
   \begin{tabular}{cccccc}
    \hline
        Teams & Sharpe Ratio & Sharpe Ratio-DRIV & Sharpe Ratio-FORM & Sharpe Ratio-JNJ & Sharpe Ratio-MSFT \\
        \hline
         Wealth Guide & 0.9263852228 & 0.485625528 & 1.585611423 & 0.078737051 & 1.555566991\\
         Upaya & 0.467489019 & 0.380232272 & 0.108506918 & -1.102831656 & -0.278385232 \\
         Albatross  & 0.48383204 & 0.251306057 & -1.435471054 & -1.558522674 & 1.309971626 \\
         CatMemo & -0.619939784 & -1.393291177 & 0.175932289 & 0.383243051 & -0.879157198 \\
        \hline
    \end{tabular}}    
    \label{tab:task3}   
\end{table}

\section{Performances of non-LLM methods}
\label{append:non_llm_result}
In this section, we present the performances of non-LLM methods on stock movement prediction and financial NLP tasks from previous papers. Note that non-LLM methods are task-oriented, each model can only run on a specific task.

\subsection{Stock movement prediction}
Stock movement prediction performance of non-LLM models are shown in Table \ref{table:stock_movement_non_llm}. The results are from~\citep{xie2023pixiu}.

\begin{table}[h]
\scriptsize
\centering
\renewcommand{\arraystretch}{1.2}
\caption{Stock movement prediction performance of non-LLM models, measured with the accuracy (ACC) and the Matthews correlation coefficient (MCC). The best performance is in bold.}
\begin{tabular}{ccccccccc}
\hline
{Method} 
& \multicolumn{2}{c}{BIGDATA22} 
& \multicolumn{2}{c}{ACL18} 
& \multicolumn{2}{c}{CIKM18} \\
 & ACC & MCC & ACC & MCC & ACC & MCC \\
\hline
Logistic regression (LR) & 0.53 & 0.02 & 0.52 & 0.04 & 0.53 & -0.04 \\
Random forest (RF) & 0.47 & -0.11 & 0.52 & 0.03 & 0.54 & 0.01 \\
LSTM & 0.51 & 0.01 & 0.53 & 0.06 & 0.53 & 0.02 \\
Attention LSTM (ALSTM) & 0.49 & -0.03 & 0.52 & 0.04 & 0.53 & -0.01 \\
Adv-ALSTM & 0.50 & 0.01 & 0.53 & 0.07 & 0.54 & 0.02 \\
DTML & 0.52 & 0.07 & 0.58 & 0.18 & 0.54 & -0.00 \\
XGBoost & 0.52 & -0.04 & 0.49 & -0.02 & \textbf{0.58} & 0.07 \\
XGBRegressor & 0.46 & -0.13 & 0.50 & -0.01 & 0.53 & -0.03 \\
ALSTM-W & 0.48 & -0.01 & 0.53 & 0.08 & 0.54 & 0.03 \\
ALSTM-D & 0.49 & 0.01 & 0.53 & 0.07 & 0.50 & -0.04 \\
StockNet & 0.53 & -0.00 & 0.54 & -0.03 & 0.52 & -0.02 \\
SLOT & \textbf{0.55} & \textbf{0.10} & \textbf{0.59} & \textbf{0.21} & 0.56 & \textbf{0.09} 
\\ \hline
\end{tabular}
\label{table:stock_movement_non_llm}
\end{table}

\subsection{Financial NLP tasks}
BERT-based model results of financial NLP tasks are shown in Table \ref{table:financial_nlp_task}. The results are from~\citep{shah2022flue}.

\begin{table}[h]
\centering
\scriptsize
\renewcommand{\arraystretch}{1.2}
\caption{Financial NLP tasks performances of BERT-based models. The best performance is in bold.}
\begin{tabular}{ccccc}
\hline
{Method} & {FPB} & {Headline} & {NER} & {FiQA SA}\\
& Accuracy & AvgF1 & F1 & MSE \\
\hline
BERT-base & 0.856 & 0.967 & 0.79 & 0.073 \\
FinBERT & 0.872 & 0.968 & 0.8 & 0.070 \\
FLANG-BERT & 0.912 & 0.972 & 0.83 & 0.054 \\
ELECTRA & 0.881 & 0.966 & 0.78 & 0.066 \\
FLANG-ELECTRA & 0.919 & 0.98 & 0.82 & 0.034 \\
\hline
\end{tabular}
\label{table:financial_nlp_task}
\end{table}

\section*{Limitations}
\label{sec:limitation}
Despite the novel efforts to benchmark LLMs in the financial domain through FinBen, we acknowledge several inherent limitations that could impact the benchmark's effectiveness and applicability:

Despite the novel efforts to benchmark LLMs in the financial domain through FinBen, several limitations could impact its effectiveness and applicability:
\textbf{Dataset Size Limitations}: The restricted size of available datasets, a common issue in open-source financial data, may affect the models' financial understanding and generalization across various contexts.
\textbf{Model Size Limitations}: Due to computational constraints, our evaluation was limited to the LLaMA 70B model, potentially overlooking the capabilities of larger or differently architected models.
\textbf{Generalizability}: The tasks, particularly trading and forecasting, are based on American market data and English texts, possibly limiting the benchmark's applicability to global financial markets.
\textbf{Potential Negative Impacts}: While FinBen aims to advance financial language understanding, it is crucial to consider potential misuse, such as propagating financial misinformation or exerting unethical influence on markets. Responsible usage and further safeguards are essential\footnote{For a detailed ethical and legal statement concerning this work, please see Appendix.}.

\section*{Ethical Statement}
\label{sec:ethical}
The development and dissemination of the FinBen by the authors carry full responsibility for any potential violation of rights or arising legal issues. All raw data we used are publicly available and do not contain any personal information. Diligent efforts have been undertaken to ensure the construction of the FinBen respects privacy and conforms to established ethical guidelines. The datasets compiled within FinBen are shared under the MIT license, with the expectation that users agree to adhere to its conditions.

This manuscript, inclusive of any associated source codes, datasets, and appendices ("Material"), is designated exclusively for academic and educational pursuits. It is crucial to acknowledge that the Material does not provide financial, legal, or investment counsel, nor should it be utilized as a foundation for any form of decision-making.

While the authors have exerted reasonable diligence to verify the accuracy and reliability of the Material, no explicit or implied warranty is extended regarding its completeness or suitability for any specific application. The authors, along with their affiliated entities, absolve themselves of liability for any losses, damages, or other consequences, whether direct or indirect, that may emanate from the employment or reliance upon the Material. It is incumbent upon the user to seek professional consultation for financial, legal, or investment determinations.

By referencing or employing this Material, individuals consent to indemnify, defend, and hold the authors, along with any affiliated organizations or persons, harmless against any claims or damages that may arise from such utilization.

\end{document}



\appendix

\section{Additional Results}
The performance analysis of models on the acronym and regulations tasks, as shown in Tables \ref{table:acronym} and \ref{table:regulations}, provides valuable insights into their capabilities. 

The acronym dataset is a QA task that requires models to decode financial acronyms. Despite not having seen this task before, FinMA, a financial LLM specially trained on financial tasks, performed exceptionally well. The FinMA7B-full model achieved the highest ROUGE-1 score of 0.12 and the highest BERTScore of 0.73, even surpassing GPT-4. This indicates that financial-specific models can leverage their domain knowledge effectively, even on short QA tasks like the acronym dataset.

On the other hand, the regulations dataset involves answering intricate questions related to financial regulations, such as EMIR. This task is long, complex, and difficult to understand, posing a significant challenge. In this scenario, the LLaMA2-70b-chat model stand out with a ROUGE-1 score of 0.30 and a BERTScore of 0.68, highlighting its ability to handle complex regulatory questions. This underscores the importance of model size and capability when dealing with more demanding and sophisticated tasks in the financial domain.

\begin{table}[h]
\centering
\caption{Results of acronym dataset. The best performance is in bold.}
\renewcommand\arraystretch{1.3}
\scalebox{0.55}{
\begin{tabular}{cccccccccccccc}
\hline
& \makecell[c]{LLaMA3\\-8B} 
& \makecell[c]{LLaMA2\\-7B-hf}
& \makecell[c]{FinMA7B\\-full} 
& \makecell[c]{InternLM\\-chat-7B} 
& \makecell[c]{FinGPT\\-mt} 
& \makecell[c]{Falcon\\-7B}
& \makecell[c]{cfGPT} 
& \makecell[c]{Baichuan\\-7B}
& \makecell[c]{CodeLLaMA}
& \makecell[c]{DISC\\-FinLLM} 
& \makecell[c]{ChatGLM3\\-6b}
& \makecell[c]{LLaMA2\\-70b-chat}
& \makecell[c]{Mistral-7B\\-instruct}
\\ \hline
ROUGE-1  
& 0.02     
& 0.03    
& \textbf{0.12}     
& 0.10      
& 0      
& 0.01    
& 0.05 
& 0  
& 0.01 
& 0.02 
& 0.04  
& 0.02
& 0.12
\\
BERTScore
& 0.55   
& 0.57  
& \textbf{0.73}    
& 0.68    
& 0.64     
& 0.54  
& 0.69
& 0.48 
& 0.45    
& 0.57      
& 0.62 
& 0.58
& 0.72
\\ \hline
\end{tabular}}
\label{table:acronym}
\end{table}

\begin{table}[h]
\centering
\caption{Results of regulations dataset. The best performance is in bold.}
\renewcommand\arraystretch{1.3}
\scalebox{0.55}{
\begin{tabular}{cccccccccccccc}
\hline
& \makecell[c]{LLaMA3\\-8B}
& \makecell[c]{LLaMA2\\-7B-hf} 
& \makecell[c]{FinMA7B\\-full}
& \makecell[c]{InternLM\\-chat-7B}
& \makecell[c]{FinGPT\\-mt} 
& \makecell[c]{Falcon\\-7B}
& \makecell[c]{cfGPT} 
& \makecell[c]{Baichuan\\-7B}
& \makecell[c]{CodeLLaMA}
& \makecell[c]{DISC\\-FinLLM}
& \makecell[c]{ChatGLM3\\-6b}
& \makecell[c]{LLaMA2\\-70b-chat}
& \makecell[c]{Mistral-7B\\-instruct}

\\ \hline
ROUGE-1 
& 0.10  
& 0.24  
& 0.12    
& 0.04     
& 0.01      
& 0.03   
& 0.14 
& 0.13 
& 0.17 
& 0.12   
& 0.26   
& \textbf{0.30}
& 0.28
\\
BERTScore 
& 0.60   
& 0.65   
& 0.59     
& 0.57     
& 0.40     
& 0.14 
& 0.57 
& 0.60 
& 0.59  
& 0.52  
& 0.65  
& \textbf{0.68}
& 0.67
\\ \hline
\end{tabular}}
\label{table:regulations}
\end{table}

\section{Motivation For Datasheet Creation}

\textcolor{blue}{\subsection{Why was the datasheet created? (e.g., was there a specific task in mind? was there a specific gap that needed to be filled?)}}

FinBen was created to address the gap in comprehensive benchmarks and evaluation studies of Large Language Models (LLMs) within the financial domain. Despite the proven capabilities of LLMs such as GPT-4 in transforming various fields including finance, a detailed understanding of their potential and limitations specific to finance is still lacking. This is partly due to the complex and specialized nature of financial tasks, which necessitates targeted datasets for thorough analysis. By creating 36 datasets covering 24 financial tasks, we aim to provide a robust benchmark that allows researchers and practitioners to evaluate the effectiveness of LLMs in financial text analysis and prediction tasks more accurately and reliably. These datasets are thus a critical step towards leveraging the full capabilities of LLMs in the finance sector, ensuring that their deployment is both effective and appropriate for the intricacies of financial applications.

\textcolor{blue}{\subsection{Has the dataset been used already? If so, where are the results so others can compare
(e.g., links to published papers)?}}

Yes, the dataset has already been used. It was employed in the FinLLM Share Task during the FinNLP-AgentScen Workshop at IJCAI 2024, known as the FinLLM Challenge. This event saw active participation, with 35 teams registering to take part and 11 of those teams submitting system description papers. This indicates that datasets of FinBen, have been utilized to test and explore the capabilities of LLMs  within the financial sector, thereby promoting further research and understanding in this area. The link of the challenge is \url{https://sites.google.com/nlg.csie.ntu.edu.tw/finnlp-agentscen/shared-task-finllm?authuser=0}.

\textcolor{blue}{\subsection{What (other) tasks could the dataset be used for?}}

The 36 datasets of FinBen, can be employed for several tasks within the financial technology (FinTech) sector, covering seven critical aspects: information extraction (IE), textual analysis, question answering (QA), text generation, risk management, forecasting, and decision-making. It can be utilized to train and evaluate LLMs for a variety of applications, including:

\noindent
\textbf{Financial Sentiment Analysis}: Analyzing sentiments in financial texts such as market news, analyst reports, and social media to gauge investor sentiment and predict market trends.

\noindent
\textbf{Fraud Detection}: Enhancing models that detect fraudulent activities by analyzing transaction patterns and communication within financial institutions.
   
\noindent
\textbf{Financial Forecasting}: Improving predictions of financial markets, stock prices, economic indicators, or company performance metrics based on historical data and current market conditions.
   
\noindent
\textbf{Personalized Financial Advice}: Generating customized financial advice for individuals based on their spending habits, investment preferences, and financial goals.
   
\noindent
\textbf{Regulatory Compliance}: Assisting in compliance monitoring by analyzing communications and transactions to ensure they meet legal and regulatory standards.

These applications demonstrate the potential of the FinBen dataset to significantly impact various aspects of financial technology by providing robust training and evaluation grounds for sophisticated LLMs tailored to specific financial tasks.

\textcolor{blue}{\subsection{Who funded the creation dataset?}}

FinBen is a collaborative project carried out by researchers from several institutions, including Wuhan University, 
the University of Manchester, 
University of Florida,
Columbia University, 
The Chinese University of Hong Kong, Shenzhen,
Sichuan University, 
Yunnan University, 
Stevens Institute of Technology,
Stony Brook University, 
Nanjing Audit University,
Jiangxi Normal University, 
Southwest Jiaotong University. 
The project is funded by the Fin AI.

\textcolor{blue}{\subsection{Any other comment?}}

FinBen represents a significant breakthrough in the domain of financial AI. It aims to bridge the gap in the lack of open-source financial benchmarks for evaluating Large Language Models (LLMs) and providing comprehensive datasets and evaluation metrics for these tasks. It is meticulously designed to enhance the performance of financial LLMs in downstream tasks by providing a novel taxonomy for organizing financial evaluation tasks and systematic evaluation metrics.

Moreover, FinBen is designed to foster transparency and collaboration in the field. It openly provides instruction-tuning data, evaluation datasets, and a structured framework included in the benchmark to encourage open research. By covering a diverse set of financial tasks, FinBen offers a versatile tool for both academic researchers and industry professionals.

By providing these resources and promoting open-source development, FinBen is set to push forward the frontier of financial AI, offering a comprehensive framework for assessing financial LLMs and fostering a deeper understanding of complex financial language and concepts. This project is a significant step towards the integration of advanced AI techniques into the financial industry, paving the way for a wide range of applications such as predicting stock price movements and advanced financial analytics.

\section{Datasheet Composition}

\textcolor{blue}{\subsection{What are the instances?(that is, examples; e.g., documents, images, people, countries) Are there multiple types
of instances? (e.g., movies, users, ratings; people, interactions between them; nodes, edges)}}
The instruction dataset within the FinBen benchmark comprises individual samples of data that are used to evaluate the model. These instances are varied and encompass multiple types, reflecting the diverse tasks and data sources prevalent in the financial domain.For instance, in the named entity recognition task, an example would involve a sentence from a financial agreement, where the model is instructed to identify named entities corresponding to a person, organization, or location.

The data types include:
\begin{itemize}
    \item {Textual Data}: Predominantly, the instructions pertain to processing textual data such as reports, news articles, news headlines, regulatory filings, tweets, and financial agreements. Tasks associated with this data type include sentiment analysis, news headline classification, and named entity recognition.
    \item {Time-Series Data}: Instructions sometimes relate to time-series data, specifically stock price data, often used in conjunction with textual data for tasks like stock movement prediction.
    \item {Tabular Data}: Instructions also cover the processing of tabular data, typically derived from companies' financial filings. This data is essential for question-answering tasks where the model needs to extract and leverage information from financial tables.
\end{itemize}

\textcolor{blue}{\subsection{How many instances are there in total (of each type, if appropriate)?}}
The test dataset is broken down by each task type, with the total number of instances for each as follows:

\begin{itemize}
\item Sentiment Analysis and Related Tasks: 4,619 instances
\item Stock Movement Prediction: 6,330 instances
\item Question Answering and Related Tasks: 2,305 instances
\item Named Entity Recognition and Related Tasks: 3,228 instances
\item Credit Scoring and Fraud Detection: 6,461 instances
\item Summarization: 990 instances
\item Trading: 3,384 instances
\end{itemize}

This results in a total of 27,317 instances in the test dataset.

\textcolor{blue}{\subsection{What data does each instance consist of ? “Raw” data (e.g., unprocessed text or images)? Features/attributes? Is there a label/target associated with instances? If the instances related to people, are subpopulations identified (e.g., by age, gender, etc.) and what is
their distribution?}}

\begin{itemize}
\item Information Extraction: Instances are sourced from various financial documents like SEC filings and financial agreements. They include tasks such as named entity recognition, relation extraction, causal classification, and detection, numeric labeling, and textual analogy parsing. These tasks require the model to identify and classify text spans, relationships, and numeric data based on specific instructions. 

\item Textual Analysis: This includes tasks like sentiment analysis, news headline classification, and various forms of argument and claim classification. Instances here are mostly textual data from financial news, earnings reports, and policy documents. They are structured to help models analyze sentiments, classify text into predefined categories, or extract argumentative structures.  

\item Question Answering: Instances include financial reports and tables, challenging models to perform tasks such as multi-step numerical reasoning and multi-turn question answering. These are designed to test a model's ability to navigate complex financial data and provide accurate, context-aware responses. 

\item Text Generation: Focuses on generating coherent summaries from financial texts like earnings call transcripts and news articles. Instances are tailored to test the model's capabilities in condensing information while retaining critical financial insights and facts.

\item Forecasting and Risk Management: These instances involve predicting future financial events or assessing risks from data like historical stock prices and customer transactions. Tasks include stock movement prediction, credit scoring, fraud detection, and financial distress identification.

\item Decision-making: Strategic decision-making instances challenge models to synthesize information for trading strategies. They include simulated trading environments with historical price data and market sentiment, designed to evaluate the profitability and risk management capabilities of financial LLMs. 
\end{itemize}

\textcolor{blue}{\subsection{Is there a label or target associated with each instance? If so, please
provide a description.}}
Each task in the dataset has a specific target or label associated with it, as described below:






\begin{itemize}
\item Information Extraction: Labels identify entities, relationships, causal links, numeric values, or roles within analogies.
\item Textual Analysis: Targets include sentiment classifications, news relevance, argument stances, and categorization of financial texts.
\item Question Answering: Answers or information extracted from financial documents and tables serve as the targets.
\item Text Generation: Summaries or generated texts are evaluated against reference summaries to assess quality and accuracy.
\item Forecasting and Risk Management: Predicted outcomes, risk categories, or fraud indicators are the labels for these tasks.
\item Decision-making: Performance metrics such as returns and risk ratios are the targets in simulated trading scenarios.
\end{itemize}

\textcolor{blue}{\subsection{Is any information missing from individual instances? If so, please
provide a description, explaining why this information is missing (e.g., because it was unavailable). This does not include intentionally removed
information, but might include, e.g., redacted text.}}

In the assembly and preprocessing stages of our current dataset, we have aimed to maintain a comprehensive dataset for each task. Despite these efforts, there are instances where certain information might be missing due to several factors:

\begin{itemize}
\item \textbf{Unavailability:} At times, specific data might not be accessible during collection. For instance, certain financial reports or articles may lack detailed information about company performance or proprietary data.

\item \textbf{Irrelevance:} Occasionally, certain details are deliberately excluded because they do not aid the task objectives. For example, the omission of 'MISCELLANEOUS' entities in Named Entity Recognition tasks, as these do not help in identifying key entities like persons, organizations, or locations.

\item \textbf{Privacy Protection:} In tasks involving stock movement prediction, tweets are sanitized to remove any personally identifiable information or offensive content, adhering to privacy regulations and maintaining ethical research standards.

\item \textbf{Token Limitation:} For tasks such as ConvFinQA, FinQA, and stock price movement prediction, the length of instances is capped. Any data exceeding 2048 tokens is truncated due to the sequence length limitations of LLMs. This truncation is carefully managed to retain the most pertinent information for the task, ensuring that crucial context (e.g., relevant financial news or the specific question context) remains within the token limit.
\end{itemize}

\textcolor{blue}{\subsection{Are relationships between individual instances made explicit (e.g., users’ movie ratings, social network links)? If so, please describe how these relationships are made explicit.}}

In our dataset, each entry is treated as an individual instance with a primary focus on specific tasks such as sentiment analysis, named entity recognition, question answering, and others. Relationships between individual instances are generally not explicitly defined, but certain inherent connections may exist:

\begin{itemize}
\item \textbf{Temporal Relationships:} In tasks like stock movement prediction, instances are connected through their temporal sequence. For example, stock prices and related tweets follow a chronological order, creating an implicit link based on their timestamps.

\item \textbf{Contextual Relationships:} For question answering tasks, such as ConvFinQA and FinQA, questions and their answers may be interconnected as part of a continuous conversation or discussion thread.

\end{itemize}

It is crucial to recognize that while these relationships are inherent to the data, they are not explicitly labeled or annotated within the dataset. The model is expected to discern and utilize these relationships during its training process.

\textcolor{blue}{\subsection{Does the dataset contain all possible instances or is it a sample (not
necessarily random) of instances from a larger set? If the dataset is
a sample, then what is the larger set? Is the sample representative of the
larger set (e.g., geographic coverage)? If so, please describe how this
representativeness was validated/verified. If it is not representative of the
larger set, please describe why not (e.g., to cover a more diverse range of
instances, because instances were withheld or unavailable).}}

The dataset is a sample drawn from a larger set of financial data sources, including financial reports, news articles, stock prices, tweets, and other financial and economic texts. Due to the sheer volume and continuously evolving nature of such data, the dataset does not cover all possible instances from these sources. The samples were selected to enable diverse and comprehensive coverage of typical tasks in the financial domain.

\begin{itemize}
\item \textbf{Representativeness:} The dataset strives to be representative of the larger set of financial data sources in terms of the variety of tasks it covers (e.g., sentiment analysis, entity recognition, question answering, stock movement prediction), modalities (text, time-series data), and types of financial texts (news articles, reports, regulatory filings). 

\item \textbf{Validation:} Due to the dynamic and vast nature of the larger data set, the representativeness of the dataset cannot be fully validated. However, it was designed to include diverse and important tasks in financial analysis, and its effectiveness is evaluated based on the performance of models trained on it.

\item \textbf{Limitations:} Some instances may have been excluded due to data unavailability, restrictions in data sharing agreements, or practical considerations such as token limitations in the model. For example, very lengthy reports or conversations may have been truncated or excluded. Additionally, some types of financial data (e.g., confidential company reports, private communications) are inherently unavailable due to privacy and confidentiality reasons.
\end{itemize}

Despite these limitations, the dataset aims to provide wide coverage of tasks, modalities, and text types, capturing the complexity and diversity of financial data analysis tasks.

\textcolor{blue}{\subsection{Are there recommended data splits (e.g., training, development/validation, testing)? If so, please provide a description of these
splits, explaining the rationale behind them.}}

For our benchmark, we have opted to include only a test split to specifically evaluate the performance of large language models (LLMs). This ensures the evaluation focuses solely on the model's ability to generalize to unseen data, a critical aspect for assessing LLM effectiveness. By excluding training and validation splits, we provide a standardized evaluation framework emphasizing the model's performance in real-world scenarios with new and unseen data. For existing datasets, we use only their test sets, and for our newly created datasets, all data is designated for testing.

\textcolor{blue}{\subsection{Are there any errors, sources of noise, or redundancies in the
dataset? If so, please provide a description.}}

Given the diverse sources of data and the complex nature of financial language, there may be inherent errors, noise, or redundancies in the dataset, including:

\begin{itemize}
    \item Sentiment Ambiguity: In the Financial Sentiment Analysis task, converting sentiment scores into three categories (negative, neutral, positive) can introduce noise. This is because sentiments expressed in financial texts are often subjective and ambiguous.
    \item Named Entity Recognition Noise: The process of discarding miscellaneous entities in the Named Entity Recognition task might inadvertently remove useful information. Additionally, sentences without any entities might introduce further noise.
    \item Question Answering Complexity: In the Question Answering task, the complexity of financial questions and their respective answers may lead to potential errors during annotation. Multiple correct or partially correct answers might exist for a given question, adding to the challenge.
    \item Stock Movement Prediction: In Stock Movement Prediction, price features and tweet data might contain redundancies, as they include multiple overlapping features such as opening, highest, lowest, closing, and adjusted closing prices. Moreover, stock movement predictions are influenced by various factors, not all of which may be captured in the features used, leading to potential errors.
    \item Token Limit: For tasks such as ConvFinQA, FinQA, and Stock Movement Prediction, instances are truncated to 2048 tokens, which might result in the loss of information and subsequently introduce errors in these tasks.
\end{itemize}

Despite these potential issues, the datasets remain valuable resources for developing and benchmarking models for financial tasks, provided these limitations are considered during the model development and evaluation process.

\textcolor{blue}{\subsection{Is the dataset self-contained, or does it link to or otherwise rely on external resources (e.g., websites, tweets, other datasets)? If it links to or relies on external resources, a) are there guarantees that they will exist, and remain constant, over time; b) are there official archival versions of the complete dataset (i.e., including the external resources as they existed at the time the dataset was created); c) are there any restrictions (e.g., licenses, fees) associated with any of the external resources that might apply to a future user? Please provide descriptions of all external resources and any restrictions associated with them, as well as links or other access points, as appropriate.}}

The dataset originates from external sources like financial news articles, tweets, and stock market data, but it has been preprocessed, labeled, and stored independently, making it self-contained. This ensures no direct links to the original sources, mitigating issues of persistence, archival consistency, and restrictions.

Hosted on a GitHub repository, the dataset is easily accessible and benefits from GitHub's version-control features, preserving its state at release. Users can clone or fork the repository under the specified terms and conditions.

Despite its external origins, the final dataset is self-contained, accessible, and minimally restricted.

\textcolor{blue}{Any other comments?}

The dataset for each task balances comprehensiveness and specificity well. Key points include:

\begin{itemize}
    \item Diversity: The dataset includes diverse text data from various sources, ensuring broad coverage of financial terms and expressions, particularly in sentiment analysis and news headline classification tasks.
    \item Representation: It represents real-world financial NLP challenges, with instances carefully chosen to simulate tasks like sentiment analysis, entity recognition, question answering, and stock movement prediction.
    \item Comprehensiveness: Covering a wide range of tasks from text classification to complex question answering and stock movement prediction, the dataset addresses various financial NLP problems.
    \item Size: It is sufficiently large, providing ample instances for training, validation, and testing robust models.
\end{itemize}

These factors enhance the dataset's utility and value for financial LLMs.

\section{Collection Process}
\textcolor{blue}{\subsection{What mechanisms or procedures were used to collect the data (e.g., hardware apparatus or sensor, manual human curation, software program, software API)? How were these mechanisms or procedures validated?}}

The dataset has been assembled from publicly available datasets and resources, supplemented by the manual creation of instruction templates. More specifically:

\begin{itemize}
\item \textbf{Publicly Available Datasets:} Existing datasets have been utilized in our tasks, including financial sentiment analysis, news headline classification, named entity recognition and question answering. These datasets are widely acknowledged and utilized in the research community, validating their reliability and quality. The raw datasets are provided along with our work for reference.

\item \textbf{Manual Creation of Instruction Templates:} For tasks that rely on conversational models, we have manually created instruction templates. These templates guide the model's responses and ensure that the output is consistent with the task's requirements. 

\end{itemize}

The full details, including the source of the raw datasets, are provided in the documentation accompanying our GitHub repository. The use of publicly available datasets, coupled with manual curation of instructions, helps ensure the data's reliability and robustness for our tasks.

\textcolor{blue}{\subsection{How was the data associated with each instance acquired? Was the data directly observable (e.g., raw text, movie ratings), reported by subjects (e.g., survey responses), or indirectly inferred/derived from other data (e.g., part-of-speech tags, model-based guesses for age or language)? If data was reported by subjects or indirectly inferred/derived from other data, was the data validated/verified? If so, please describe how.}}

We constructed the multi-task and multi-modal instruction data by collecting publicly available training data from a range of diverse tasks. This data was directly observable and derived from multiple open-released financial datasets. The data we used included both textual and time-series data modalities, which allowed our model to handle tasks such as sentiment analysis, news headline classification, named entity recognition, question answering, and stock movement prediction. For each task, we wrote specific instructions that were combined with the data samples to create our large-scale instruction tuning data.

The instructions for each task were carefully designed by domain experts to ensure that they accurately reflect the nuances and requirements of the different tasks. This approach allowed us to tailor our large language model, FinMA, to perform a diverse range of financial tasks.

\textcolor{blue}{\subsection{If the dataset is a sample from a larger set, what was the sampling strategy (e.g., deterministic, probabilistic with specific sampling probabilities)?}}

Our dataset is not a sample from a larger dataset but an assembly of publicly available multi-task and multi-modal data derived from multiple open-released financial datasets. We didn't use a specific sampling strategy since we weren't sampling from a larger set. Instead, we collected the complete datasets that were available and relevant to our study

\textcolor{blue}{\subsection{Who was involved in the data collection process (e.g., students,
crowd workers, contractors) and how were they compensated (e.g.,
how much were crowd workers paid)?}}

 For the FinBen project, our data collection process involved both domain experts and students. The domain experts, who are professionals in the financial field such as fund managers, were invited to design the task-specific instructions for each dataset. On the other hand, students were responsible for the collection of publicly available datasets.

This project was conducted as collaborative research, so those who contributed to the data collection and instruction design were not compensated in a traditional paid manner. Instead, they contributed their expertise and time to the project as part of their research activities or professional engagement.

\textcolor{blue}{\subsection{Over what timeframe was the data collected? Does this timeframe
match the creation timeframe of the data associated with the instances
(e.g., a recent crawl of old news articles)? If not, please describe the timeframe in which the data associated with the instances was created.}}
\begin{itemize}
\item FPB, while it does not provide a specific timeframe for data collection, was published in 2014.
\item FIQASA, like FPB, does not specify a timeframe for data collection, but it was published in 2018.
\item The Headlines dataset contains human-annotated news headlines that were collected over a span of 19 years, from 2000 to 2019. It was published in 2021.
\item NER, again, does not indicate a specific timeframe for data collection, but it was published in 2015.
\item The FinQA dataset spans two decades, collecting data from 1999 to 2019. It was published in 2021.
\item ConvFinQA, like FinQA, covers a period from 1999 to 2019 and was published in 2022.
\item The BigData22 dataset has a more confined range of data collection, from July 5th, 2019 to June 30th, 2020.
\item The ACL18 dataset spans approximately two years, from January 2nd, 2014 to December 30th, 2015.
\item The CIKM18 dataset covers data collected within a single year, from January 3rd, 2017 to December 28th, 2017.
\item FOMC, with the meeting minutes and speeches data spanned from January 1st, 1996 to October 15th, 2022, and the press conferences data from April 27th, 2011 to October 15th, 2022, was published in 2023. 
\item FinArg-ACC does not specify a timeframe for data collection, but it was published in 2023.
\item FinArg-ARC does not specify a timeframe for data collection, but it was published in 2023.
\item MultiFin, covering a period from 2015 to 2021, was published in 2023.
\item The MA dataset covers data collected from January 1st, 2007 to August 12th, 2019, and was published in 2020.
\item MLESG, again, does not specify a timeframe for data collection, but it was published in 2023.
\item FiNER-ORD does not specify a timeframe for data collection, but it was published in 2023.
\item FinRED collected the raw data during the timespan of 2015,2019, respectively. it was published in 2023.
\item SC extracted data during 2019 and was published in 2020.
\item CD extracted data during 2019 and was published in 2020.
\item TATQA collected data from 2019 to 2020 and was published in 2021.
\item The ECTSum dataset spans over three years, from January 2019 to April 2022.
\item EDTSum does not specify a timeframe for data collection, but it was published in 2021.
\item The fintrade dataset spans approximately two years, from August 15, 2021, to April 25, 2023. it was published in 2023.
\item FNXL covering a period from 2019 to 2021, was published in 2023.
\end{itemize}

In summary, the data associated with the instances in the FinBen dataset spans a wide range of timeframes, from specific periods within a year to a stretch of 20 years. Some of these match the publication years of their respective sources, like the Headlines dataset, FinQA, and ConvFinQA, while others, like FPB, FIQASA, and NER, do not specify a timeframe for data collection. Therefore, the creation timeframe of the data associated with the instances varies, and in some cases, it may be assumed to be close to their respective publication years.

\section{Data Preprocessing}

\textcolor{blue}{\subsection{Was any preprocessing/cleaning/labeling of the data done (e.g., discretization or bucketing, tokenization, part-of-speech tagging, SIFT
feature extraction, removal of instances, processing of missing values)? If so, please provide a description. If not, you may skip the remainder of the questions in this section.}}
To convert raw data into a structured instruction dataset, we defined a pipeline with the following steps:

\begin{enumerate}
    \item \textbf{Define Instruction Templates:} We began by defining clear and succinct instruction templates for each task.
    
    \item \textbf{Extract Relevant Information:} The next step involved extracting the requisite pieces of information from the raw data. These would be used to fill in the instruction templates.
    
    \item \textbf{Match to Template:} Once the relevant information was extracted, it was matched to the instruction template.
    
    \item \textbf{Verify and Clean Instructions:} Following the creation of the instructions, we verified that they were both coherent and accurate representations of the task.
    
    \item \textbf{Standardize Instructions:} Finally, we standardized the instructions in terms of language, structure, and style. This ensured consistency across the dataset, which subsequently enabled the model to understand and learn the tasks more efficiently.
\end{enumerate}

This pipeline was utilized to process each task within the FinBen model. For example, consider a task involving sentiment analysis. The raw data consisted of a series of tweets along with their associated sentiment labels. The instruction template for this task could be: ``Classify the sentiment of the following tweet as 'positive', 'negative', or 'neutral'". The preprocessing steps involved extracting the tweet text, fitting it into the template, and verifying that the resulting instruction accurately represented the task.

\textcolor{blue}{\subsection{Was the “raw” data saved in addition to the preprocessed/cleaned/labeled data (e.g., to support unanticipated
future uses)? If so, please provide a link or other access point to the
“raw” data.}}

The raw data can be accessed through the original papers:

\begin{enumerate}
\item \textbf{NER}: Julio Cesar Salinas Alvarado, Karin Verspoor, and Timothy Baldwin. 2015. Domain adaption of named entity recognition to support credit risk assessment. In Proceedings of the Australasian Language Technology Association Workshop 2015. 84–90.
\item \textbf{FiNER-ORD}: Agam Shah, Ruchit Vithani, Abhinav Gullapalli, and Sudheer Chava. 2023b. Finer: Financial named entity recognition dataset and weak-supervision model. arXiv preprint arXiv:2302.11157 (2023).
\item \textbf{FinRED}: Soumya Sharma, Tapas Nayak, Arusarka Bose, Ajay Kumar Meena, Koustuv Dasgupta, Niloy
Ganguly, and Pawan Goyal. 2022. FinRED: A dataset for relation extraction in financial domain.
In Companion Proceedings of the Web Conference 2022. 595–597.
\item \textbf{SC, CD}: Dominique Mariko, Hanna Abi Akl, Estelle Labidurie, Stephane Durfort, Hugues De Mazancourt, and Mahmoud El-Haj. 2020. Financial document causality detection shared task (fincausal 2020). arXiv preprint arXiv:2012.02505 (2020).
\item \textbf{FNXL}: Soumya Sharma, Subhendu Khatuya, Manjunath Hegde, Afreen Shaikh, Koustuv Dasgupta, Pawan Goyal, and Niloy Ganguly. 2023. Financial Numeric Extreme Labelling: A dataset and bench-marking. In Findings of the Association for Computational Linguistics: ACL 2023. 3550–3561.
\item \textbf{FSRL}: Matthew Lamm, Arun Tejasvi Chaganty, Christopher D Manning, Dan Jurafsky, and Percy Liang. 2018. Textual analogy parsing: What’s shared and what’s compared among analogous facts. arXiv
preprint arXiv:1809.02700 (2018).
\item \textbf{FPB}: Pekka Malo, Ankur Sinha, Pekka Korhonen, Jyrki Wallenius, and Pyry Takala. 2014. Good debt or bad debt: Detecting semantic orientations in economic texts. Journal of the Association for
Information Science and Technology 65, 4 (2014), 782–796.
\item \textbf{FiQA-SA}: Macedo Maia, Siegfried Handschuh, Andre Freitas, Brian Davis, Ross McDermott, Manel Zarrouk, and Alexandra Balahur. 2018B. WWW’18 Open Challenge: Financial Opinion Mining and Question Answering. WWW’18: Companion Proceedings of the The Web Conference 2018, 1941–1942.
\item \textbf{TSA}: Keith Cortis, André Freitas, Tobias Daudert, Manuela Huerlimann, Manel Zarrouk, Siegfried Hand-schuh, and Brian Davis. 2017. Semeval-2017 task 5: Fine-grained sentiment analysis on financial microblogs and news. In Proceedings of the 11th international workshop on semantic evaluation (SemEval-2017). 519–535.
\item \textbf{Headlines}: Ankur Sinha and Tanmay Khandait. 2021. Impact of news on the commodity market: Dataset and results. In Advances in Information and Communication: Proceedings of the 2021 Future of Information and Communication Conference (FICC), Volume 2. Springer, 589–601.
\item \textbf{FOMC}: Agam Shah, Suvan Paturi, and Sudheer Chava. 2023a. Trillion Dollar Words: A New Financial Dataset, Task \& Market Analysis. In Proceedings of the 61st Annual Meeting of the Association for Computational Linguistics (Volume 1: Long Papers), Anna Rogers, Jordan Boyd-Graber, and Naoaki Okazaki (Eds.). Association for Computational Linguistics, Toronto, Canada, 6664–6679
\item \textbf{FinArg-ACC, FinArg-ARC}: Eugene Sy, Tzu-Cheng Peng, Shih-Hsuan Huang, Heng-Yu Lin, and Yung-Chun Chang. 2023. Fine-Grained Argument Understanding with BERT Ensemble Techniques: A Deep Dive into Financial Sentiment Analysis. In Proceedings of the 35th Conference on Computational Linguistics and Speech Processing (ROCLING 2023). 242–249.
\item \textbf{MultiFin}: Rasmus Jørgensen, Oliver Brandt, Mareike Hartmann, Xiang Dai, Christian Igel, and Desmond Elliott. 2023. MultiFin: A Dataset for Multilingual Financial NLP. In Findings of the Association for Computational Linguistics: EACL 2023. 864–879.
\item \textbf{MA}: Linyi Yang, Eoin M Kenny, Tin Lok James Ng, Yi Yang, Barry Smyth, and Ruihai Dong. 2020a. Generating plausible counterfactual explanations for deep transformers in financial text classification.
arXiv preprint arXiv:2010.12512 (2020).
\item \textbf{MLESG}: Chung-Chi Chen, Yu-Min Tseng, Juyeon Kang, Anaïs Lhuissier, Min-Yuh Day, Teng-Tsai Tu, and Hsin-Hsi Chen. 2023a. Multi-Lingual ESG Issue Identification. In Proceedings of the Fifth Workshop on Financial Technology and Natural Language Processing and the Second Multimodal AI For Financial Forecasting. 111–115.
\item \textbf{FinQA}: Zhiyu Chen, Wenhu Chen, Charese Smiley, Sameena Shah, Iana Borova, Dylan Langdon, Reema
Moussa, Matt Beane, Ting-Hao Huang, Bryan R Routledge, et al. 2021. FinQA: A Dataset of Numerical Reasoning over Financial Data. In Proceedings of the 2021 Conference on Empirical Methods in Natural Language Processing. 3697–3711.
\item \textbf{TATQA}: Fengbin Zhu, Wenqiang Lei, Youcheng Huang, Chao Wang, Shuo Zhang, Jiancheng Lv, Fuli Feng, and Tat-Seng Chua. 2021. TAT-QA: A question answering benchmark on a hybrid of tabular and textual content in finance. arXiv preprint arXiv:2105.07624 (2021).
\item \textbf{ConvFinQA}: Zhiyu Chen, Shiyang Li, Charese Smiley, Zhiqiang Ma, Sameena Shah, and William Yang Wang.
2022. ConvFinQA: Exploring the Chain of Numerical Reasoning in Conversational Finance Question Answering. arXiv:2210.03849 [cs.CL]
\item \textbf{ECTSum}: Rajdeep Mukherjee, Abhinav Bohra, Akash Banerjee, Soumya Sharma, Manjunath Hegde, Afreen Shaikh, Shivani Shrivastava, Koustuv Dasgupta, Niloy Ganguly, Saptarshi Ghosh, et al. 2022. Ectsum: A new benchmark dataset for bullet point summarization of long earnings call transcripts. arXiv preprint arXiv:2210.12467 (2022).
\item \textbf{BigData22}: Yejun Soun, Jaemin Yoo, Minyong Cho, Jihyeong Jeon, and U Kang. 2022. Accurate Stock Movement Prediction with Self-supervised Learning from Sparse Noisy Tweets. In 2022 IEEE International Conference on Big Data (Big Data). IEEE, 1691–1700.
\item \textbf{ACL18}: Yumo Xu and Shay B Cohen. 2018. Stock movement prediction from tweets and historical prices. In Proceedings of the 56th Annual Meeting of the Association for Computational Linguistics (Volume 1: Long Papers). 1970–1979.
\item \textbf{CIKM18}: Huizhe Wu, Wei Zhang, Weiwei Shen, and Jun Wang. 2018. Hybrid deep sequential modeling for social text-driven stock prediction. In Proceedings of the 27th ACM international conference on information and knowledge management. 1627–1630.
\item \textbf{German}: Hans Hofmann. 1994. Statlog (German Credit Data). UCI Machine Learning Repository. DOI: https://doi.org/10.24432/C5NC77.
\item \textbf{Australian}: Ross Quinlan. [n. d.]. Statlog (Australian Credit Approval). UCI Machine Learning Repository. DOI: https://doi.org/10.24432/C59012.
\item \textbf{LendingClub}: Duanyu Feng, Yongfu Dai, Jimin Huang, Yifang Zhang, Qianqian Xie, Weiguang Han, Alejandro Lopez-Lira, and Hao Wang. 2023. Empowering many, biasing a few: Generalist credit scoring through large language models. arXiv preprint arXiv:2310.00566 (2023).
\item \textbf{ccf}: Duanyu Feng, Yongfu Dai, Jimin Huang, Yifang Zhang, Qianqian Xie, Weiguang Han, Alejandro Lopez-Lira, and Hao Wang. 2023. Empowering many, biasing a few: Generalist credit scoring through large language models. arXiv preprint arXiv:2310.00566 (2023).
\item \textbf{ccfraud}: Duanyu Feng, Yongfu Dai, Jimin Huang, Yifang Zhang, Qianqian Xie, Weiguang Han, Alejandro Lopez-Lira, and Hao Wang. 2023. Empowering many, biasing a few: Generalist credit scoring through large language models. arXiv preprint arXiv:2310.00566 (2023).
\item \textbf{polish}: Duanyu Feng, Yongfu Dai, Jimin Huang, Yifang Zhang, Qianqian Xie, Weiguang Han, Alejandro Lopez-Lira, and Hao Wang. 2023. Empowering many, biasing a few: Generalist credit scoring through large language models. arXiv preprint arXiv:2310.00566 (2023).
\item \textbf{taiwan}: Duanyu Feng, Yongfu Dai, Jimin Huang, Yifang Zhang, Qianqian Xie, Weiguang Han, Alejandro Lopez-Lira, and Hao Wang. 2023. Empowering many, biasing a few: Generalist credit scoring through large language models. arXiv preprint arXiv:2310.00566 (2023).
\item \textbf{ProtoSeguro}: Duanyu Feng, Yongfu Dai, Jimin Huang, Yifang Zhang, Qianqian Xie, Weiguang Han, Alejandro Lopez-Lira, and Hao Wang. 2023. Empowering many, biasing a few: Generalist credit scoring through large language models. arXiv preprint arXiv:2310.00566 (2023).
\item \textbf{travelinsurance}: Duanyu Feng, Yongfu Dai, Jimin Huang, Yifang Zhang, Qianqian Xie, Weiguang Han, Alejandro Lopez-Lira, and Hao Wang. 2023. Empowering many, biasing a few: Generalist credit scoring through large language models. arXiv preprint arXiv:2310.00566 (2023).
\end{enumerate}

\textcolor{blue}{\subsection{Is the software used to preprocess/clean/label the instances available? If so, please provide a link or other access point.}}

Yes, the software used for preprocessing, cleaning, and labeling the instances is publicly available. It can be accessed via our GitHub repository at \href{https://github.com/The-FinAI/PIXIU}{https://github.com/The-FinAI/PIXIU}.

\textcolor{blue}{\subsection{Does this dataset collection/processing procedure
achieve the motivation for creating the dataset
stated in the first section of this datasheet? If not,
what are the limitations?}}

Yes, the dataset collection and processing procedure does fulfill the initial motivation. We have created a comprehensive set of resources specifically tailored for the financial sector. However, as with any novel project, there are limitations. The financial domain is vast and complex, and while FinBen covers a broad range of topics, it may not encompass all possible financial tasks or scenarios. We hope that future iterations of FinBen can further expand and diversify the instruction dataset and evaluation benchmarks.

\textcolor{blue}{\subsection{Any other comments}}
No.

\section{Dataset Distribution}

\textcolor{blue}{\subsection{How will the dataset be distributed? (e.g., tarball on
website, API, GitHub; does the data have a DOI and is it
archived redundantly?)}}
The dataset, along with the necessary code, will be distributed through our GitHub repository (\url{https://github.com/The-FinAI/PIXIU}).

\textcolor{blue}{\subsection{When will the dataset be released/first distributed?
What license (if any) is it distributed under?}}
The dataset was first released on February 20th, 2024. It is distributed under the MIT License, which permits use, copy, modify, merge, publish, distribute, sublicense, and/or sell copies of the Software with few restrictions, provided that the above copyright notice and this permission notice are included in all copies or substantial portions of the Software.

\textcolor{blue}{\subsection{Are there any copyrights on the data?}}
 The transformed and aggregated dataset, as well as the code and models we provide, are released under the MIT License.

\textcolor{blue}{\subsection{Are there any fees or access/export restrictions?}}
There are no fees associated with the FinBen, and it can be freely accessed and exported for research purposes.

\textcolor{blue}{\subsection{Any other comments?}}
We encourage researchers to use this dataset and related resources to advance the field of financial AI. However, proper attribution should be given when using these resources in line with standard academic practices.

\section{Dataset Maintenance}

\textcolor{blue}{\subsection{Who is supporting/hosting/maintaining the
dataset?}}
The dataset is maintained and supported by a team of researchers from various institutions. 

The team includes:

Qianqian Xie$^{a}$,
Weiguang Han$^b$,  
Zhengyu Chen$^b$,   
Ruoyu Xiang$^a$, 
Xiao Zhang$^a$,
Yueru He$^a$,\\
Mengxi Xiao$^b$, Dong Li$^b$, Yongfu Dai$^g$, Duanyu Feng$^g$, Yijing Xu$^a$, Haoqiang Kang$^e$,\\
Ziyan Kuang$^l$, Chenhan Yuan$^c$, Kailai Yang$^c$, Zheheng Luo$^c$, Tianlin Zhang$^c$,\\
Zhiwei Liu$^c$, Guojun Xiong$^j$, Zhiyang Deng$^i$, Yuechen Jiang$^i$, Zhiyuan Yao$^i$,\\
Haohang Li$^i$, Yangyang Yu$^i$, Gang Hu$^h$, Jiajia Huang$^k$, Xiao-Yang Liu$^e$,\\
Alejandro Lopez-Lira$^d$, Benyou Wang$^f$, Yanzhao Lai$^m$, Hao Wang$^g$, Min Peng$^b$,\\
Sophia Ananiadou$^c$, Jimin Huang$^a$.

And the indicators for institutions are:

$^a$The Fin AI, $^b$Wuhan University, $^c$The University of Manchester, $^d$University of Florida,
$^e$Columbia University, $^f$The Chinese University of Hong Kong, Shenzhen,\\
$^g$Sichuan University, $^h$Yunnan University, $^i$Stevens Institute of Technology\\
$^j$Stony Brook University, $^k$Nanjing Audit University,\\
$^l$Jiangxi Normal University, $^m$Southwest Jiaotong University.

\textcolor{blue}{\subsection{Will the dataset be updated? If so, how often and
by whom?}}
The updates will be managed by Zhengyu Chen from the School of Computer Science, Wuhan University. The frequency of updates will be determined by the availability of new data and feedback from users.

\textcolor{blue}{\subsection{How will updates be communicated? (e.g., mailing
list, GitHub)}}
Updates about the dataset will be communicated through GitHub, where the dataset is hosted. Users are encouraged to keep track of the GitHub repository for any updates.

\textcolor{blue}{\subsection{If the dataset becomes obsolete how will this be
communicated?}}
If the dataset becomes obsolete, this will be communicated through the same GitHub repository where the dataset is hosted.

\textcolor{blue}{\subsection{Is there a repository to link to any/all papers/systems that use this dataset?}}
Currently, there is no specific repository for linking to papers or systems that use the dataset. However, users are encouraged to cite the FinBen paper in their work.

\textcolor{blue}{\subsection{If others want to extend/augment/build on this
dataset, is there a mechanism for them to do so?
If so, is there a process for tracking/assessing the
quality of those contributions? What is the process
for communicating/distributing these contributions
to users?}}

The dataset is open source and users are welcome to extend, augment, or build on it. Users can submit pull requests to the GitHub repository where the dataset is hosted. The quality of contributions will be assessed by the maintenance team before being integrated into the dataset. These contributions, once approved, will be made available to other users through the GitHub repository.

\section{Legal and Ethical Considerations}

\textcolor{blue}{\subsection{Were any ethical review processes conducted (e.g., by an institutional review board)? If so, please provide a description of these review
processes, including the outcomes, as well as a link or other access point
to any supporting documentation.}}

No specific ethical review processes were conducted for the development of FinBen as it leverages previously published and publicly accessible datasets. These original datasets underwent their own ethical review processes.

\textcolor{blue}{\subsection{Does the dataset contain data that might be considered confidential
(e.g., data that is protected by legal privilege or by doctor-patient confidentiality, data that includes the content of individuals non-public
communications)? If so, please provide a description.}}
The datasets used in FinBen do not contain any data that might be considered confidential. Any potential personal information in the original datasets has been removed by the original authors before they were made publicly accessible.

\textcolor{blue}{\subsection{Does the dataset contain data that, if viewed directly, might be offensive, insulting, threatening, or might otherwise cause anxiety? If so,
please describe why}}
The datasets used in FinBen do not contain data that, if viewed directly, might be offensive, insulting, threatening, or might otherwise cause anxiety. The original authors of the datasets ensured this during their preprocessing and cleaning stages.

\textcolor{blue}{\subsection{Does the dataset relate to people? If not, you may skip the remaining
questions in this section.}}
While some of the datasets may originally have pertained to people, for instance, those containing tweets, any direct identifiers have been removed from the original datasets. Therefore, FinBen does not directly relate to identifiable individuals.

\textcolor{blue}{\subsection{Does the dataset identify any subpopulations (e.g., by age, gender)?
If so, please describe how these subpopulations are identified and provide
a description of their respective distributions within the dataset.}}
No.

\textcolor{blue}{\subsection{Is it possible to identify individuals (i.e., one or more natural persons), either directly or indirectly (i.e., in combination with other
data) from the dataset? If so, please describe how.}}
It is not possible to identify individuals either directly or indirectly from the datasets used in FinBen. All potentially identifiable information was removed from the original datasets before they were made publicly accessible.

\textcolor{blue}{\subsection{Does the dataset contain data that might be considered sensitive in
any way (e.g., data that reveals racial or ethnic origins, sexual orientations, religious beliefs, political opinions or union memberships, or
locations; financial or health data; biometric or genetic data; forms of
government identification, such as social security numbers; criminal
history)? If so, please provide a description.}}
The datasets used in FinBen do not contain data that might be considered sensitive in any way. All potentially sensitive information was removed from the original datasets before they were made publicly accessible.

\textcolor{blue}{\subsection{Did you collect the data from the individuals in question directly, or
obtain it via third parties or other sources (e.g., websites)?}}
The data used in FinBen was not directly collected from individuals. Instead, it was obtained from previously published datasets that were collected by other researchers and made publicly available.

\textcolor{blue}{\subsection{Were the individuals in question notified about the data collection?
If so, please describe (or show with screenshots or other information) how
notice was provided, and provide a link or other access point to, or otherwise reproduce, the exact language of the notification itself.}}
As the data used in FinBen were obtained from previously published datasets, the original data collectors were responsible for notifying individuals about the data collection.

\textcolor{blue}{\subsection{Did the individuals in question consent to the collection and use of
their data? If so, please describe (or show with screenshots or other
information) how consent was requested and provided, and provide a link
or other access point to, or otherwise reproduce, the exact language to
which the individuals consented.}}
As the data used in FinBen were obtained from previously published datasets, the original data collectors were responsible for obtaining consent from individuals for the collection and use of their data.

\textcolor{blue}{\subsection{If consent was obtained, were the consenting individuals provided
with a mechanism to revoke their consent in the future or for certain
uses? If so, please provide a description, as well as a link or other access
point to the mechanism (if appropriate).}}
The issue of revoking consent does not directly apply to FinBen as the data used was obtained from previously published datasets. The original data collectors were responsible for providing mechanisms for revoking consent.

\textcolor{blue}{\subsection{Has an analysis of the potential impact of the dataset and its use
on data subjects (e.g., a data protection impact analysis)been conducted? If so, please provide a description of this analysis, including the
outcomes, as well as a link or other access point to any supporting documentation.}}
Given that the data used in FinBen is de-identified and does not contain sensitive information, it is not anticipated that there will be direct impacts on data subjects. However, as with any research involving human-related data, there is always a responsibility to use the data ethically and with respect to potential implications.

\textcolor{blue}{\subsection{Any other comments?}}
No.

\medskip

\section{Responsibility Statement}
The authors of FinBen bear all responsibility in case of any violation of rights or any other legal issues that arise from the use of this dataset. The authors have taken all possible measures to ensure the respect of privacy and ethical guidelines in the construction of this dataset.

The datasets included in FinBen are distributed under the MIT license. By using these datasets, users agree to comply with the terms of this license.

\textcolor{black}{
This paper, including any associated source codes, datasets, and appendices ("Material"), is intended solely for academic and educational purposes. The Material does not constitute financial, legal, or investment advice and is not intended to be a basis for any decision-making.}

\textcolor{black}{While the authors have taken reasonable measures to ensure the accuracy of the Material, no warranty, express or implied, is made as to its completeness, reliability, or suitability for any specific purpose. The authors and their affiliated organizations shall not be liable for any losses, damages, or consequences, whether direct or indirect, arising from the use or reliance on the Material. It is the responsibility of the user to consult with professionals for any financial, legal, or investment decisions.}

\textcolor{black}{By referencing or utilizing this Material, the reader agrees to indemnify, defend, and hold harmless the authors and any affiliated organizations or persons from any and all claims or damages arising from such use.
}

\section{Reproducibility}

Ensuring the reproducibility of research results is of utmost importance in promoting transparency and enabling further scientific advancements. In the context of the benchmarks presented in this datasheet, we have taken several measures to facilitate the reproducibility of our reported results.

To begin with, we have made all the necessary code, data, and instructions available in the FinBen GitHub repository (\url{https://github.com/The-FinAI/PIXIU}). This repository serves as a centralized hub where researchers can access the resources required to reproduce the benchmarks. The repository is organized and well-documented, providing a clear and structured framework for replication.

In order to enhance reproducibility, we have adhered to the ML reproducibility checklist, a framework that promotes best practices for ensuring the replicability of machine learning experiments. By following this checklist, we have prioritized the inclusion of all essential components, such as code, datasets, and evaluation procedures, making it easier for researchers to reproduce our reported results.

Furthermore, we have provided detailed instructions within the repository, outlining the steps needed to replicate the benchmarks. These instructions serve as a guide for researchers, ensuring that they have the necessary information and resources at their disposal to validate and verify our findings.

We encourage researchers to leverage the resources available in the FinBen GitHub repository to replicate our benchmarks and explore further extensions and improvements. By promoting a culture of reproducibility, we aim to foster collaboration and drive advancements in the field of financial artificial intelligence.

\section{Response to Previous Review Concerns}

In the previous submission to ACL ARR 2024 April, reviewers raised a few minor concerns about the justification of the datasets and the depth of insights derived. To address these, the following improvements have been made:

\textbf{Enhanced Dataset Justification:} The paper now includes a more detailed motivation for the inclusion of each dataset, highlighting their unique contributions and relevance to the financial sector.

\textbf{Incorporation of New Findings:} The paper presents new findings and insights derived from the comprehensive benchmark, showcasing how these results add to the current knowledge in the field of financial LLMs.

By addressing these concerns, the revised version of the paper aims to provide a more robust and insightful evaluation of LLMs in the financial sector while building on the previously recognized strengths.